\documentclass[10pt,journal,compsoc]{IEEEtran}
\ifCLASSOPTIONcompsoc
  \usepackage[nocompress]{cite}
\else
  \usepackage{cite}
\fi
\ifCLASSINFOpdf
\else
\fi

\usepackage{times}
\usepackage{epsfig}
\usepackage{graphicx}
\usepackage{amsmath}
\usepackage{amssymb}
\usepackage{subfigure}
\usepackage{multirow}
\usepackage{booktabs}
\usepackage{color}
\usepackage[colorinlistoftodos]{todonotes}

\begin{document}
%
\title{AGConv: Adaptive Graph Convolution on 3D Point Clouds}
%
%
%
%

\author{\renewcommand{\thefootnote}{\fnsymbol{footnote}} Mingqiang Wei, Zeyong Wei, Haoran Zhou, Fei Hu, Huajian Si, Zhilei Chen, Zhe Zhu, Jingbo Qiu, Xuefeng Yan, Yanwen Guo, Jun Wang, and Jing Qin

 \thanks{M. Wei, Z. Wei, F. Hu, H. Si, Z. Chen, Z. Zhu, J. Qiu, X. Yan and J. Wang are with the School of Computer Science and Technology, Nanjing University of Aeronautics and Astronautics, Nanjing, China.}
\thanks{H. Zhou and Y. Guo are with the State Key Laboratory for Novel Software Technology, Nanjing University, Nanjing, China.}
 \thanks{J. Qin is with the Hong Kong Polytechnic University, Hong Kong, China.
 }

}

\IEEEtitleabstractindextext{%
\begin{abstract}
Convolution on 3D point clouds is widely researched yet far from perfect in geometric deep learning. 
The traditional wisdom of convolution characterises feature correspondences indistinguishably among 3D points, arising an intrinsic limitation of poor distinctive feature learning. 
In this paper, we propose Adaptive Graph Convolution (AGConv) for wide applications of point cloud analysis. 
AGConv generates adaptive kernels for points according to their dynamically learned features. 
Compared with the solution of using fixed/isotropic kernels, AGConv improves the flexibility of point cloud convolutions, effectively and precisely capturing the diverse relations between points from different semantic parts. 
Unlike the popular attentional weight schemes, AGConv implements the adaptiveness inside the convolution operation instead of simply assigning different weights to the neighboring points. 
Extensive evaluations clearly show that our method outperforms state-of-the-arts of point cloud classification and segmentation on various benchmark datasets.
Meanwhile, AGConv can flexibly serve more point cloud analysis approaches to boost their performance. To validate its flexibility and effectiveness, we explore AGConv-based paradigms of completion, denoising, upsampling, registration and circle extraction, which are comparable or even superior to their competitors.
Our code is available at \emph{\textcolor{magenta}{ https://github.com/hrzhou2/AdaptConv-master}}.
\end{abstract}

}

\maketitle

\IEEEdisplaynontitleabstractindextext

%
\IEEEpeerreviewmaketitle

\IEEEraisesectionheading{\section{Introduction}\label{sec:introduction}}
\IEEEPARstart{P}oint clouds are a standard output of 3D sensors, e.g., LiDAR scanners and RGB-D cameras \cite{pami/JiangZD20}. They preserve the original geometric information of objects in 3D space with a very simple and flexible data structure \cite{corr/abs-2112-04148}. 
A variety of applications, such as robotics, autonomous driving, and Metaverse, arise with the fast advance of point cloud acquisition techniques.
%
Recent years have witnessed considerable attempts to generalize convolutional neural networks (CNNs) to point clouds for 3D analysis and generation \cite{cvm/GuoCLMMH21,pami/GuoWHLLB21}. 
However, convolution on point clouds is still far from perfect, since unlike 2D images organized as regular grid-like structures, 3D points are unstructured and unordered, discretely distributed on the underlying surfaces of sampled objects.
%

The common ways of learning on point clouds are to convert them into regular 2D grids, 3D voxels or to develop hand-crafted feature descriptors, on which traditional 2D/3D CNNs can be naturally applied \cite{riegler2017octnet, cvpr/ZhouCF000WW020,cad/LiZFXWWH20,pami/haoranzhou2022}. 
Such solutions, however, often introduce excessive memory cost, and are difficult to capture fine-grained geometric details. 
To handle the irregularity of point clouds without conversions, PointNet \cite{QiSMG17} applies multi-layer perceptrons (MLPs) independently on each point, which is the pioneering work to directly process sparse 3D points.
\begin{figure}[t]
	\includegraphics[width=0.99\linewidth]{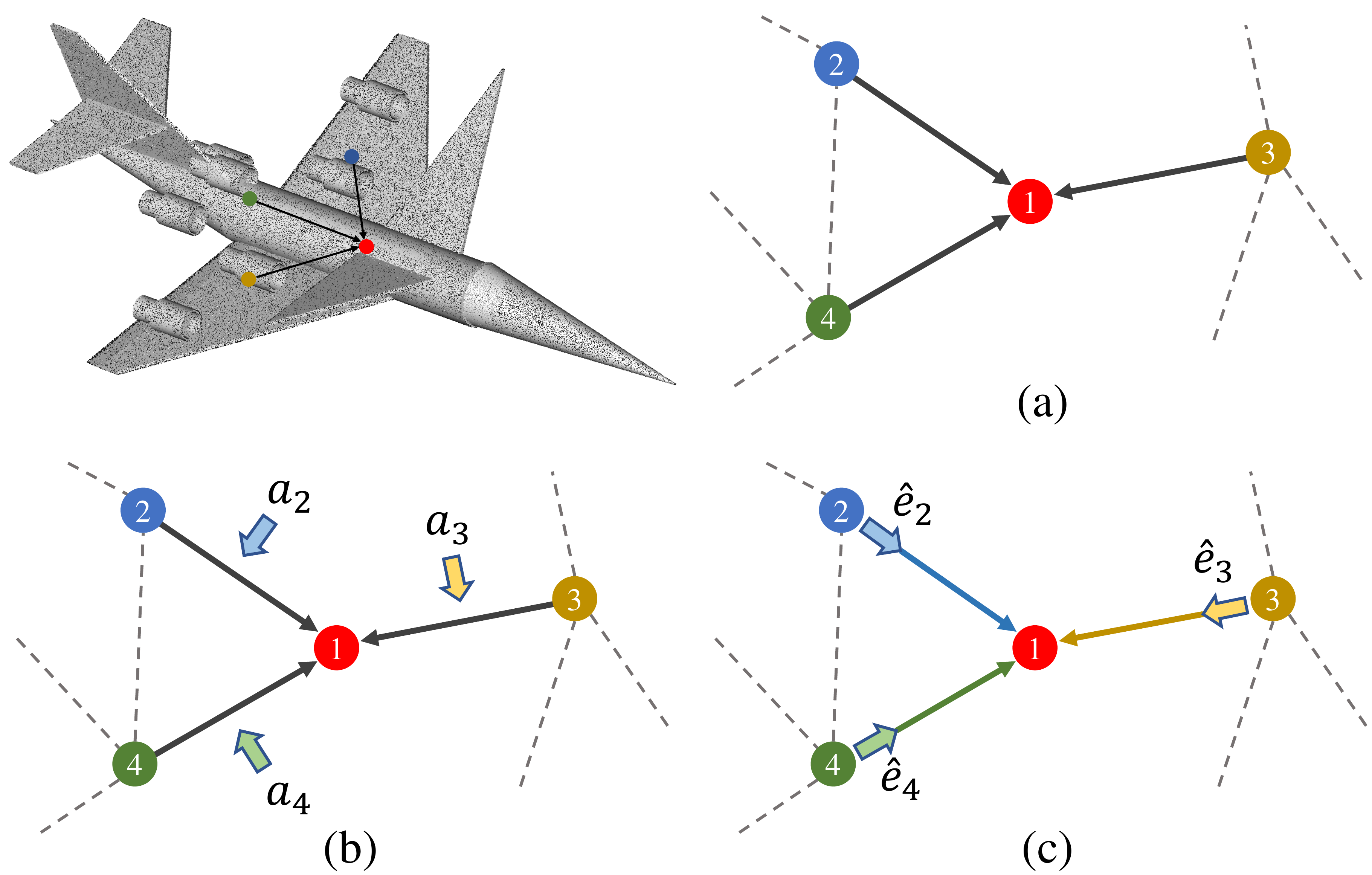}
	\caption{Illustration of adaptive kernels and fixed kernels in the convolution. (a) The standard graph convolution applies a fixed/isotropic kernel (black arrow) to compute features for each point indistinguishably. (b) Based on these features in (a), several attentional weights $a_i$ are assigned to determine their importance. (c) Differently, AGConv generates an adaptive kernel $\hat{e}_i$ that is unique to the learned features of each point.}
	\label{fig:intro}
\end{figure}

More recently, promising graph-like structures are explored for point cloud analysis. 
Graph CNNs (GCNs) \cite{wang2019dynamic,lin2020convolution,wang2019graph,fujiwara2020neural} describe a point cloud as graph data according to the spatial/feature similarity between points and generalize 2D convolutions on images to 3D data. 
GCN-based methods
have shown a powerful ability to understand contextual features and achieved much higher processing accuracy (e.g., point cloud segmentation) than  pointwise feature-based methods.
In order to process an unordered set of points with varying neighborhood sizes, standard graph convolutions harness shared weight functions over each pair of points to extract the corresponding edge feature. 
This leads to a fixed/isotropic convolution kernel, which is applied identically to all point pairs while neglecting their different feature correspondences. 
Intuitively, for points from different semantic parts of a same point cloud (see the neighboring points in Fig.~\ref{fig:intro}), the convolution kernel should be able to distinguish them and determine their individual contributions. 

To address the aforementioned shortcoming, several approaches \cite{wang2019graph,velivckovic2017graph} are proposed inspired by the idea of attention mechanism \cite{bahdanau2014neural,gehring2016convolutional}. 
As shown in Fig.~\ref{fig:intro} (b), proper attentional weights $a_i$ corresponding to the neighboring points are assigned, trying to identify their different importance when performing the convolution.
%
However, these methods are, in principle, still based on the fixed kernel convolution, as the attentional weights are just applied to the features obtained similarly (see the black arrows in Fig.~\ref{fig:intro} (b)). 
In this regard, attentional convolutions cannot solve the inherent limitations of current graph convolutions, making it still difficult to capture the delicate geometric features of a point by considering its structural connections to its neighboring points distinctively rather than uniformly. 
Considering the intrinsic isotropy of current graph convolutions, these attempts are still limited for detecting the most relevant part of the neighborhood.

In this paper, we propose a novel graph convolution operator to more thoroughly address the inherent yet long-standing limitation of traditional GCNs, in order to more effectively capture the geometric features of a point by more precisely harnessing its geometric correlations with its neighboring points; we call the operator Adaptive Graph Convolution (AGConv).
In the proposed AGConv, we adaptively establish the relationship between a pair of points according to their feature attributes instead of using fixed kernels; \textit{to our knowledge, this is the first time}. 
Such adaptiveness represents the diversity of kernels applied to each pair of points deriving from their individual features, which are capable of more accurately reflecting the underlying geometric characteristics of the objects when compared with the uniform kernels. 
Furthermore, we explore several design choices for feature convolution, offering more flexibility to the implementation of AGConv.
%
AGConv can be easily integrated into existing GCNs for point cloud analysis by simply replacing existing isotropic kernels with the adaptive kernels $\hat{e}_i$ generated from AGConv, as shown in Fig.~\ref{fig:intro} (c). 
%
%
%

This paper is extended from our previous work \cite{iccv/Zhouhaoran}. The contents and key
features newly added from \cite{iccv/Zhouhaoran}  are listed as:

1) The generalization-and-flexibility of AGConv is validated across various challenging yet important tasks of point cloud analysis, including the low-level ones, i.e.,
completion, denoising, upsampling and registration, and the high-level ones, i.e., classification, segmentation and circle extraction. Also, to verify the practicability of AGConv, more large-scale yet really captured point clouds which possess complex structures are involved.

2) To validate AGConv's effectiveness for point cloud completion, we improve ECG-Net \cite{9093117} by replacing its original graph convolution with AGConv. The improved version of ECG-Net is called iECG-Net. iECG-Net employs the so-called coarse-to-fine strategy, i.e., first recovering its global yet coarse shape and then increasing its local details to output the missing point cloud of input. The difference from the original graph convolution is that our AGConv not only extracts adequate spatial structure information but also extracts local features more efficiently and precisely. Therefore, the final completion results can better represent the missing parts on the tested benchmarks.

3) To validate AGConv's effectiveness for point cloud denoising, we improve Pointfilter \cite{9207844} by replacing its original encoder with AGConv. The improved version of Pointfilter is called iPointfilter. iPointfilter is an encoder-decoder network, which directly takes the raw neighboring points of each noisy point as input, and regresses a displacement vector to encourage this noisy point back to its ground-truth position. The difference from the original encoder is that AGConv better obtains a compact representation for each input patch. Experiments show that iPointfilter outperforms the state-of-the-art deep learning techniques in terms of noise-robustness and sharp feature preservation.

4) To validate AGConv's effectiveness for point cloud upsampling, we improve PU-Transformer \cite{2021PU} by adding the AGConv module in the upsampling head. The improved version of PU-Transformer is called iPU-Transformer. iPU-Transformer takes a sparse point cloud as input, and generates a dense point cloud. The difference from the original upsampling head is that AGConv better captures potential detailed features from the sparse point cloud. Therefore, iPU-Transformer can achieve a better upsampling effect in the region with detailed features.

5) To validate AGConv's effectiveness for point cloud circle extraction, we improve Circle-Net \cite{Honghua2022} by replacing its original graph convolution module with AGConv. The improved version of Circle-Net is called iCircle-Net. 
Most of existing approaches leverage classification and fitting operations independently, not synergizing with each other to accurately extract geometric primitives. Differently, iCircle-Net is an end-to-end classification-and-fitting network, in which the two operations of classifying circle-boundary points and fitting the circle can synergize with each other to improve the performance of circle extraction. The difference from the original graph convolution is that, our AGConv better perceives the circle spatial structure information. Therefore, iCircle-Net can achieve a better precision of circle boundary detection.

6) To validate AGConv's effectiveness for point cloud registration, we improve RGM \cite{fu2021robust} by utilizing AGConv in the local feature extractor instead of the original graph convolution. We denote the improved RGM with AGConv as iRGM. iRGM first utilizes a local feature extractor to obtain point-wise features, then both the edge generator and graph feature extractor are leveraged to excavate graph features between the source and target point clouds. In addition, the AIS module predicts the soft correspondence matrix and the LAP solver converts soft correspondences to hard correspondences. Finally, the transformation is solved by SVD. Compared with the original version, AGConv extracts more discriminative and robust features for each point, thus boosting the performance of registration.

Extensive experiments demonstrate the effectiveness and generalization of our AGConv. It achieves state-of-the-art  performances in the tasks of classification, segmentation, denoising, completion, upsampling, circle extraction and registration on many benchmark datasets.

\label{sec:introduction}

\section{Related work}
Although achieving tremendous success in 2D grid-like structures, deep learning is still not well explored for 3D point cloud analysis. We will review previous research categorized as conversion-based, point-based and graph-based methods, followed by the introduction of dynamic convolutions.

\vspace{5pt}\noindent\textbf{Conversion-based methods.} Conversion-based methods convert a point cloud into regular representations by voxelization (3D voxels), multi-view projection (2D grids), or hand-crafted feature descriptors. These regular representations can be easily fed into the powerful CNN/Transformer architectures.
For example, 1) multi-view methods usually extract and fuse view-wise features by projecting a 3D shape into multiple views, such as MVCNN \cite{su2015multi}, GVCNN \cite{feng2018gvcnn}, MHBN \cite{yu2018multi}, RN \cite{yang2019learning}, and View-GCN \cite{wei2020view}.
MVCNN \cite{su2015multi} extracts multi-view features based on 2D image classification networks and aggregates them by max-pooling to obtain a compact shape descriptor. Kalogerakis et al. \cite{kalogerakis20173d} present a surface-based projection layer that aggregates FCN outputs across multiple views and a surface-based CRF to favor coherent shape segmentation. GVCNN \cite{feng2018gvcnn} groups multi-view features and designs feature pooling on view groups. MHBN \cite{yu2018multi} presents a compact global descriptor by harmonizing bilinear pooling to integrate local convolutional features. RN \cite{yang2019learning} further models relations over a group of views and integrates them into a shape descriptor using relation scores. Unlike previous methods, View-GCN \cite{wei2020view} constructs a directed graph from multiple views, and uses graph convolution on hierarchical view-graphs to learn a global shape descriptor.
However, it is fundamentally difficult to apply these methods to large-scale scanned data, considering the struggle of covering the entire scene from single-point perspectives. 2) Voxelization-based methods usually voxelize a point cloud into 3D voxels which are easily fed into 3D CNNs, such as VoxNet \cite{maturana2015voxnet} and 3D ShapeNets \cite{wu20153d}.
Although encouraging performance has been achieved, they inevitably suffer from the loss of geometry information, as well as extensive computational costs. It is worth to noting that, some efficient data structures are developed to alleviate the computational cost, such as OctNet \cite{riegler2017octnet}, O-CNN\cite{wang2017cnn}, Kd-Net \cite{klokov2017escape}, PointGrid \cite{le2018pointgrid}.
OctNet \cite{riegler2017octnet} first uses a hybrid grid-octree structure to hierarchically partition a point cloud, and significantly reduces the memory and runtime required for high-resolution point clouds. Octree-based CNN \cite{wang2017cnn} feeds the average normal vectors of a 3D model sampled in the finest leaf octants into the network, and applies 3D-CNN on the octants occupied by the 3D shape surface. Kd-Net \cite{klokov2017escape} attempts to exploit more efficient data structures and skips the computations on empty voxels. PointGrid \cite{le2018pointgrid} integrates the point and grid representation by sampling a constant number of points within each embedding volumetric grid cell to efficiently extract geometric details by using 3D convolutions. Ben-Shabat et al. \cite{ben20173d} represent the point cloud by 3D grids and 3D modified Fisher Vectors, and employ a conventional CNN to learn the global representation.
3) The hand-crafted feature descriptors often leverage geometry operations (e.g., bilateral normal filtering) \cite{tog/WangLT16} or geometry priors (e.g., non-local similarity) \cite{cad/LiZFXWWH20} to construct regular representations. As known, these hand-crafted descriptors cannot fully utilize deep neural networks to automatically learn and extract geometry features. Hence, the valuable information provided in the training data may not be fully exhausted.


\vspace{5pt}\noindent\textbf{Point-based methods.} To handle the irregularity of point clouds, state-of-the-art deep networks are designed to directly manipulate raw point clouds, instead of introducing various intermediate representations. PointNet \cite{QiSMG17} proposes to use MLPs independently on each point and aggregate global features through a symmetric function. Thanks to this design, PointNet is invariant to input point orders, but fails to encode local geometric information, which is important for many geometric tasks. To solve this issue, PointNet++ \cite{qi2017pointnet++} proposes to apply PointNet layers locally in a hierarchical architecture to capture the regional information. Alternatively, Huang et al. \cite{huang2018recurrent} sort unordered 3D points into an ordered list and employ Recurrent Neural Networks (RNN) to extract features according to different dimensions. In order to process a set of points that are unordered and discrete, there also exist efforts of sorting the 3D points into an ordered list. Klokov et al. \cite{klokov2017escape}, Gadelha et al. \cite{gadelha2018multiresolution} propose to apply the Kd-tree structure to build a 1D list for points according to their coordinates. Although alleviating the unstructured problem, the sorting process is critical to the weight functions, and local geometric information may not be easily preserved in a specific ordered list. 

More recently, various approaches have been proposed for effective local feature learning. PointCNN \cite{li2018pointcnn} aligns points in a certain order by predicting a transformation matrix for the local point set. It inevitably leads to sensitivity in point orders, since the operation is not permutation-invariant. SpiderCNN \cite{xu2018spidercnn} defines its convolution kernel as a family of polynomial functions, relying on the neighbors' order. PCNN \cite{atzmon2018point} designs point kernels based on the spatial coordinates, and further KPConv \cite{thomas2019kpconv} presents a scalable convolution using explicit kernel points.
RS-CNN \cite{liu2019relation} assigns channel-wise weights to neighboring point features according to the geometric relations learned from 10-D vectors. ShellNet \cite{zhang2019shellnet} splits a local point set into several shell areas, from which features are extracted and aggregated.
Zhao et al. \cite{zhao2021point} and Guo et al. \cite{cvm/GuoCLMMH21} independently utilize Transformer 
to build dense self-attentions between the local and global features. Goyal et al.\cite{goyal2021revisiting} control auxiliary factors orthogonal to the network architecture to make PointNet++ competitive with more recent methods. Xu et al. \cite{xu2021learning } apply the attention mechanism to explore the relations between different variation components. PAConv \cite{xu2021paconv} constructs the convolution kernel by dynamically assembling basic weight matrices stored in a weight bank. Without modifying network configurations, PAConv is seamlessly integrated into classical MLP-based pipelines. CurveNet \cite{xiang2021walk} uses a feature aggregation paradigm for point cloud shape analysis, which propagates curve features by a curve grouping operator along with a curve aggregation operator. Qiu et al. \cite{qiu2021geometric} design a back-projection CNN module leveraging  error-correcting feedback structures to learn local features of point clouds. PointMLP \cite{ma2021rethinking} notes that a sophisticated local geometric extractor may not be crucial for performance and only uses residual feed-forward MLPs, without any delicate local feature exploration.

\begin{figure*}
	\centering
	\includegraphics[width=0.95\linewidth]{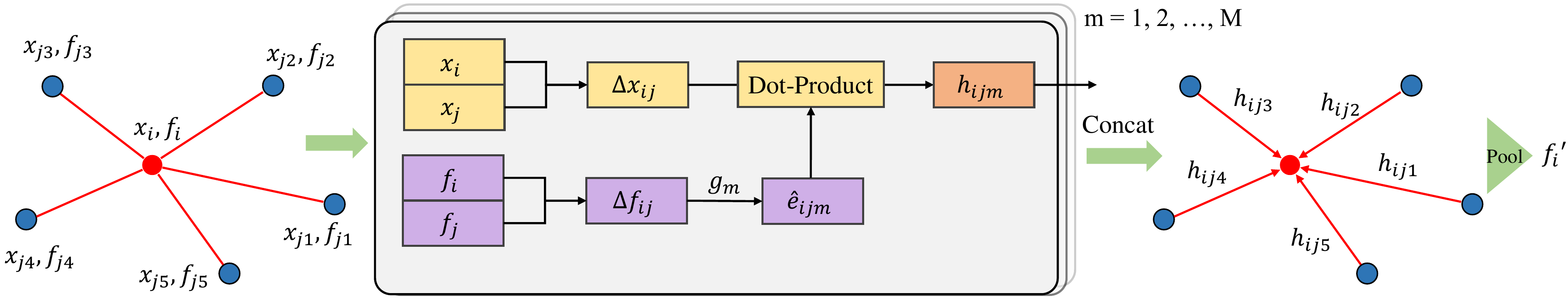}
	\caption{The illustration of AGConv processed in the neighborhood of a target point $x_i$. An adaptive kernel $\hat{e}_{ijm}$ is generated from the feature input $\Delta f_{ij}$ of a pair of points on the edge, which is then convolved with the corresponding spatial input $\Delta x_{ij}$. Concatenating $h_{ijm}$ of all dimensions yields the edge feature $h_{ij}$. Finally, the output feature $f_i'$ of the central point is obtained through a pooling function. AGConv differs from the other graph convolutions in that the convolution kernel is unique for each pair of points.}
	\label{fig:kernel}
\end{figure*}

\vspace{5pt}\noindent\textbf{Graph-based methods}. The graph-based methods treat points as nodes of a graph and establish edges according to their spatial/feature relationships. Graph is a natural representation of a point cloud to model local geometric structures. 
The notion of Graph Convolutional Network is proposed by Kipf et al. \cite{kipf2016semi}. It generalizes convolution operations over graphs by averaging features of adjacent nodes. Similar ideas \cite{shen2018mining,wang2019dynamic,hua2018pointwise,li2018pointcnn,lei2020spherical} are explored to extract local geometric features from local points. For example, 
Shen et al. \cite{shen2018mining} define kernels according to Euclidean distances and geometric affinities in the neighboring points. DGCNN \cite{wang2019dynamic} gathers the nearest neighboring points in the feature space, followed by the EdgeConv operators for feature extraction, in order to identify semantic cues dynamically. MoNet \cite{monti2017geometric} defines convolution as Gaussian mixture models in a local pseudo-coordinate system. Inspired by the attention mechanism, many efforts  \cite{velivckovic2017graph, wang2019graph, verma2018feastnet} are made to assign proper attentional weights to different points/filters. 3D-GCN \cite{lin2020convolution} develops deformable kernels, focusing on shift and scale-invariant properties in point cloud analysis.
diffConv \cite{corr/abs-2111-14658} operates on spatially-varying and density-dilated neighborhoods, which are further adapted by a learned masked attention mechanism. 

In order to expand the receptive field of graph convolutions, they either use graph pooling \cite{thomas2019kpconv,lin2020convolution} to gradually reduce the point numbers or a dynamic graph mechanism \cite{wang2019dynamic,pistilli2020learning} to connect similar points in the feature space. Both methods will change the graph construction structure which is easier to propagate local features throughout the point cloud. Therefore, the neighborhood of each point is varying between layers, determined dynamically according to the sampling strategy or feature similarity. That is, the relationships (edge features) are largely diverse among points not only in the neighborhood of a central point but also between any pair of points in the point cloud. Previous methods try to use a static function over each pair of points, neglecting their different feature correspondence from the previous layers. Differently, we propose to adaptively establish this varying relationship between a pair of points according to their feature attributes. This adaptiveness represents the diversity of  weight kernels applied on each pair of points deriving from their individual features.



\vspace{5pt}\noindent\textbf{Dynamic convolutions.} 
To handle the irregular and unordered point clouds, many efforts are made to adapt the traditional convolutions with dynamic strategies. For example, by building multiple network branches as experts, Ma et al. \cite{ma2018modeling} adopt real-valued weights to dynamically rescale the representations obtained from different experts; all the branches need to be executed, and thus the computation cannot be reduced at the test time. For the shape-adaptive convolution kernel, Gao et al. \cite{gao2019deformable} sample weights in the kernel space to reshape the convolutional kernels and achieve dynamic reception of fields. But the irregular memory access and computation pattern require customized CUDA kernels for the implementation. To improve the basic designs using fixed MLPs in PointNet/PointNet++, a variety of works \cite{velivckovic2017graph, wang2019graph, verma2018feastnet,thomas2019kpconv,liu2019relation} introduce weights based on the learned features, with more variants of convolution inputs \cite{wang2019dynamic,monti2017geometric,xu2018spidercnn}. Several methods boost the model capacity and improve the representation power by applying feature-conditioned attention weights on an ensemble of convolutional kernels, with a minor increase in computation \cite{chen2020dynamic, su2019pixel, yang2019condconv}. Shan et al. \cite{shan2020meta} adapt the weights of each input sample based on the similarity to the neighbors. 

Compared to the modification of model parameters, kernel weight prediction is more straightforward. It directly generates input-adaptive kernels at the test time.
Some efforts \cite{simonovsky2017dynamic,wu2019pointconv,jia2016dynamic} attempt to learn a dynamic weight for the convolution. PointConv \cite{wu2019pointconv} designs weight and density functions to fully approximate the 3D continuous convolution.
WeightNet \cite{ma2020weightnet} predicts the convolutional weights via simple grouped FC layers. Li et al. \cite{li2021involution} present Involution Kernel by reversing the design principles of convolution and generalizing the formulation of self-attention. Bello \cite{bello2021lambdanetworks}  predicts the weights of linear projections based on the contexts and the relative position embedding, which is advantageous in terms of computational cost and memory footprint.
However, this wisdom is to approximate weight functions from the direct 3D coordinates while our AGConv exploits features to learn the kernels exhibiting more adaptiveness. In addition, their implementations are heavily memory-consuming when convolving with high-dimensional features.

In summary, the main focus of this work is to handle the isotropy of point cloud convolutions, by developing an adaptive kernel that is unique to each point in the convolution; to our knowledge, this is the first time.


\label{sec:related}

\begin{figure*}
	\centering
	\includegraphics[width=0.95\linewidth]{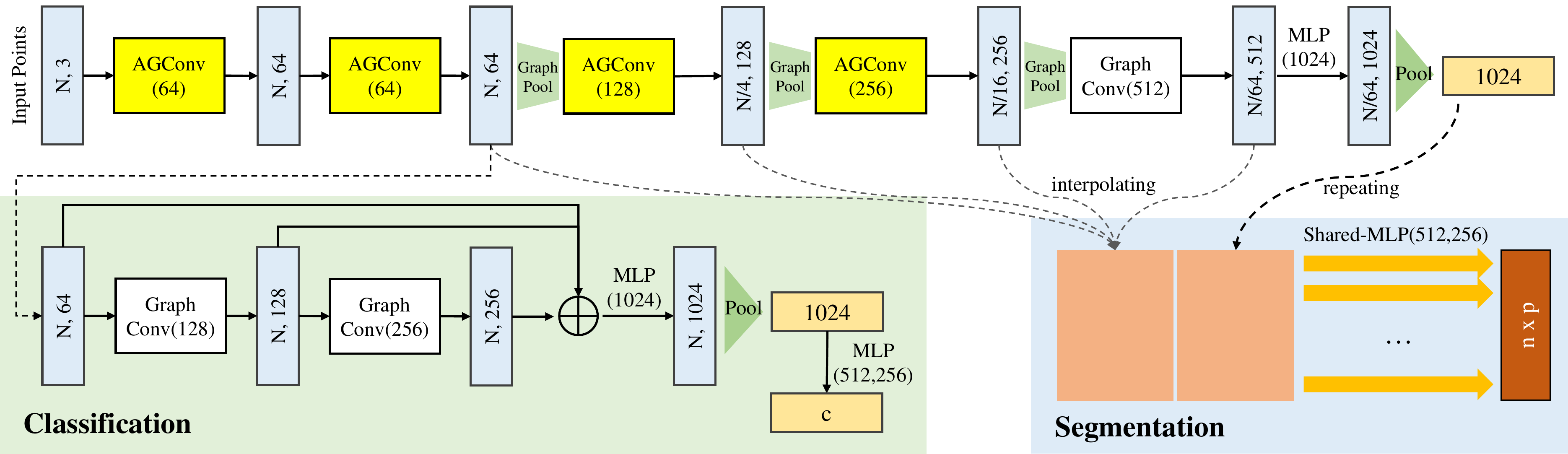}
	
	\caption{AGConv for classification and segmentation. GraphConv denotes our standard convolution without an adaptive kernel. The segmentation model uses pooling and interpolating to build a hierarchical graph structure, while the classification model applies a dynamic structure \cite{wang2019dynamic}. The subsampled features are resized to the same resolution as the input points by interpolation.}
	\label{fig:architecture}
\end{figure*}

\section{Methodology}
We exploit local geometric characteristics in point clouds by proposing a novel adaptive graph convolution (AGConv) in the spirit of graph neural networks (Sec.~\ref{sec:method:adapt}). 
The details of the constructed networks are shown in Sec.~\ref{sec:method:architecture}. 

\subsection{Adaptive graph convolution}
\label{sec:method:adapt}
%
We denote the input point cloud as $\mathcal{X} = \{x_i | i=1,2,...,N\} \in \mathbb{R}^{N \times 3}$ with the corresponding features defined as $\mathcal{F} = \{f_i | i=1,2,...,N\} \in \mathbb{R}^{N \times D}$. 
Here, $x_i$ processes the $(\mathbf{x},\mathbf{y},\mathbf{z})$ coordinates of the i-th point, and, in other cases, can be potentially combined with a vector of additional attributes, such as normal and color. 
%
We then compute a directed graph $\mathcal{G}(\mathcal{V},\mathcal{E})$ from the given point cloud where $\mathcal{V} = \{1,...,N\}$ is the set of points (nodes), and $\mathcal{E} \subseteq \mathcal{V} \times \mathcal{V}$ represents the set of edges. 
We construct the graph by employing the $k$-nearest neighbors (KNN) of each point including self-loop. 
Given the input $D$-dimensional features, our AGConv layer is designed to produce a new set of $M$-dimensional features with the same number of points while attempting to more accurately reflect local geometric characteristics than previous graph convolutions.

Denote that $x_i$ is the central point in the graph convolution, and $\mathcal{N}(i) = \{j | (i,j) \in \mathcal{E}\}$ is a set of point indices in its neighborhood. 
%
Due to the irregularity of point clouds, previous methods usually apply a fixed kernel function on all neighbors of $x_i$ to capture the geometric information of the patch.
However, different neighbors may reflect different feature correspondences with $x_i$, particularly when $x_i$ is located at salient regions, such as corners or edges.
In this regard, the fixed kernel may incapacitate the geometric representations generated from the graph convolution for classification and, particularly, segmentation.
%

In contrast, we endeavor to design an adaptive kernel to capture the distinctive relationships between each pair of points. 
To achieve this, for each channel in the output $M$-dimensional feature, our AGConv dynamically generates a kernel using a function over the point features $(f_i, f_j)$:
\begin{equation}
\hat{e}_{ijm} = g_m(\Delta f_{ij}), j \in \mathcal{N}(i).
\end{equation}
Here, $m = 1,2,...,M$ indicates one of the $M$ output dimensions corresponding to a single filter defined in our AGConv. 
In order to combine the global shape structure and feature differences captured in a local neighborhood \cite{wang2019dynamic}, we define $\Delta f_{ij} = [f_i, f_j-f_i]$ as the input feature for the adaptive kernel, where $[\cdot,\cdot]$ is the concatenation operation. 
$g(\cdot)$ is a feature mapping function, and here we use a multilayer perceptron. 

Like the computations in 2D convolutions, which obtain one of the $M$ output dimensions by convolving the $D$ input channels with the corresponding filter weights, our adaptive kernel is convolved with the corresponding points $(x_i, x_j)$:
\begin{equation}
h_{ijm} = \sigma \left \langle \hat{e}_{ijm}, \Delta x_{ij} \right \rangle, \label{equ:convolution}
\end{equation}
where $\Delta x_{ij}$ is defined as $[x_i, x_j-x_i]$ similarly, $\langle \cdot, \cdot \rangle$ represents the inner product of two vectors outputting $h_{ijm} \in \mathbb{R}$ and $\sigma$ is a nonlinear activation function. As shown in Fig.~\ref{fig:kernel} (middle), the m-th adaptive kernel $\hat{e}_{ijm}$ is combined with the spatial relations $\Delta x_{ij}$ of the corresponding point $x_j \in \mathbb{R}^3$, which means the size of the kernel should be matched in the dot product, i.e., the aforementioned feature mapping is $g_m: \mathbb{R}^{2D} \rightarrow \mathbb{R}^{6}$. In this way, the spatial positions in the input space are efficiently incorporated into each layer, combined with the feature correspondences extracted dynamically from our kernel. Stacking $h_{ijm}$ of each channel yields the edge feature $h_{ij} = [h_{ij1}, h_{ij2}, ..., h_{ijM}] \in \mathbb{R}^M$ between the connected points $(x_i, x_j)$.
Finally, we define the output feature of the central point $x_i$ by applying aggregation over all the edge features in the neighborhood (see Fig.~\ref{fig:kernel} (right)):
\begin{equation}
f_i' = \max_{j \in \mathcal{N}(i)} h_{ij},
\label{fifeature}
\end{equation}
where $\max$ is a channel-wise max-pooling function.
Overall, the convolution weights of AGConv are defined as $\Theta = (g_1, g_2, ..., g_M)$.

AGConv generates an adaptive kernel for each pair of points according to their individual features $(f_i, f_j)$. Then, the kernel $\hat{e}_{ijm}$ is applied to the point pair of $(x_i, x_j)$ in order to describe their spatial relations in the input space. The feature decision of $\Delta x_{ij}$ in the convolution of Eq.~\ref{equ:convolution} is an important design. In other cases, the inputs can be $x_i \in \mathbb{R}^E$ including additional dimensions representing other valuable point attributes, such as point normals and colors. 
By modifying the adaptive kernel to $g_m: \mathbb{R}^{2D} \rightarrow \mathbb{R}^{2E}$, our AGConv can also capture the relationships between feature dimensions and spatial coordinates which are from different domains. Note that, 
we use the spatial positions as input $x_i$ by default in the convolution.

\subsection{Network architectures for classification and segmentation}
\label{sec:method:architecture}
We design individual network architectures for the point cloud classification and segmentation tasks using the proposed AGConv layer. The network architectures are shown in Fig.~\ref{fig:architecture}. 
The standard graph convolution layer with a fixed kernel uses the same feature inputs $\Delta f_{ij}$ as in the adaptive kernels.

\vspace{5pt}\noindent\textbf{Dynamic graph update.} Following \cite{wang2019dynamic}, we update the graph structure in each layer according to the feature similarity among points, rather than fixed using spatial positions. That is, in each layer, the edge set $\mathcal{E}^{(l)}$ is recomputed where the neighborhood of point $x_i$ is $\mathcal{N}(i) = \{j_{i_1}, j_{i_2}, ..., j_{i_k}\}$ such that the corresponding features $f_{j_{i_1}}, f_{j_{i_2}}, ..., f_{j_{i_k}}$ are closest to $f_i$. This encourages the network to organize the graph semantically, grouping together similar points in the feature space but not solely considering their proximity in the spatial inputs. Thus, the receptive field of local points is expanded, leading to a propagation of local information throughout the point cloud. 
Note that, in the convolution with adaptive kernel in Eq.~\ref{equ:convolution}, $\Delta x_{ij}$ corresponds to the feature pair $(f_i, f_j)$ which may not be spatially close.

\begin{figure}[t]
	\centering
	
	\includegraphics[width=0.99\linewidth]{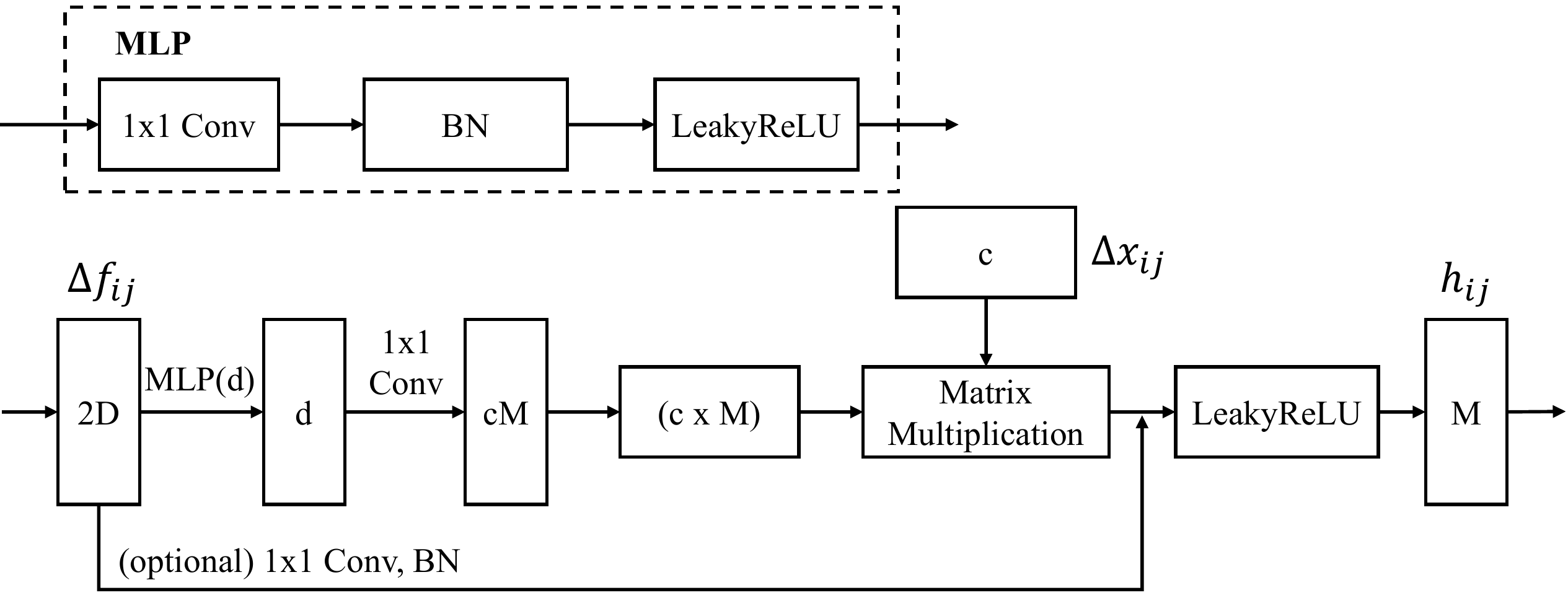}
	
	\caption{Kernel function in our adaptive convolution. We apply a two-layer MLP for the adaptive weight matrix. The output edge feature is obtained by matrix multiplication between $\Delta x_{ij}$ and the weight matrix. Optional ResNet block: shortcut $1 \times 1$ convolution and batch normalization layer.}
	\label{fig:kernel2}
\end{figure}

\vspace{5pt}\noindent\textbf{Kernel function for adaptive convolution.} In our experiments, the AGConv kernel function $g_m$ is implemented as a two-layer shared MLP with residual connections to extract important geometric information. It is an inevitable choice to use a shared mapping as the kernel function. However, $g_m$ is not the convolution kernel (fixed kernel) that is applied to points, but is to explore the different feature correspondences for different pairs of points. In the implementation, we process all $g_m$ ($m = 1,2,...,M$) together and obtain the adaptive kernels for the following convolution (see Fig.~\ref{fig:kernel2}). The first layer is one shared MLP($d$) for all $g_m$, and we organize the kernels as a weight matrix ($c \times M$) which is then applied to the corresponding $\Delta x_{ij}$ of dimension $c$ by matrix multiplication. After a LeakyReLU, the edge feature $h_{ij}$ is obtained and finally we apply Eq. \ref{fifeature} for the output feature of the central point. The ResNet connection is an optional block used in our segmentation model. From this perspective, the generation of adaptive kernels can be regarded as the generation of a weight matrix which directly produces the $M$-dimensional feature.

\vspace{5pt}\noindent\textbf{Graph pooling.} For the segmentation task, we reduce the number of points progressively in order to build the network in a hierarchical architecture. The point cloud is subsampled by the furthest point sampling algorithm \cite{QiSMG17} with a sampling rate of 4, and is applied by a pooling layer to output aggregated features on the coarsened graph. In each graph pooling layer, a new graph is constructed corresponding to the sampled points. The feature pooled at each point in the sub-cloud can be simply obtained by a max-pooling function within its neighborhood. Alternatively, we can use an AGConv layer to aggregate this pooled features.
To predict point-wise labels for the segmentation purpose, we interpolate deeper features from the subsampled cloud to the original points. Here, we use the nearest upsampling scheme to get the features for each layer, which are concatenated for the final point-wise features.

\vspace{5pt}\noindent\textbf{Segmentation network.} Our segmentation network architecture is illustrated in Fig.~\ref{fig:architecture}. The AGConv encoder includes 5 layers of convolutions in which the last one is a standard graph convolution layer, as well as several graph pooling layers that use an AGConv layer to aggregate features. The graph structures are updated in each layer according to the feature similarity among points. Besides, the subsampled features are resized to the same resolution as the input points by interpolation. The interpolation uses an inverse distance weighted average based on the $k$-nearest neighbors. The resized features are concatenated for the final point features to feed to the decoder. The final features of all points are regressed to the segmentation results by the shared MLP with hidden dimensions (512, 256).
In addition, our segmentation network includes a spatial transformer network \cite{jaderberg2015spatial} before the convolution layers. It processes the input points and outputs a $3 \times 3$ matrix in order to apply a global transformation. We apply standard graph convolutions (64, 128, 1024), followed by a max pooling function and fully-connected layers with hidden dimensions (512, 256). The output matrix is initialized as an identity matrix. Here, it is also possible to replace these graph convolutions with AGConv layers, but this does not lead to a significant improvement. Since the input contains normals as additional attributes, we apply the $3 \times 3$ matrix separately to the point and normal dimensions. The STN module can be seen as a global adaptive kernel that is convolved with all input points similar as in our AGConv. We report the results of networks with and without STN in Tab.~\ref{table:stn}.

\begin{table}[t]
	\centering
	\small
	\setlength{\tabcolsep}{3.5mm}
	\begin{tabular}{c|cc} 
		\toprule[1pt]
		Method & mcIoU(\%) & mIoU(\%) \\
		\midrule[0.3pt]
		\midrule[0.3pt]
		w/o STN							& 83.2 & 86.2 \\
		STN 						& \textbf{83.4} & \textbf{86.4} \\
		\bottomrule[1pt]
	\end{tabular}
	\vspace{5pt}
	\caption{Segmentation results on the ShapeNetPart dataset. The model using STN achieves better results. }
	\label{table:stn}
\end{table}

\vspace{5pt}\noindent\textbf{Classification network.} The classification network uses a similar encoder as in the segmentation model (see Fig.~\ref{fig:architecture}). For sparser point clouds used in ModelNet40, we simply apply dynamic graph structures \cite{wang2019dynamic} without pooling and interpolation. Specifically, the graph structure is updated in each layer according to the feature similarity among points, rather than fixed using spatial positions. That is, in each layer, the edge set $\mathcal{E}_l$ is recomputed where the neighborhood of point $x_i$ is $\mathcal{N}(i) = \{j_1, j_2, ..., j_k\}$ such that the corresponding features $f_{j_1}, f_{j_2}, ..., f_{j_k}$ are closest to $f_i$. This encourages the network to organize the graph semantically and expands the receptive field of local neighborhood by grouping together similar points in the feature space.

\label{sec:method}



\begin{table}[t]
	\centering
	\footnotesize
	\begin{tabular}{c|cccc} 
		\toprule[1pt]
		Method & Input & \#points & mAcc(\%) & OA(\%) \\
		\midrule[0.3pt]
		\midrule[0.3pt]
		3DShapeNetParts \cite{wu20153d}						& voxel & - & 77.3 & 84.7 \\
		VoxNet \cite{maturana2015voxnet}				& voxel & - & 83.0 & 85.9 \\
		Subvolume \cite{qi2016volumetric}				& voxel & - & 86.0 & 89.2 \\
		\midrule[0.3pt]
		PointNet \cite{QiSMG17}							& xyz & 1k & 86.0 & 89.2 \\
		PointNet++ \cite{qi2017pointnet++}				& xyz, normal & 5k & - & 91.9 \\
		Kd-Net \cite{klokov2017escape}					& xyz & 1k & - & 90.6 \\
		SpecGCN \cite{wang2018local}					& xyz & 1k & - & 92.1 \\
		SpiderCNN \cite{xu2018spidercnn}				& xyz, normal & 5k & - & 92.4 \\
		PointCNN \cite{li2018pointcnn}					& xyz & 1k & 88.1 & 92.2 \\
		SO-Net \cite{li2018so}							& xyz, normal & 5k & - & \textbf{93.4} \\
		DGCNN \cite{wang2019dynamic}					& xyz & 1k & 90.2 & 92.9 \\
		KPConv \cite{thomas2019kpconv}					& xyz & 6.8k & - & 92.9 \\
		3D-GCN \cite{lin2020convolution}				& xyz & 1k & - & 92.1 \\
		PointASNL \cite{yan2020pointasnl}				& xyz, normal & 1k & - & 93.2 \\
		\midrule[0.3pt]
		Ours											& xyz & 1k & \textbf{90.7} & \textbf{93.4} \\
		\bottomrule[1pt]
	\end{tabular}
	\vspace{5pt}
	\caption{Classification results on ModelNet40. Our network achieves the best results according to both mAcc and OA.}
	\label{table:cls_results}
\end{table}

\begin{table*}
	\centering
	\footnotesize
	\setlength{\tabcolsep}{1mm}
	\begin{tabular}{c|cc|cccccccccccccccc} 
		\toprule[1pt]
		Method & mcIoU & mIoU & air & bag & cap & car & chair & ear & guitar & knife & lamp & laptop & motor & mug & pistol & rocket & skate & table  \\
		& & & 					plane &  &  	&	  &	     & phone &      &		&		&	 	& bike & 		&		&		 & board &	\\
		\midrule[0.3pt]
		Kd-Net \cite{klokov2017escape} 	& 77.4 & 82.3 & 80.1 & 74.6 & 74.3 & 70.3 & 88.6 & 73.5 &90.2 & 87.2 & 81.0 & 94.9 & \textbf{87.4} & 86.7 & 78.1 & 51.8 & 69.9 & 80.3 \\
		PointNet \cite{QiSMG17}	& 80.4 & 83.7 & 83.4 & 78.7 & 82.5 & 74.9 & 89.6 & 73.0 & 91.5 & 85.9 & 80.8 & 95.3 & 65.2 & 93.0 & 81.2 & 57.9 & 72.8 & 80.6 \\
		PointNet++ \cite{qi2017pointnet++} & 81.9 & 85.1 & 82.4 & 79.0 & 87.7 & 77.3 & 90.8 & 71.8 & 91.0 & 85.9 & 83.7 & 95.3 & 71.6 & 94.1 & 81.3 & 58.7 & 76.4 & 82.6\\
		SO-Net \cite{li2018so} & 81.0 & 84.9 & 82.8 & 77.8 & 88.0 & 77.3 & 90.6 & 73.5 &90.7 & 83.9 & 82.8 & 94.8 & 69.1 & 94.2 & 80.9 & 53.1 & 72.9 & 83.0 \\
		DGCNN \cite{wang2019dynamic} & 82.3 & 85.2 & 84.0 & 83.4 & 86.7 & 77.8 & 90.6 & 74.7 & 91.2 & 87.5 & 82.8 & 95.7 & 66.3 & 94.9 & 81.1 &    63.5 & 74.5 & 82.6\\
		PointCNN \cite{li2018pointcnn} & - & 86.1 & 84.1 & \textbf{86.4} & 86.0 & 80.8 & 90.6 & 79.7 & 92.3 & 88.4 & \textbf{85.3} & 96.1 & 77.2 & 95.3 & 84.2 & 64.2 & 80.0 & 83.0 \\
		PointASNL \cite{yan2020pointasnl} & - & 86.1 & 84.1 & 84.7 & \textbf{87.9} & 79.7 & \textbf{92.2} & 73.7 & 91.0 & 87.2 & 84.2 & 95.8 & 74.4 & 95.2 & 81.0 & 63.0 & 76.3 & 83.2\\
		3D-GCN \cite{lin2020convolution} & 82.1 & 85.1 & 83.1 & 84.0 & 86.6 & 77.5 & 90.3 & 74.1 & 90.9 & 86.4 & 83.8 & 95.6 & 66.8 & 94.8 & 81.3 & 59.6 & 75.7 & 82.8 \\
		KPConv \cite{thomas2019kpconv} & \textbf{85.1} & \textbf{86.4} & 84.6 & 86.3 & 87.2 & \textbf{81.1} & 91.1 & 77.8 & \textbf{92.6} & 88.4 & 82.7 & \textbf{96.2} & 78.1 & \textbf{95.8} & \textbf{85.4} & \textbf{69.0} & \textbf{82.0} & 83.6\\
		\midrule[0.3pt]
		Ours  & 83.4 & \textbf{86.4} & \textbf{84.8} & 81.2 & 85.7 & 79.7 & 91.2 & \textbf{80.9} & 91.9 & \textbf{88.6} & 84.8 & \textbf{96.2} & 70.7 & 94.9 & 82.3 & 61.0 & 75.9 & \textbf{84.2} \\
		\bottomrule[1pt]
	\end{tabular}
	\vspace{5pt}
	\caption{Part segmentation results on ShapeNetPart evaluated by the mean class IoU (mcIoU) and mean instance IoU (mIoU).}
	\label{table:partseg_results}
\end{table*}

\section{Evaluation}
In this section, we evaluate our AGConv for point cloud classification, part segmentation and indoor/outdoor segmentation.

\subsection{Classification}
\vspace{5pt}\noindent\textbf{Data.} We evaluate our model on ModelNet40 \cite{wu20153d} for classification. It contains 12,311 meshed CAD models from 40 categories, where 9,843 models are used for training and 2,468 models for testing. We follow the experimental setting of \cite{QiSMG17}. 
We sample 1024 points for each object uniformly and only use the $(\mathbf{x}, \mathbf{y}, \mathbf{z})$ coordinates of sampled points as input.
Data augmentation includes shifting, scaling and perturbing of the points.

\vspace{5pt}\noindent\textbf{Network configuration.} The network architecture is shown in Fig.~\ref{fig:architecture}. Following \cite{wang2019dynamic}, we recompute the graph based on the feature similarity in each layer. The number $k$ of neighborhood size is set to 20 for all layers. Shortcut connections are included and one shared fully-connected layer (1024) is applied to aggregate the multi-scale features. The global feature is obtained using a max-pooling function. All layers are with LeakyReLU and batch normalization.
We use the SGD optimizer with the momentum set to 0.9. The initial learning rate is 0.1 and is dropped until 0.001 using cosine annealing \cite{loshchilov2016sgdr}. The batch size is set to 32 for all training models. We use PyTorch 
for implementation and train the network on a RTX 2080 Ti GPU. The hyperparameters are chosen in a similar way for other tasks.

\vspace{5pt}\noindent\textbf{Results.} We show the results for classification  in Tab.~\ref{table:cls_results}. The evaluation metrices on this dataset are the mean class accuracy (mAcc) and the overall accuracy (OA). Our model achieves the best scores on this dataset. For a clear comparison, we show the input data types and the number of points corresponding to each method. Our AGConv only considers the point coordinates as input with a relatively small size of 1k points, which already outperforms other methods using larger inputs.

\subsection{Part segmentation}
\vspace{5pt}\noindent\textbf{Data.} We further test our model for the part segmentation task on the ShapeNetPart dataset \cite{yi2016scalable}. This dataset contains 16,881 shapes from 16 categories, with 14,006 for training and 2,874 for testing. Each point is annotated with one label from 50 parts and each point cloud contains 2-6 parts. We follow the experimental setting of \cite{qi2017pointnet++} and use their provided data for the benchmarking purpose. 2,048 points are sampled from each shape. The input attributes include the point normals apart from the 3D coordinates. 

\vspace{5pt}\noindent\textbf{Network configuration.} Following \cite{QiSMG17}, we include a one-hot vector representing category types for each point. It is stacked with the point-wise features to compute the segmentation results. 
Other training parameters are set the same as in our classification task. We use spatial positions (without normals) as $\Delta x_{ij}$ as in Sec.~\ref{sec:method:adapt}. Other choices will be evaluated later in Sec.~\ref{sec:eval:ablation}.

\vspace{5pt}\noindent\textbf{Results.} We report the mean class IoU (mcIoU) and mean instance IoU (mIoU) in Tab.~\ref{table:partseg_results}. Following  \cite{QiSMG17}, IoU of a shape is computed by averaging IoU of each part. The mean IoU (mIoU) is computed by averaging the IoUs of all testing instances. The class IoU (mcIoU) is the mean IoU over all shape categories. We also show the class-wise segmentation results. Our model achieves the state-of-the-art performance compared with other methods.

\begin{table}[t]
	\centering
	\small
	\setlength{\tabcolsep}{3.5mm}
	\begin{tabular}{c|cc} 
		\toprule[1pt]
		Ablations & mcIoU(\%) & mIoU(\%) \\
		\midrule[0.3pt]
		\midrule[0.3pt]
		GraphConv							& 81.9 & 85.5 \\
		Attention Point						& 78.0 & 83.3 \\
		Attention Channel					& 77.9 & 83.0\\
		\midrule[0.3pt]
		Feature								& 82.2 & 85.9 \\
		Normal								& 83.2 & 86.2 \\
		Initial attributes					& 83.2 & 86.1 \\
		Ours								& \textbf{83.4} & \textbf{86.4} \\
		\bottomrule[1pt]
	\end{tabular}
	\vspace{5pt}
	\caption{Ablation studies on ShapeNetPart for part segmentation.}
	\label{table:ab:seg}
\end{table}

\begin{table}[t]
	\centering
	\small
	\setlength{\tabcolsep}{3.5mm}
	\begin{tabular}{c|cc} 
		\toprule[1pt]
		Method & mAcc(\%) & OA(\%) \\
		\midrule[0.3pt]
		\midrule[0.3pt]
		GraphConv								& 88.8 & 92.5 \\
		Attention Point							& 88.5 & 92.1 \\
		Attention Channel						& 89.2 & 92.2 \\
		Ours									& \textbf{90.7} & \textbf{93.4} \\
		\bottomrule[1pt]
	\end{tabular}
	\vspace{5pt}
	\caption{Results of ablation networks on ModelNet40.}
	\label{table:ablation}
\end{table}

\begin{table*}
	\centering
	\footnotesize
	\setlength{\tabcolsep}{1mm}
	\begin{tabular}{c|ccc|ccccccccccccc} 
		\toprule[1pt]
		Method & OA & mAcc & mIoU & ceiling & floor & wall & beam & column & window & door & table & chair & sofa & bookcase & board & clutter  \\
		\midrule[0.3pt]
		PointNet \cite{QiSMG17} & – & 49.0 & 41.1 & 88.8 & 97.3 & 69.8 & 0.1 & 3.9 & 46.3 & 10.8 & 59.0 & 52.6 & 5.9 & 40.3 & 26.4 & 33.2 \\
		SegCloud \cite{tchapmi2017segcloud} & – & 57.4 & 48.9& 90.1& 96.1& 69.9& 0.0& 18.4& 38.4& 23.1& 70.4& 75.9& 40.9& 58.4& 13.0& 41.6 \\
		PointCNN \cite{li2018pointcnn} &85.9& 63.9& 57.3& 92.3& 98.2& 79.4& 0.0& 17.6& 22.8& 62.1& 74.4& 80.6& 31.7& 66.7& 62.1& 56.7 \\
		PCCN \cite{wang2018deep} &–& 67.0& 58.3& 92.3& 96.2& 75.9& \textbf{0.3}& 6.0& \textbf{69.5}& 63.5& 66.9& 65.6& 47.3& 68.9& 59.1& 46.2 \\
		PointWeb \cite{zhao2019pointweb} & 87.0& 66.6& 60.3& 92.0& \textbf{98.5}& 79.4& 0.0& 21.1& 59.7& 34.8& 76.3& 88.3& 46.9& 69.3& 64.9& 52.5 \\
		HPEIN \cite{jiang2019hierarchical} & 87.2& 68.3& 61.9& 91.5& 98.2& 81.4& 0.0& 23.3& 65.3& 40.0& 75.5& 87.7& 58.5& 67.8& 65.6& 49.4 \\
		GAC \cite{wang2019graph} &87.7 &-& 62.8& 92.2& 98.2& 81.9& 0.0& 20.3& 59.0& 40.8& 78.5& 85.8& 61.7& 70.7& \textbf{74.6}& 52.8 \\
		KPConv \cite{thomas2019kpconv} &– &72.8& 67.1& 92.8& 97.3& \textbf{82.4}& 0.0& 23.9& 58.0& 69.0& 81.5& \textbf{91.0}& 75.4& \textbf{75.3}& 66.7& \textbf{58.9} \\
		PointASNL \cite{yan2020pointasnl} &87.7& 68.5& 62.6& \textbf{94.3}& 98.4& 79.1& 0.0& \textbf{26.7}& 55.2& 66.2& 83.3& 86.8& 47.6& 68.3& 56.4& 52.1 \\
		\midrule[0.3pt]
		Ours &\textbf{90.0}& \textbf{73.2}&\textbf{67.9} & 93.9& 98.4& 82.2&  0.0& 23.9& 59.1& \textbf{71.3}& \textbf{91.5}& 81.2& \textbf{75.5}& 74.9& 72.1& 58.6 \\ 
		\bottomrule[1pt]
	\end{tabular}
	\vspace{5pt}
	\caption{Semantic segmentation results on S3DIS evaluated on Area 5. We report the mean classwise IoU (mIoU), mean classwise accuracy (mAcc) and overall accuracy (OA). IoU of each class is also provided.}
	\label{table:scene_results}
\end{table*}

\begin{table*}
	\centering
	\footnotesize
	\setlength{\tabcolsep}{1mm}
	\begin{tabular}{c|c|cccccccccc} 
		\toprule[1pt]
		Method & mIoU & ground & building & pole & bollard & trash can & barrier & pedestrian & car & natural \\
		\midrule[0.3pt]
		RF\textunderscore MSSF \cite{thomas2018RFMSSF} & 61.5& 99.1& 90.5& 66.4& 62.6& 5.8& 52.1& 5.7& 86.2& 84.7 \\
		HDGCN \cite{liang2019hierarchical}& 68.3& 99.4& 93.0& 67.7& 75.7& 25.7& 44.7& 37.1& 81.9 & 89.6\\
		MS3\textunderscore DVS \cite{roynard2018classification}& 66.9& 99.0& 94.8& 52.4& 38.1& 36.0& 49.3& 52.6& 91.3& 88.6\\ 
		ConvPoint \cite{boulch2020convpoint}& 75.9 & \textbf{99.5}& 95.1& \textbf{71.6}&  \textbf{88.7}& 46.7& 52.9& 53.5& 89.4& 85.4\\
		KPConv \cite{thomas2019kpconv} & 75.9 & -& -& -&  -& -& -& -& -& -\\ 
		\midrule[0.3pt]
		Ours &\textbf{76.9}& 99.4& \textbf{97.3}& 67.8&  77.1& \textbf{49.4}& \textbf{59.4}& \textbf{55.5}& \textbf{93.2}& \textbf{93.2}\\ 
		\bottomrule[1pt]
	\end{tabular}
	\vspace{5pt}
	\caption{Semantic segmentation results on NPM3D. We report the mean classwise IoU (mIoU) and IoU of each class.}
	\label{table:outdoor_scene_results}
\end{table*}

\subsection{Indoor scene segmentation}
\label{sec:eval:indoor}
\vspace{5pt}\noindent\textbf{Data.} Our third experiment shows the semantic segmentation performance of our model on the S3DIS dataset \cite{armeni20163d}. This dataset contains 3D RGB point clouds from six indoor areas of three different buildings, covering a total of 271 rooms. Each point is annotated with one semantic label from 13 categories. For a common evaluation protocol \cite{tchapmi2017segcloud,QiSMG17,landrieu2018large}, we choose Area 5 as the test set which is not in the same building as other areas.

\vspace{5pt}\noindent\textbf{Real scene segmentation.} The large-scale indoor datasets reveal more challenges, covering larger scenes in a real-world environment with noise and outliers. Thus, we follow the experimental settings of KPConv \cite{thomas2019kpconv}, and train the network using randomly sampled clouds in spheres. The subclouds contain more points with varying sizes, and are stacked into batches for training. During testing, spheres are uniformly picked in the scenes, and each point is tested several times using a voting scheme. The input point attributes include the RGB colors and the original heights.

\begin{figure}[t]
	\newlength{\unitxx}
	\setlength{\unitxx}{0.27\linewidth}
	\centering
	
	\includegraphics[width=\unitxx]{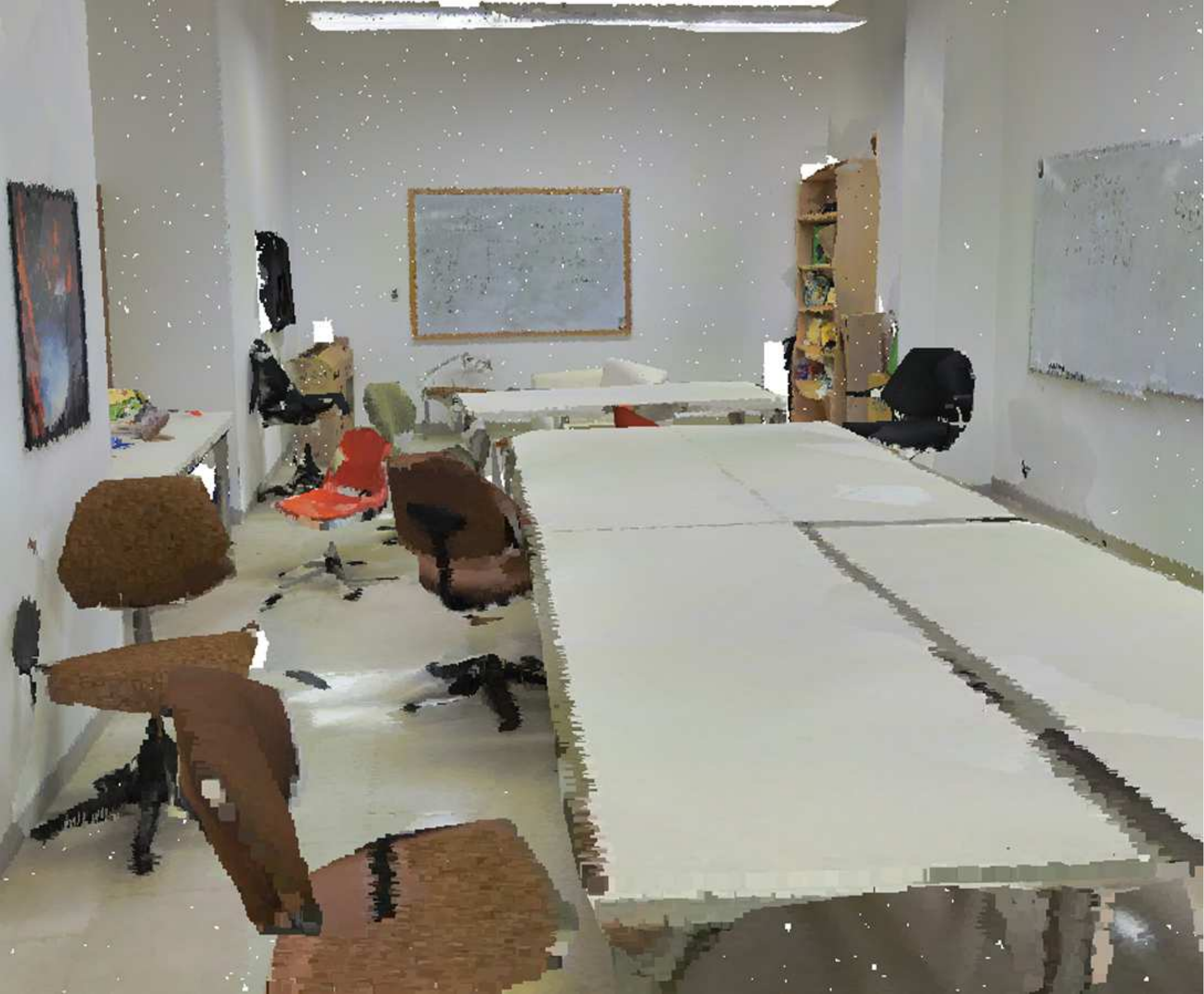}%
	\hspace{2pt}\includegraphics[width=\unitxx]{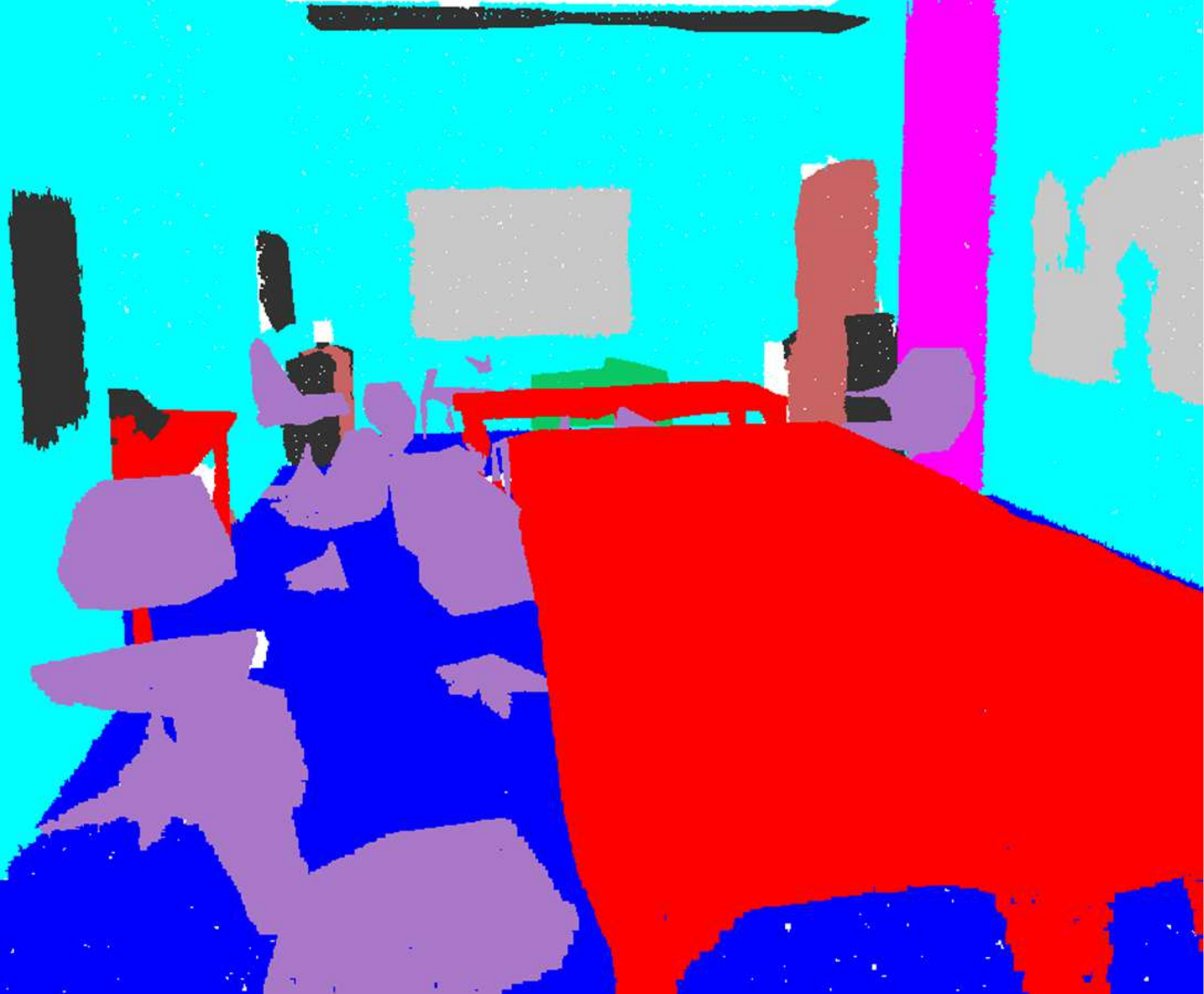}%
	\hspace{2pt}\includegraphics[width=\unitxx]{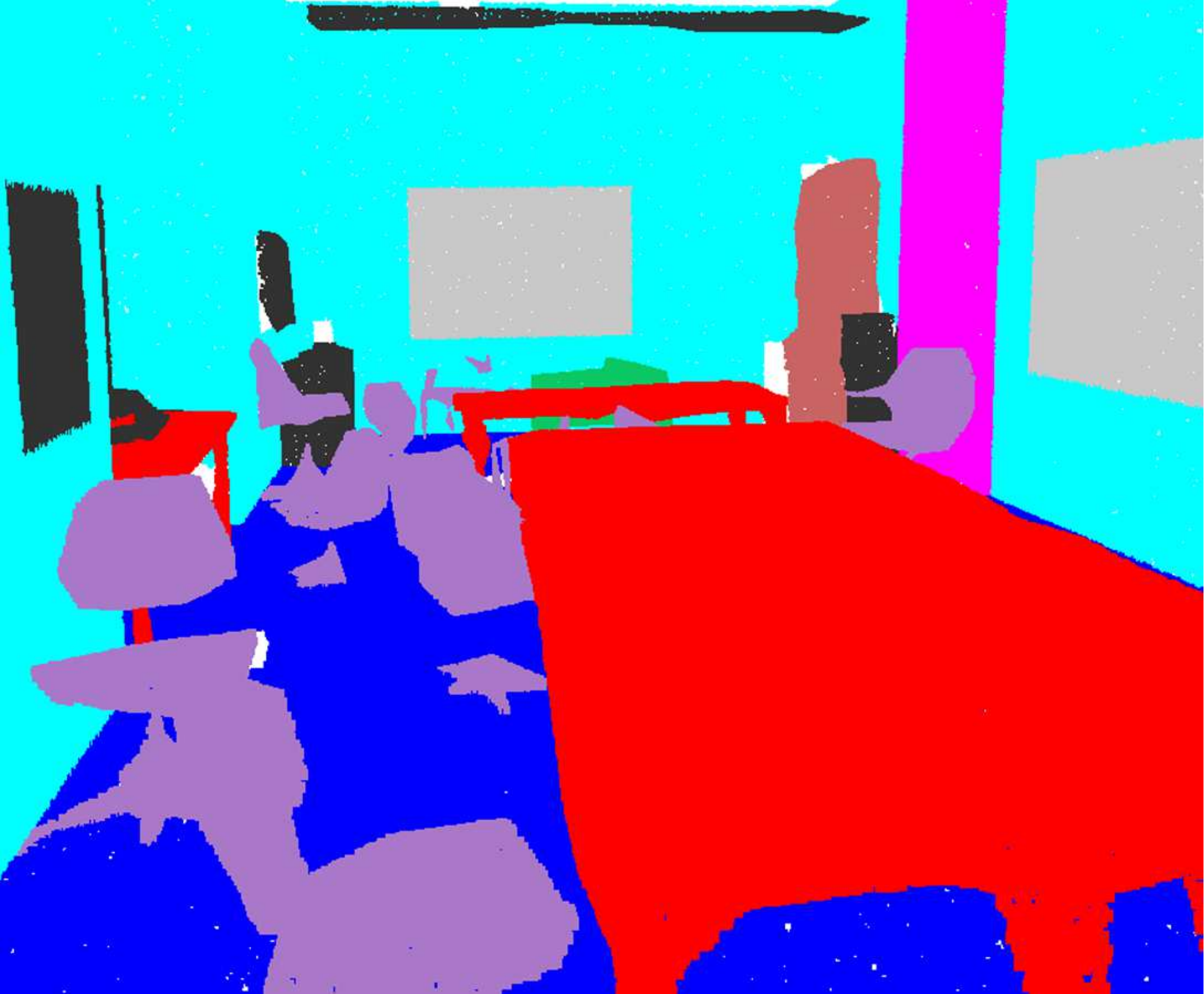}
	
	\vspace{3pt}
	
	\includegraphics[width=\unitxx]{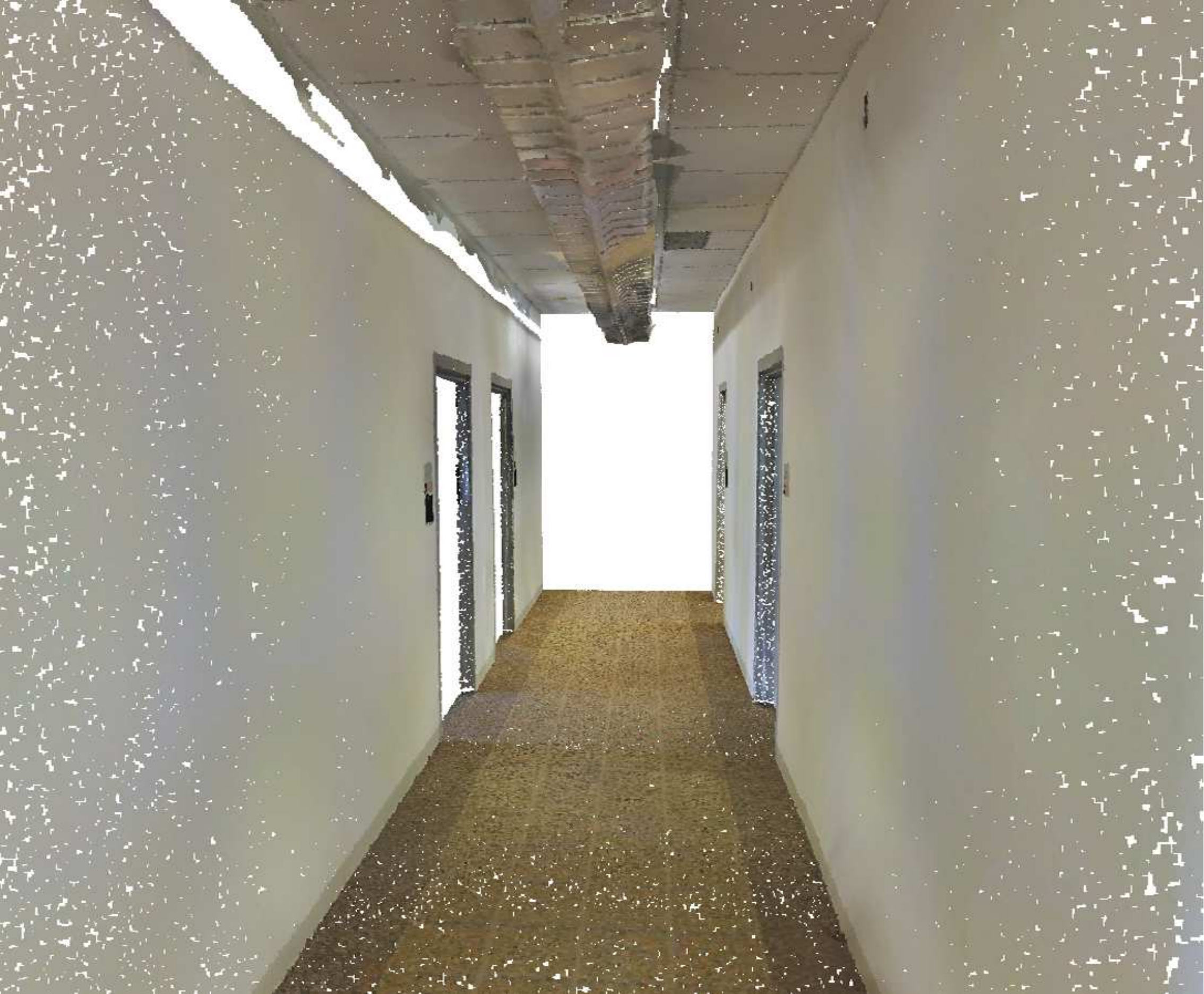}%
	\hspace{2pt}\includegraphics[width=\unitxx]{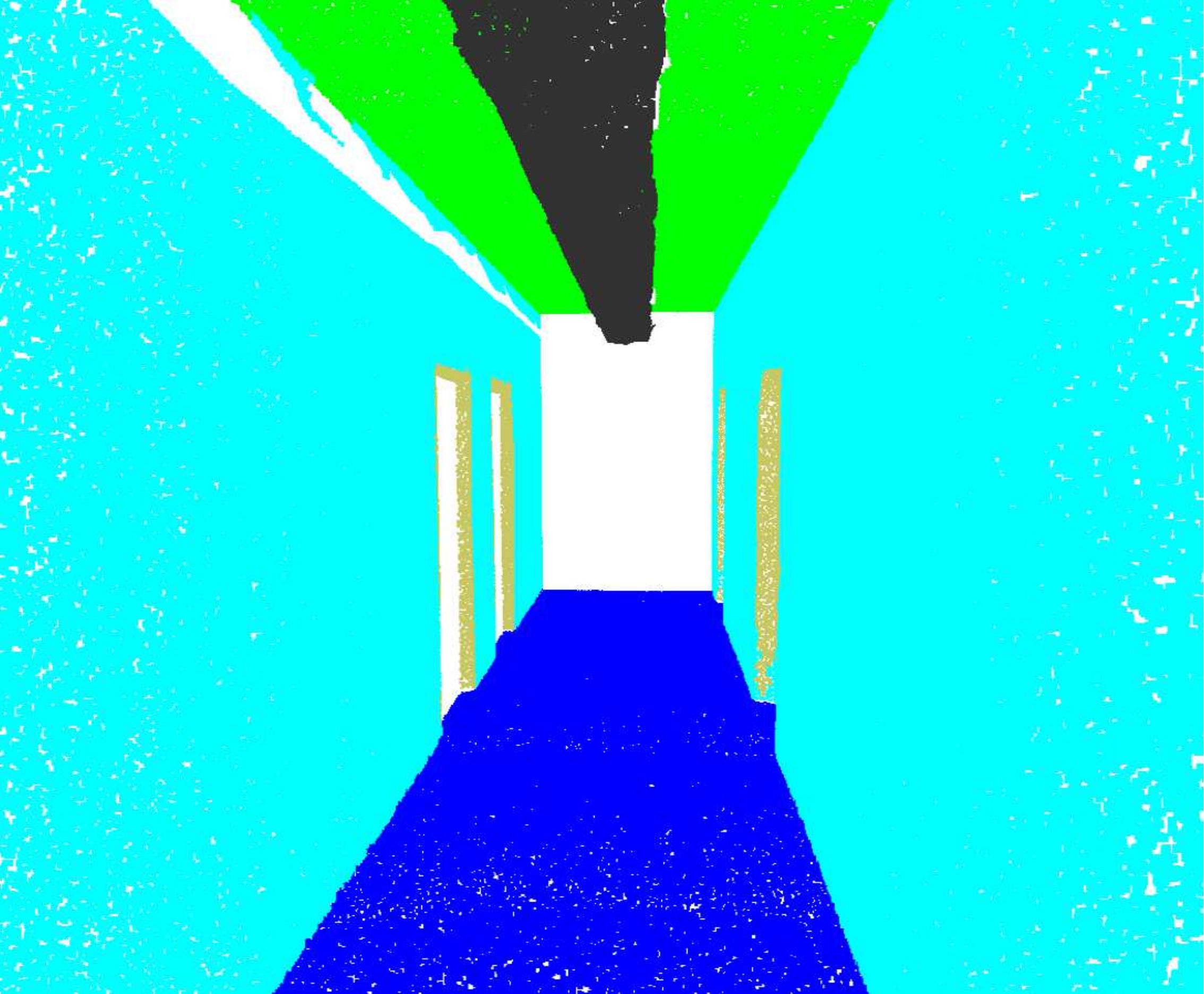}%
	\hspace{2pt}\includegraphics[width=\unitxx]{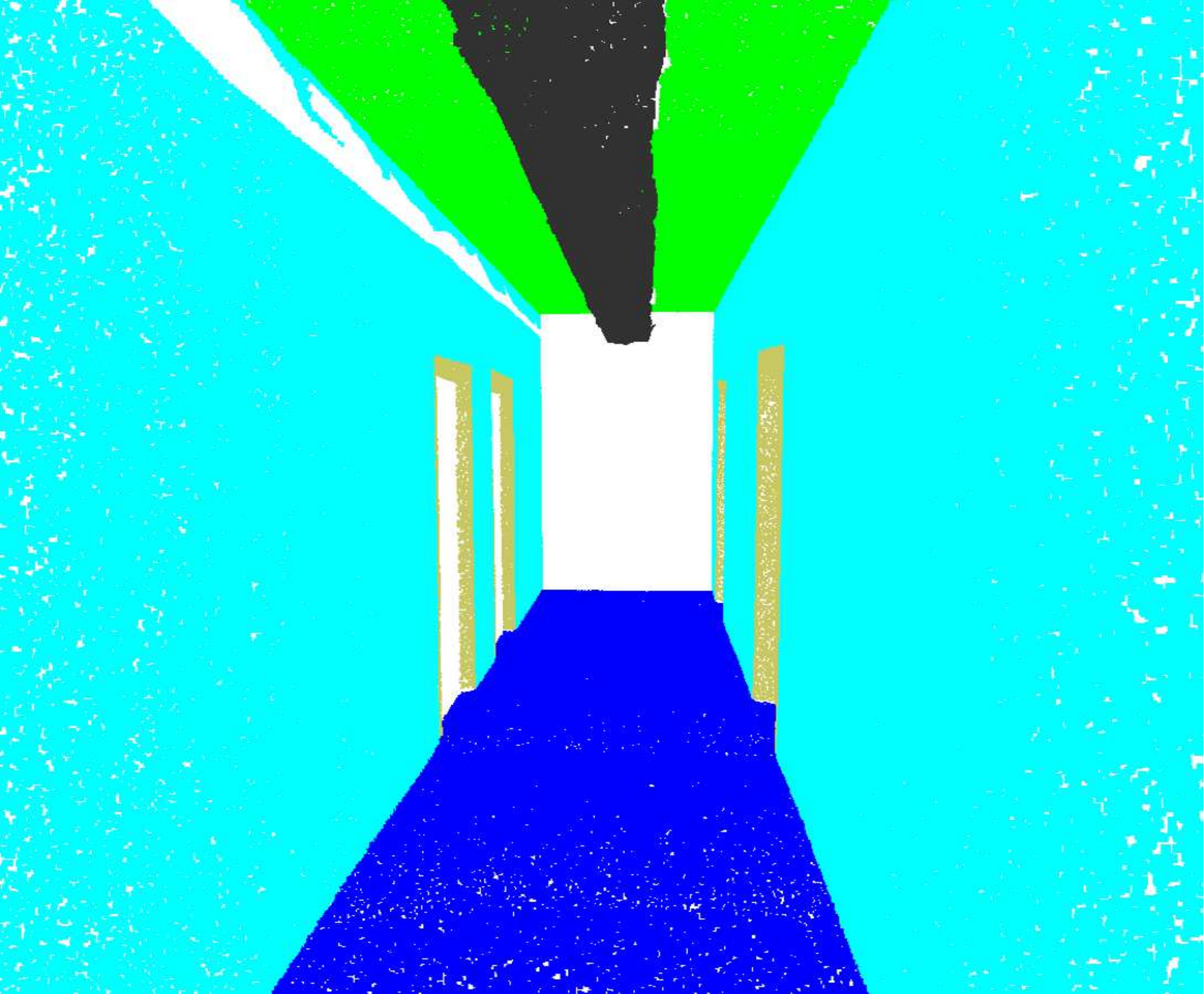}
	
	\vspace{-2pt}
	
	\subfigure[Input]{\includegraphics[width=\unitxx]{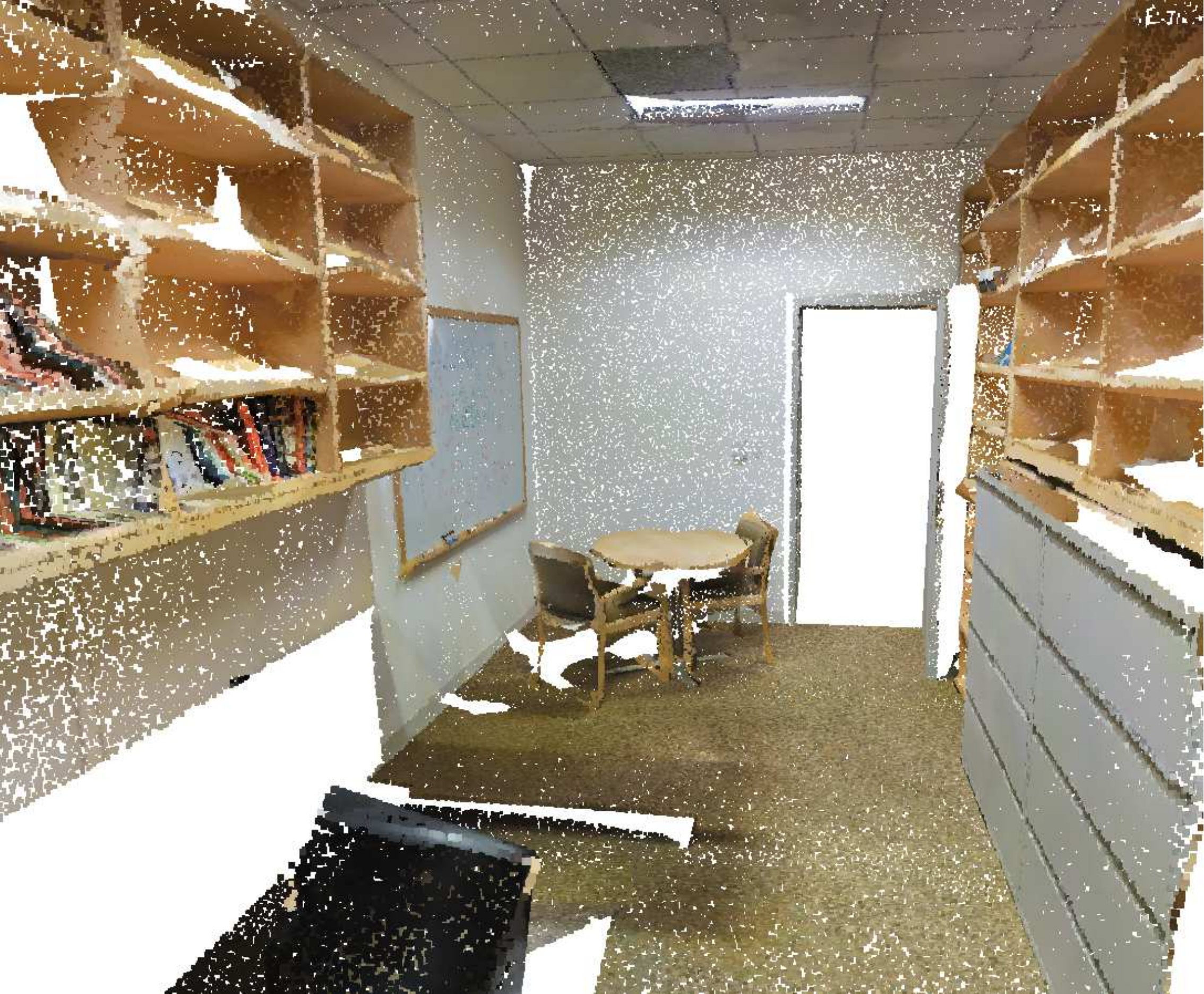}}%
	\hspace{2pt}\subfigure[Prediction]{\includegraphics[width=\unitxx]{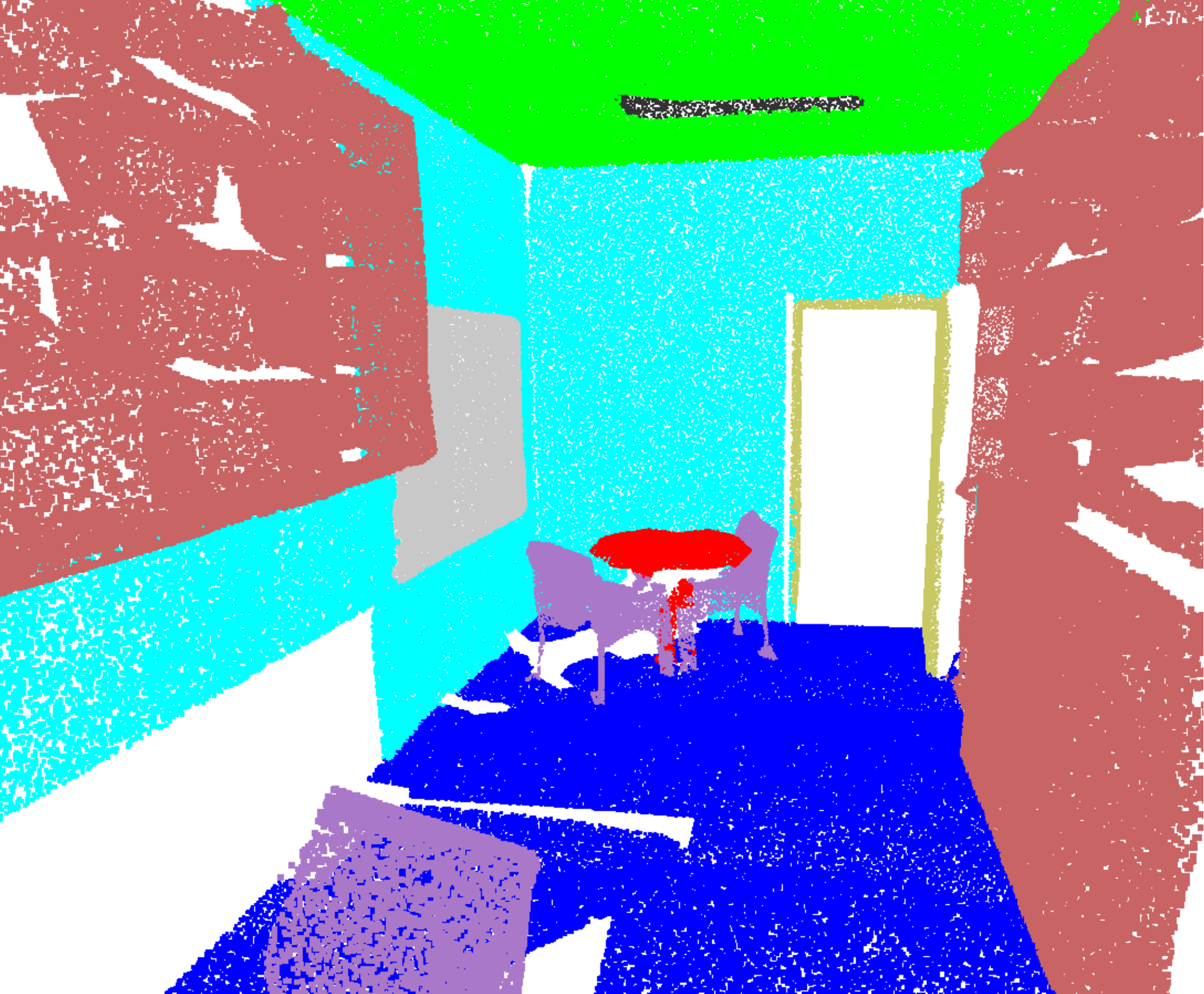}}%
	\hspace{2pt}\subfigure[Ground Truth]{\includegraphics[width=\unitxx]{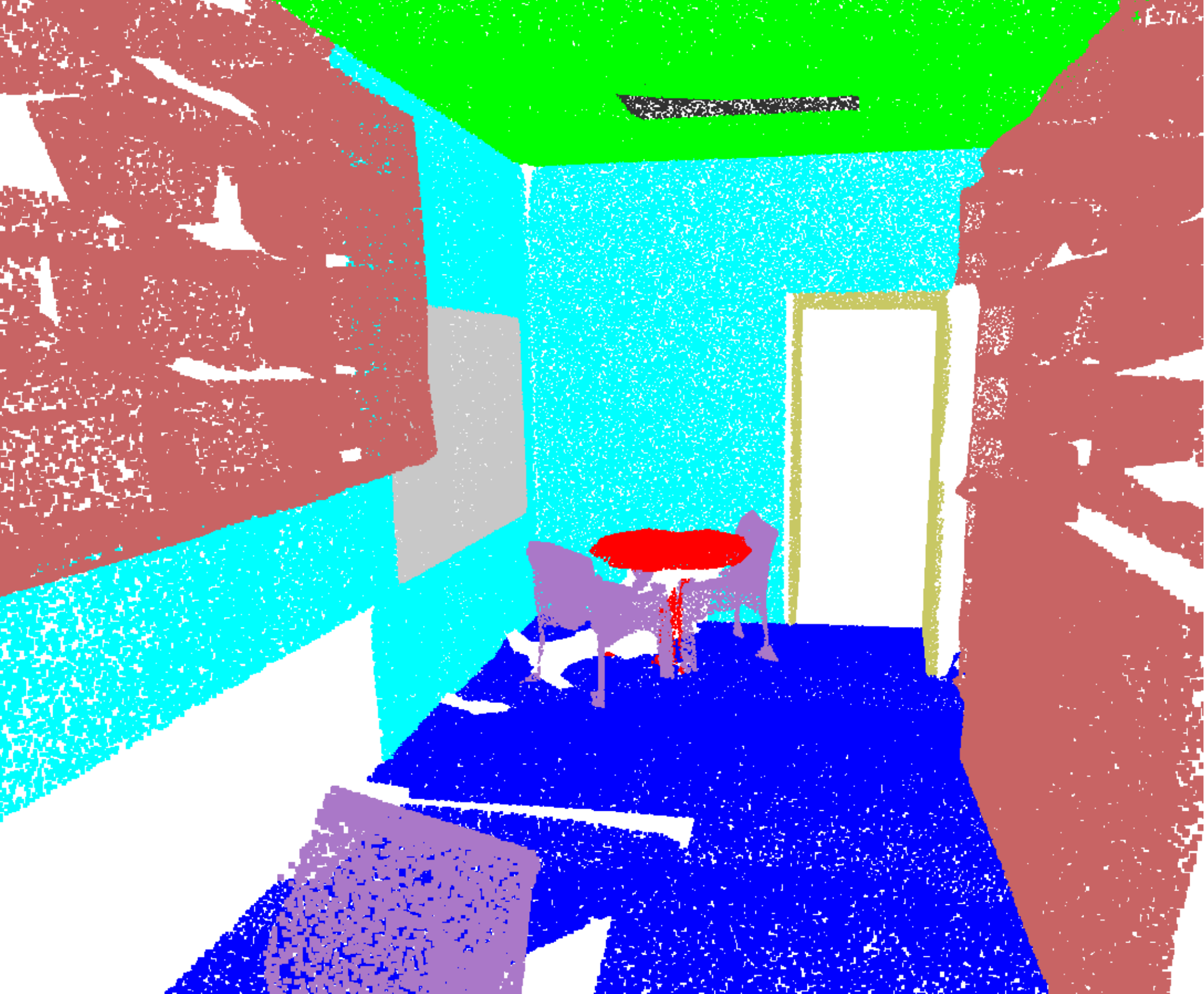}}
	
	\caption{Visualization of semantic segmentation results on S3DIS. We show the input point cloud, and labelled points mapped to RGB colors.}
	\label{fig:indoor}
\end{figure}

\vspace{5pt}\noindent\textbf{Results.} We report the mean classwise intersection over union (mIoU), mean classwise accuracy (mAcc) and overall accuracy (OA) in Tab.~\ref{table:scene_results}. The IoU of each class is also provided. The proposed AGConv outperforms the state-of-the-arts in most of the categories, which further demonstrates the effectiveness of adaptive convolutions over fixed kernels. The qualitative results are visualized in Fig.~\ref{fig:indoor} where we show rooms from different areas of the building. Our method can correctly detect less obvious edges of, e.g., pictures and boards on the wall.

\begin{figure}[t]
	\newlength{\unit}
	\setlength{\unit}{0.45\linewidth}
	\centering
	
	\includegraphics[width=\unit]{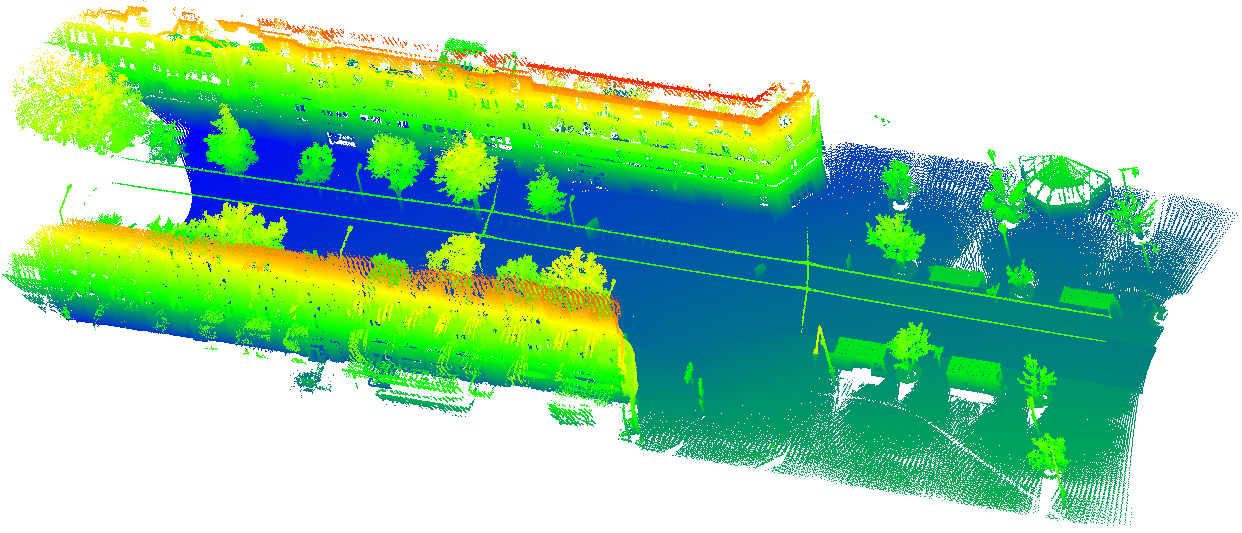}%
	\hspace{ 5pt}\includegraphics[width=\unit]{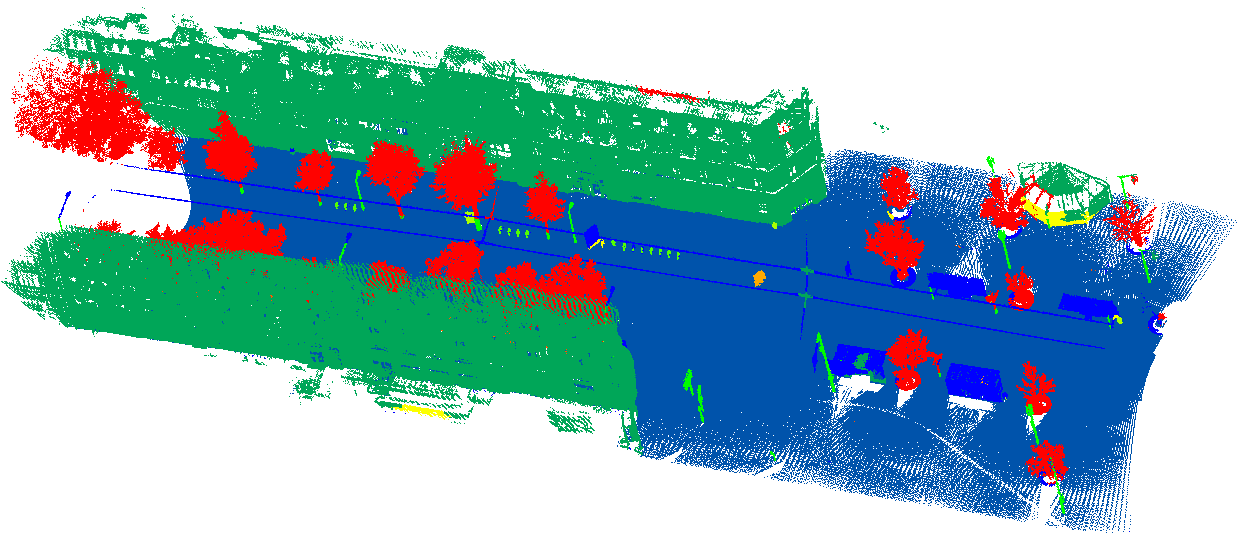}
	
	\vspace{3pt}
	
	\includegraphics[width=\unit]{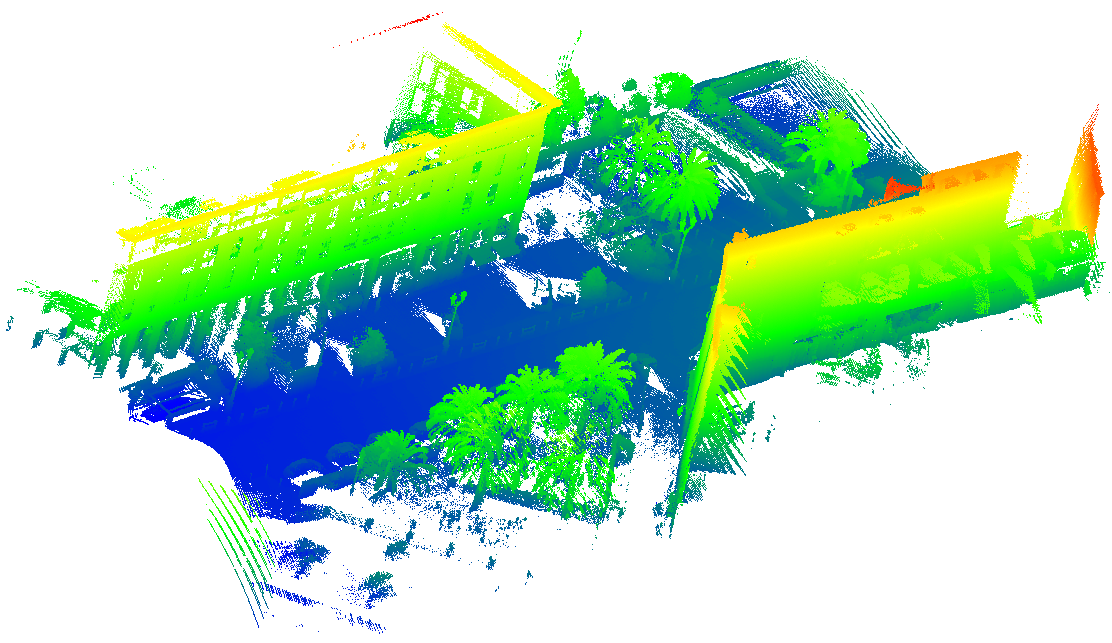}%
	\hspace{3pt}\includegraphics[width=\unit]{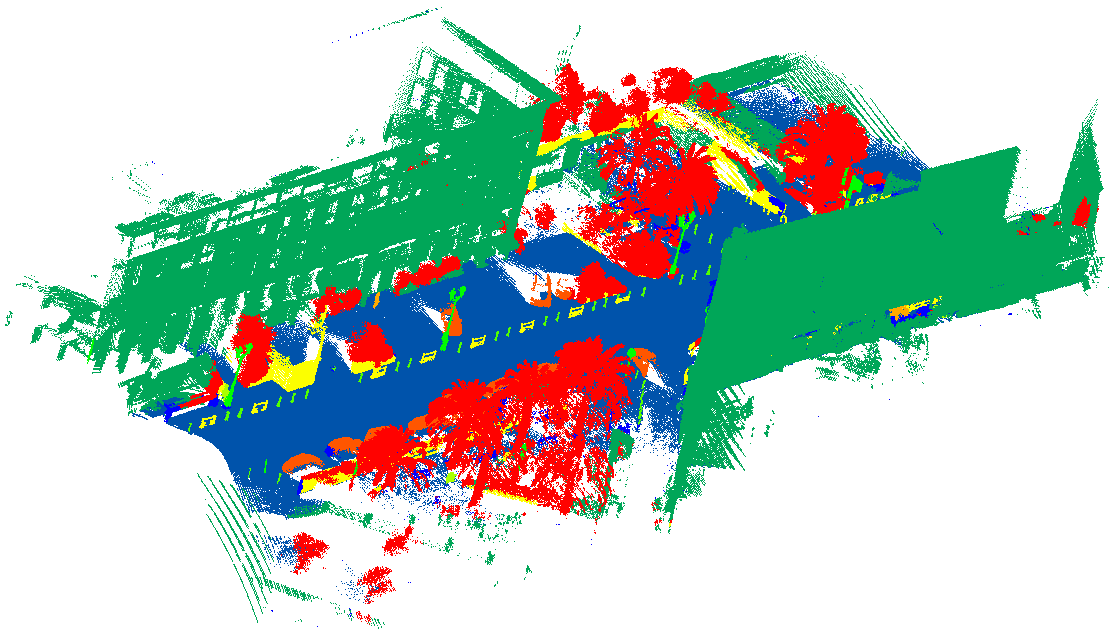}

	\vspace{-2pt}
	
	\subfigure[Input (LiDAR point clouds)]{\includegraphics[width=\unit]{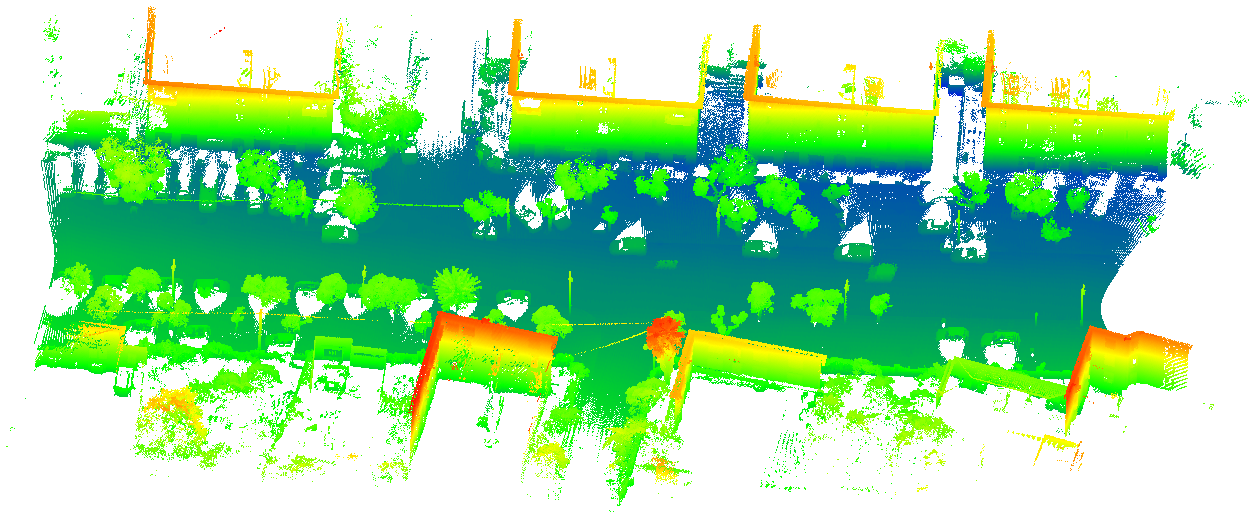}}%
	\hspace{5pt}\subfigure[Prediction]{\includegraphics[width=\unit]{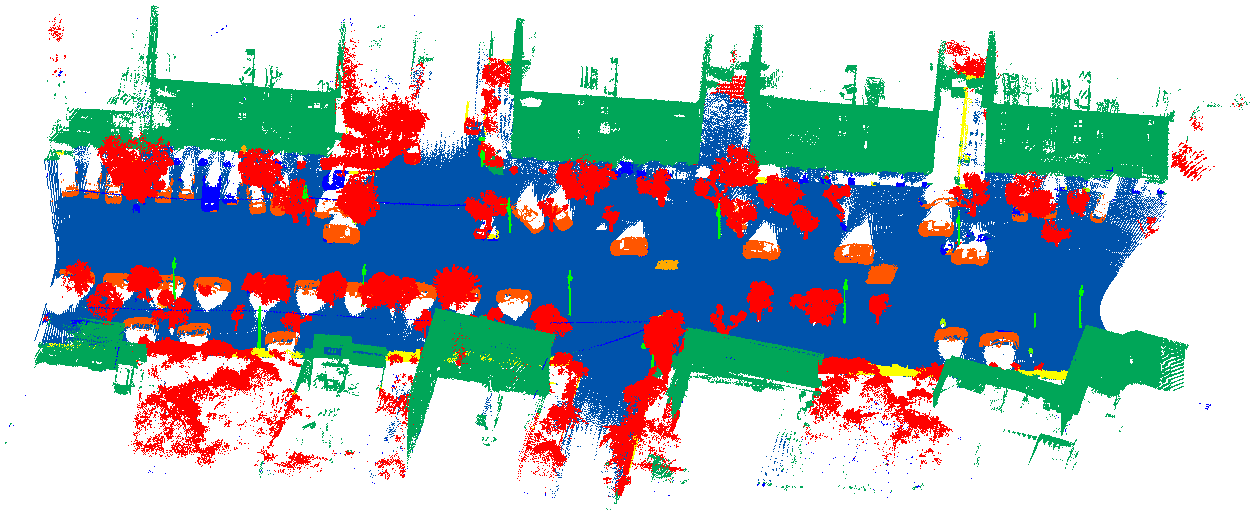}}

	\caption{Visualization of semantic segmentation results on NPM3D. 
	}
	\label{fig:outdoor}
\end{figure}

\subsection{Real-world outdoor scene segmentation}
\label{sec:eval:outdoor}

\vspace{5pt}\noindent\textbf{Data.} Paris-Lille 3D (NPM3D) is a large-scale yet real-world urban point cloud dataset acquired by a Mobile Laser System (MLS).
It contains 160 million points in total scanned from four different cities. 
To help the segmentation and classification tasks, the point cloud has been annotated to  10 coarse classes. 
To ensure a fair comparison, test labels are hidden and they provide only an online benchmark.

\vspace{5pt}\noindent\textbf{Challenge.} Compared with the indoor scenes, outdoor objects are often larger in scale, more complex in structure and contain much more noise. Moreover, the captured point clouds of outdoor scenes are sparse and have a large amount of object classes. However, single objects, especially the small targets, are usually sampled with very few points. These challenges cannot make the downstream applications (e.g., segmentation, detection and recognition)  operate smoothly. To alleviate the aforementioned problems, a deep network that possesses a powerful ability of extracting features is more than welcome. Following the experimental setup in Sec.~\ref{sec:eval:indoor}, we implement AGConv by replacing the GraphConv layer with AGConv in the KPConv~\cite{thomas2019kpconv} architecture.

\vspace{5pt}\noindent\textbf{Results.} To demonstrate the effectiveness of AGConv, we compare our method with KPConv\cite{thomas2019kpconv}, ConvPoint\cite{boulch2020convpoint}, MS3\textunderscore DVN\cite{roynard2018classification}, HDGCN\cite{liang2019hierarchical}, and RF-MSSF \cite{thomas2018RFMSSF}. 
The final mIoU results and the IoU of each class are reported in Tab. \ref{table:outdoor_scene_results}. All quantitative results come from the official website of the dataset or the corresponding papers. 
After performing AGConv on KPConv, mIoU of the scene segmentation has been improved, which is superior to the competitors. 
This indicates that our adaptive convolution can identify effective information of each feature map, and capture different relations between points from different semantic components. 
We show visualized segmentation results in Fig. \ref{fig:outdoor} where our method correctly identifies outdoor sparse point cloud scenes.

\subsection{Ablation studies}
\label{sec:eval:ablation}
In this section, we explain some of the architecture choices used in our network, and demonstrate the effectiveness of AGConv compared to several ablation networks.

\vspace{5pt}\noindent\textbf{Adaptive Convolution \textit{vs} Fixed-kernel Convolution.} We compare our AGConv with the fixed-kernel convolutions, including those using the attention mechanism and standard graph convolution (DGCNN\cite{wang2019dynamic}). We design several ablation networks by replacing the AGConv layers with fixed kernel layers and keeping the other architectures the same.


Specifically, Velickovic et al. \cite{velivckovic2017graph} assign attentional weights to different neighboring points and Wang et al. \cite{wang2019graph} further design a channel-wise attentional function. We use their layers and denote these two ablations as Attention Point and Attention Channel, respectively. Following  \cite{velivckovic2017graph}, the output feature is formulated as:
	\begin{equation}
		f_i' = \max_{j \in \mathcal{N}(i)} a_{ij} * h(f_j),
	\end{equation}
	where $h: \mathbb{R}^{D} \rightarrow \mathbb{R}^{M}$ is a shared MLP and $a_{ij}$ is the attentional weight calculated as:
	\begin{equation}
		a_{ij} = \mathbf{softmax}_j (\alpha(\Delta f_{ij})).
	\end{equation}
Here, $\alpha(\cdot)$ is a mapping function, and $\Delta f_{ij} = [h(f_i), h(f_j) - h(f_i)]$ since the attentional weights are applied to $h(f_j)$ instead of $f_j$. A softmax is used to make $\sum_j a_{ij} = 1, j \in \mathcal{N}(i)$. Attention Point uses $a_{ij} \in \mathbb{R}$, i.e., the function is $\alpha: \mathbb{R}^{2M} \rightarrow \mathbb{R}$. Attention Channel uses $a_{ij} \in \mathbb{R}^M$ and, in this case, $*$ denotes the element-wise product. The attentional weights $a_{ij}$ are based on the produced features $h(f_j)$ in order to determine the different contributions of the neighboring points. 
However, since $h(\cdot)$ is still a fixed/isotropic one as discussed, they still cannot solve the intrinsic limitation of current graph convolutions. 

To compare our model with attentional graph convolutions, we only replace the AGConv layers in our network and the feature inputs $\Delta f_{ij}$ are the same as our model. Besides, we also show the result by standard graph convolutions (GraphConv), which is a similar version of DGCNN \cite{wang2019dynamic}. In Tab.~\ref{table:ab:seg}, these ablation networks are trained on the ShapeNetPart dataset for segmentation. Tab.~\ref{table:ablation} shows the classification comparison on ModelNet40. In both the classification and segmentation tasks, AGConv achieves better results than the fixed-kernel graph convolutions.

\vspace{5pt}\noindent\textbf{Feature decisions.} In AGConv, the adaptive kernel is generated from the feature input $\Delta f_{ij}$, and subsequently convolved with the corresponding  $\Delta x_{ij}$. 
Note that, in our experiments, $\Delta x_{ij}$ corresponds to the $(\mathbf{x}, \mathbf{y}, \mathbf{z})$ spatial coordinates of the points. We discuss several other choices of $\Delta x_{ij}$ in Eq.~\ref{equ:convolution} in Sec.~\ref{sec:method:adapt}, which are evaluated by designing these ablations:

$\bullet$ Feature - In Eq.~\ref{equ:convolution}, we convolve the adaptive kernel $\hat{e}_{ijm}$ with their current point features. That is, $\Delta x_{ij}$ is replaced with $\Delta f_{ij}$ and the kernel function is $g_m: \mathbb{R}^{2D} \rightarrow \mathbb{R}^{2D}$. The kernel learns to adapt to the features of the previous layer and extract feature relations, which is a more general convolution operator.

$\bullet$ Initial attributes - The point normals $(n_x, n_y, n_z)$ are included in the part segmentation task on ShapeNetPart, leading to a 6-dimensional initial feature attributes for each point. Thus, we design three ablations where we use only spatial inputs (Ours), only normal inputs (Normal) and both of them (Initial attributes). The kernel function is modified correspondingly.

The resulting IoU scores are shown in Tab.~\ref{table:ab:seg}. As one can see, $(\mathbf{x}, \mathbf{y}, \mathbf{z})$ is the most critical initial attribute (probably the only attribute) in point clouds, thus it is recommended to use them in the convolution with adaptive kernels. Although achieving a promising result, the computational cost for the Feature ablation is extremely high since the network expands heavily when it is convolved with a high-dimensional feature.

In summary, we recommend xyz rather than feature in that: (i) the point feature $f_j$ has been already included in the adaptive kernel and convolving again with $f_j$ leads to redundancy of feature information; (ii) it is easier to learn spatial relations through MLPs, instead of detecting feature correspondences in a high-dimensional space (e.g., 64, 128 dimensional features); and (iii) the last reason is the memory cost and more specifically the large computational graph in the training stage which cannot be avoided.

\begin{figure}[t]
	\centering
	
	\includegraphics[width=0.98\linewidth]{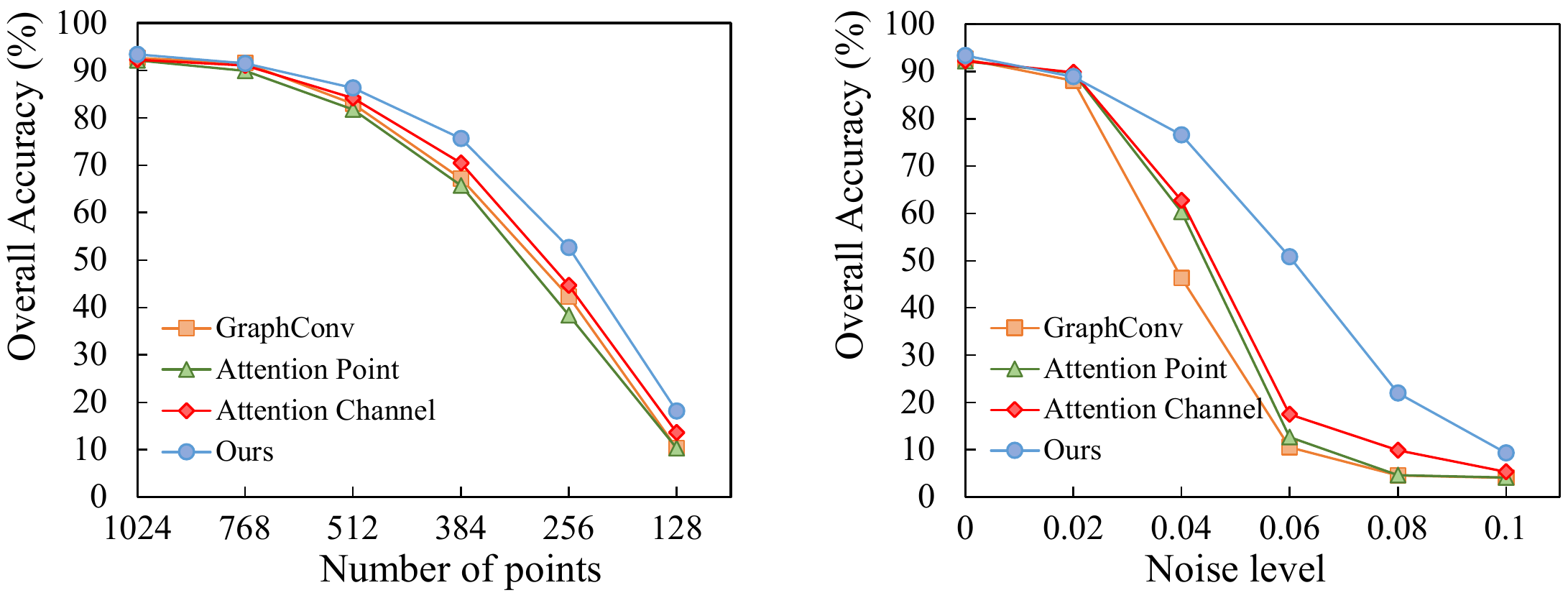}
	
	\caption{Robustness test on ModelNet40 for classification. GraphConv indicates the standard graph convolution network. Attention Point and Attention Channel indicate the ablations where we replace the AGConv layers with graph attention layers (point- and channel-wise), respectively. Our model is more robust to point density and noise perturbation.}
	\label{fig:robust}
\end{figure}

\begin{table}[t]
	\centering
	\small
	\setlength{\tabcolsep}{5mm}
	\begin{tabular}{c|cc} 
		\toprule[1pt]
		Number $k$ & mAcc(\%) & OA(\%) \\
		\midrule[0.3pt]
		\midrule[0.3pt]
		5						& 89.4 & 92.8 \\
		10						& \textbf{90.7} & 93.2 \\
		20						& \textbf{90.7} & \textbf{93.4} \\
		40						& 90.4 & 93.0 \\
		\bottomrule[1pt]
	\end{tabular}
	\vspace{5pt}
	\caption{Our classification network with different $k$ of the nearest neighbors.}
	\label{table:numberk}
\end{table}

\begin{figure}[t]
	\centering
	
	\includegraphics[width=0.245\linewidth]{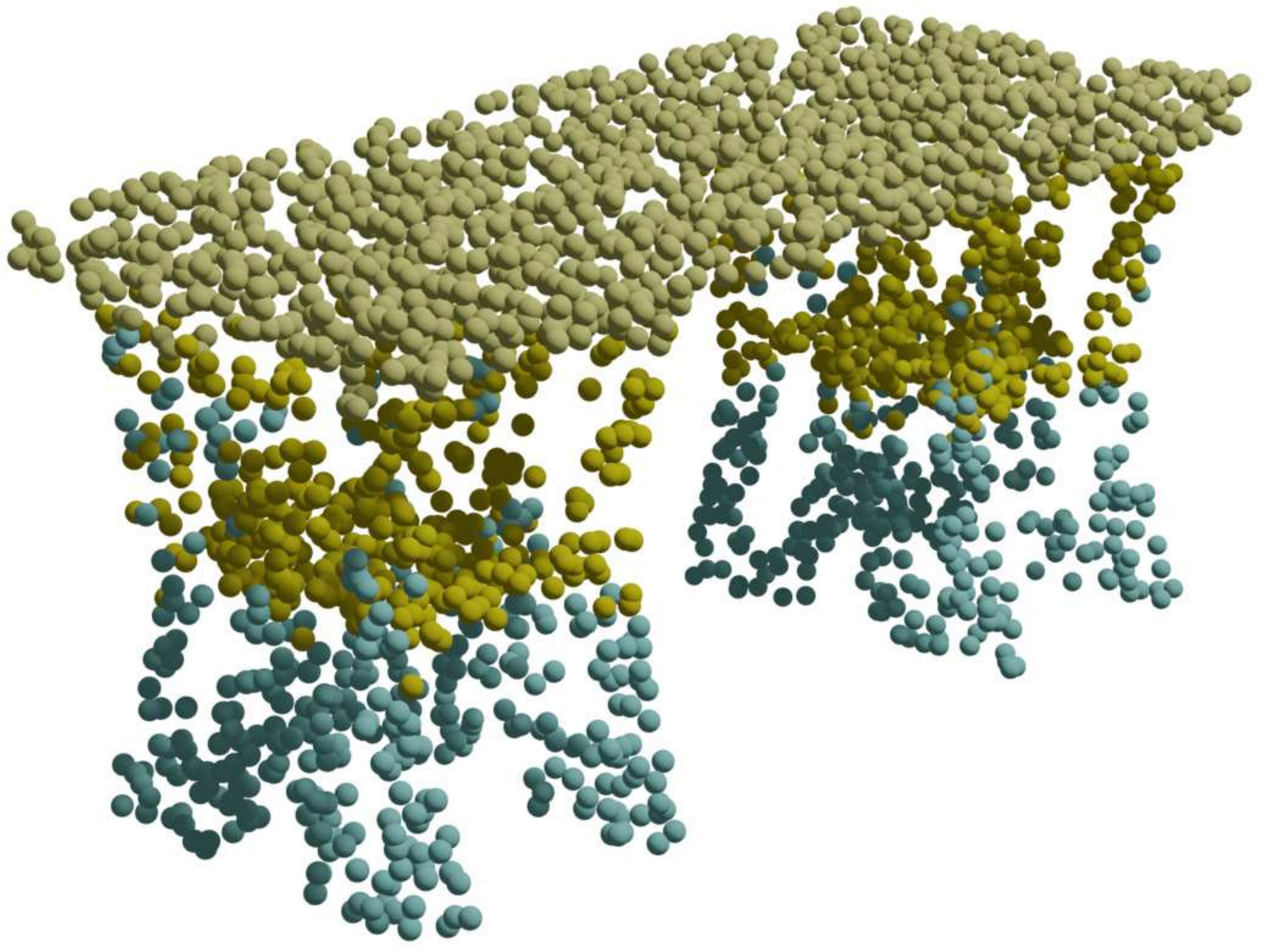}%
	\hspace{0pt}\includegraphics[width=0.245\linewidth]{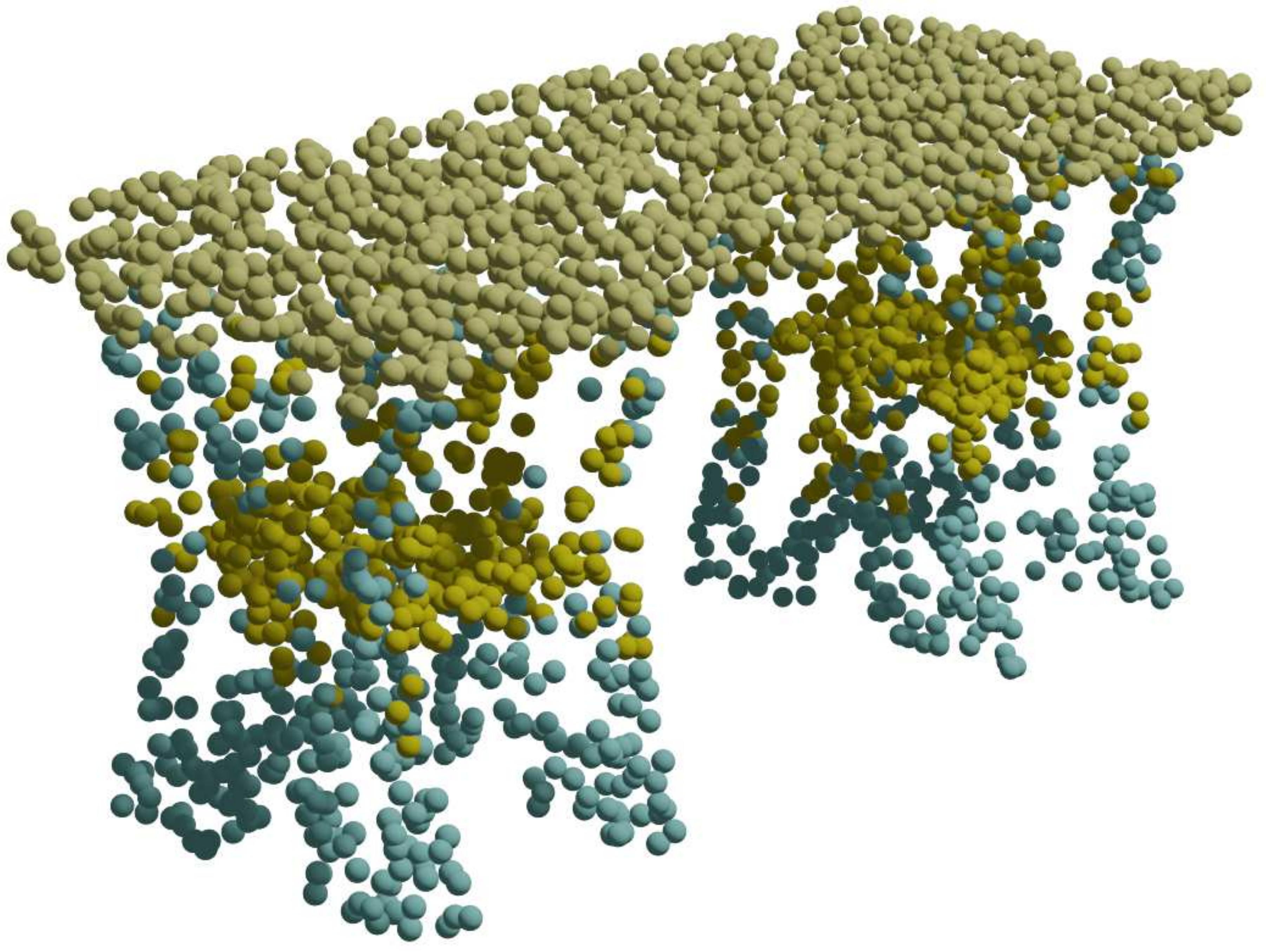}
	\hspace{0pt}\includegraphics[width=0.245\linewidth]{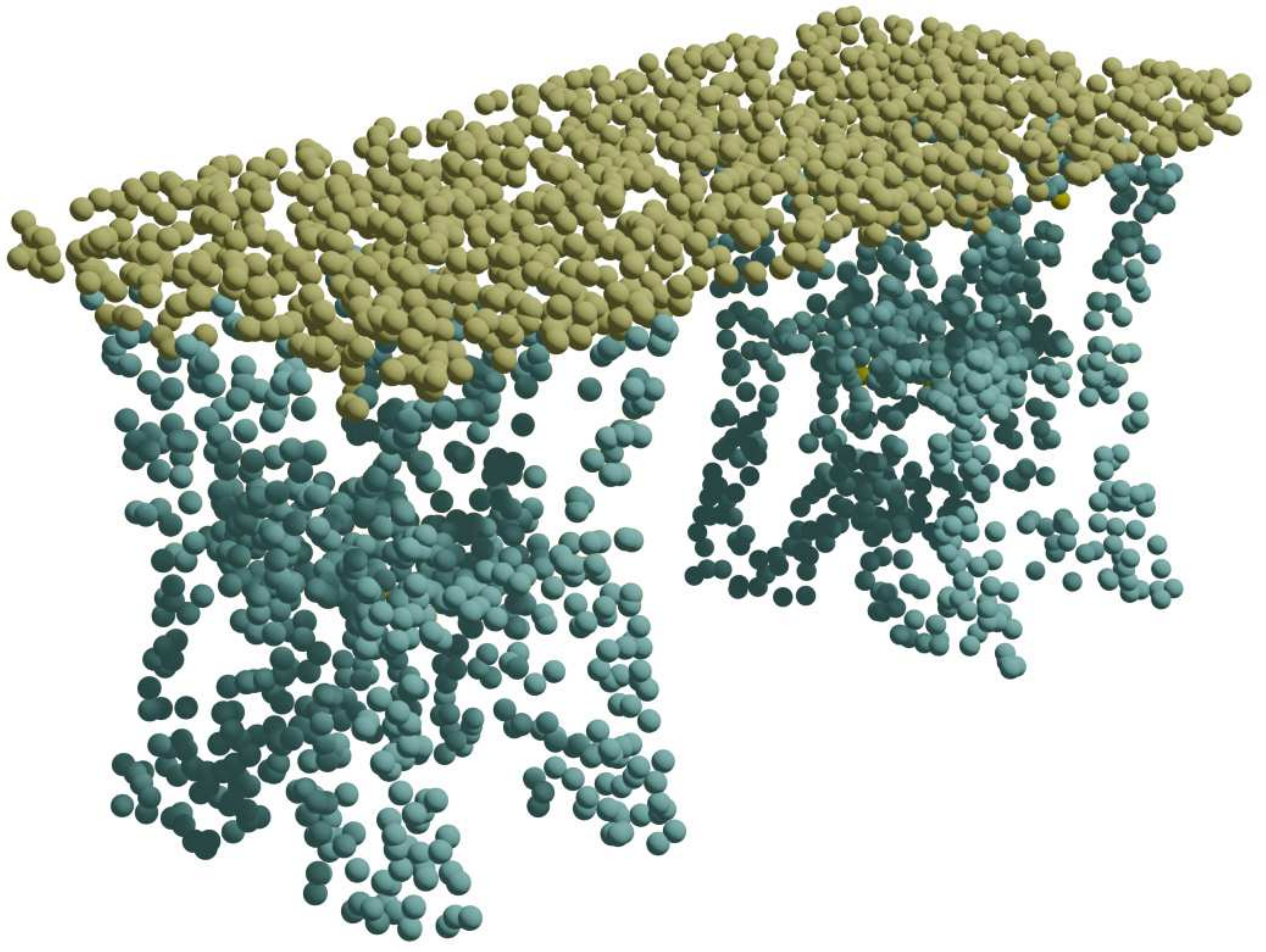}%
	\hspace{0pt}\includegraphics[width=0.245\linewidth]{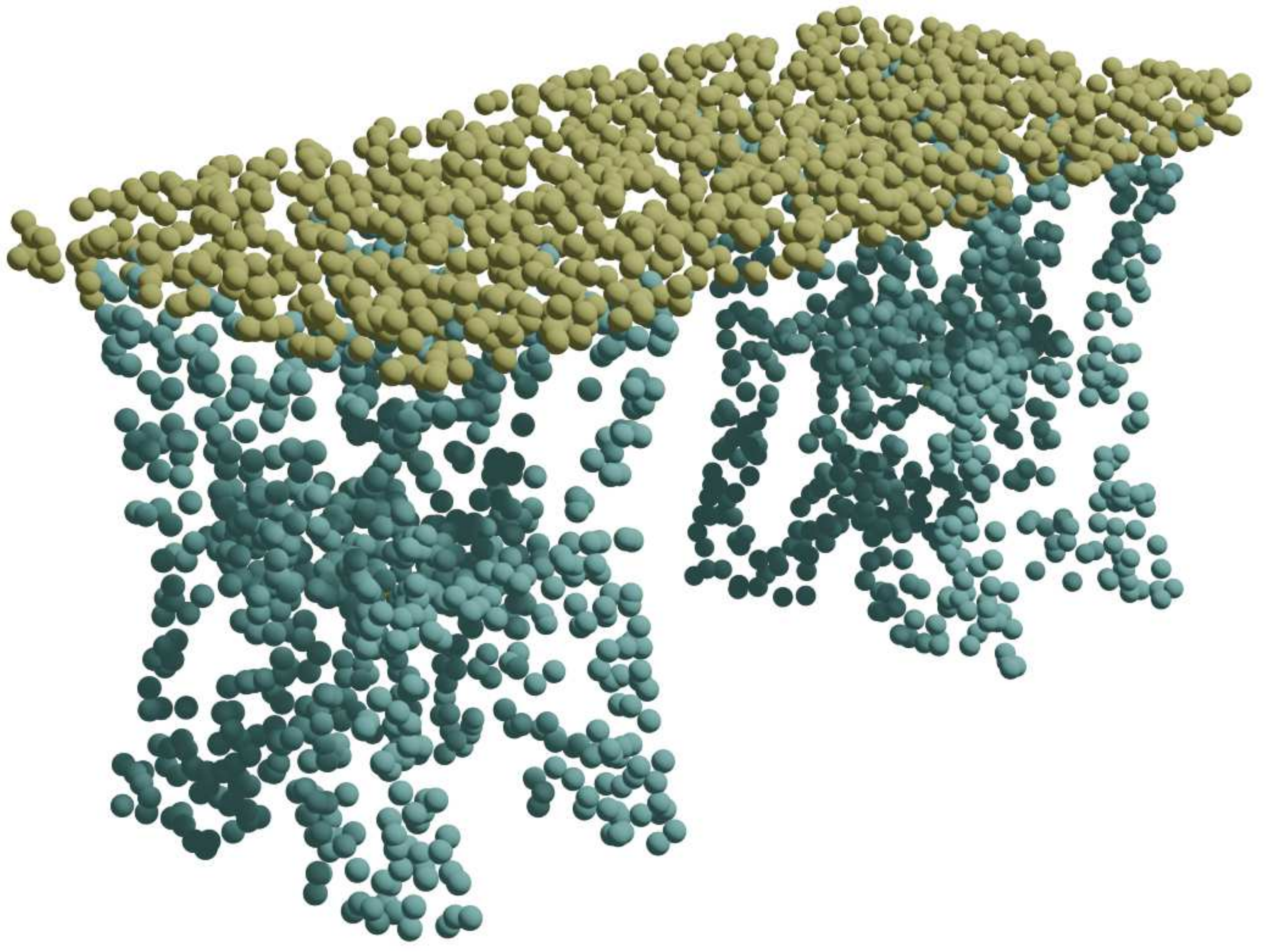}
	
	
	
	

	\subfigure[DGCNN]{\includegraphics[width=0.2\linewidth]{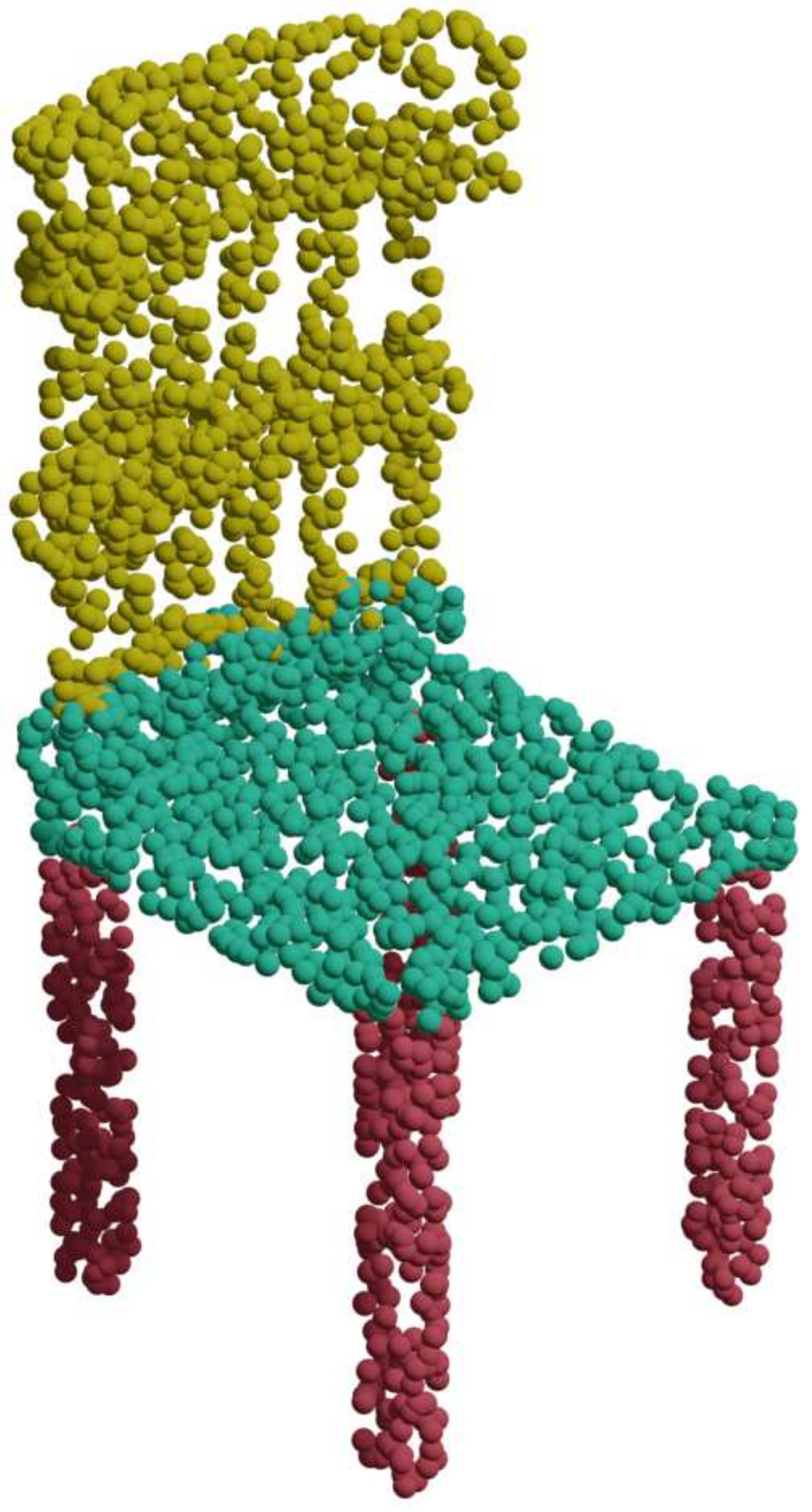}}%
	\hspace{10pt}\subfigure[Attention]{\includegraphics[width=0.2\linewidth]{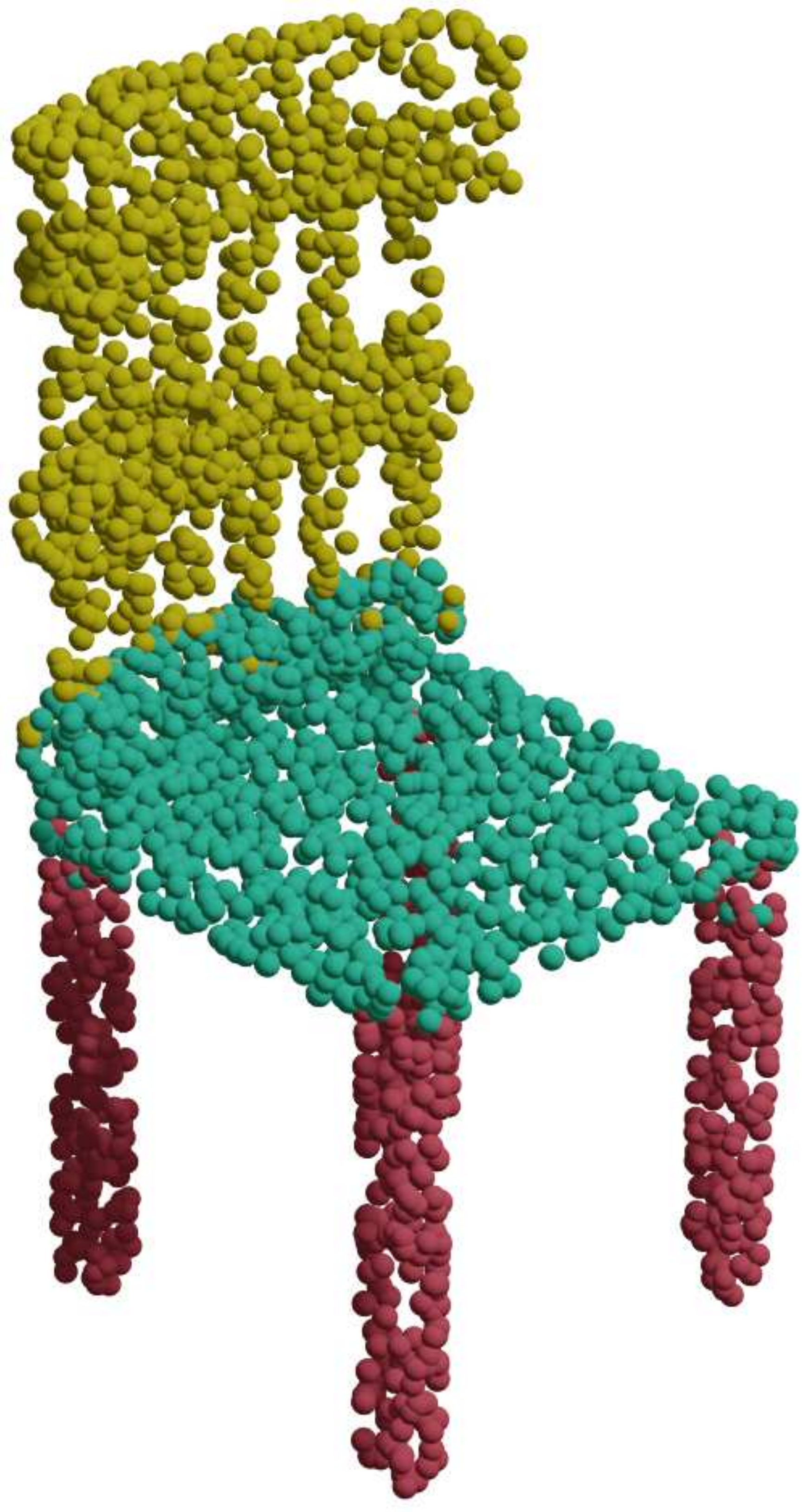}}%
	\hspace{10pt}\subfigure[Ours]{\includegraphics[width=0.2\linewidth]{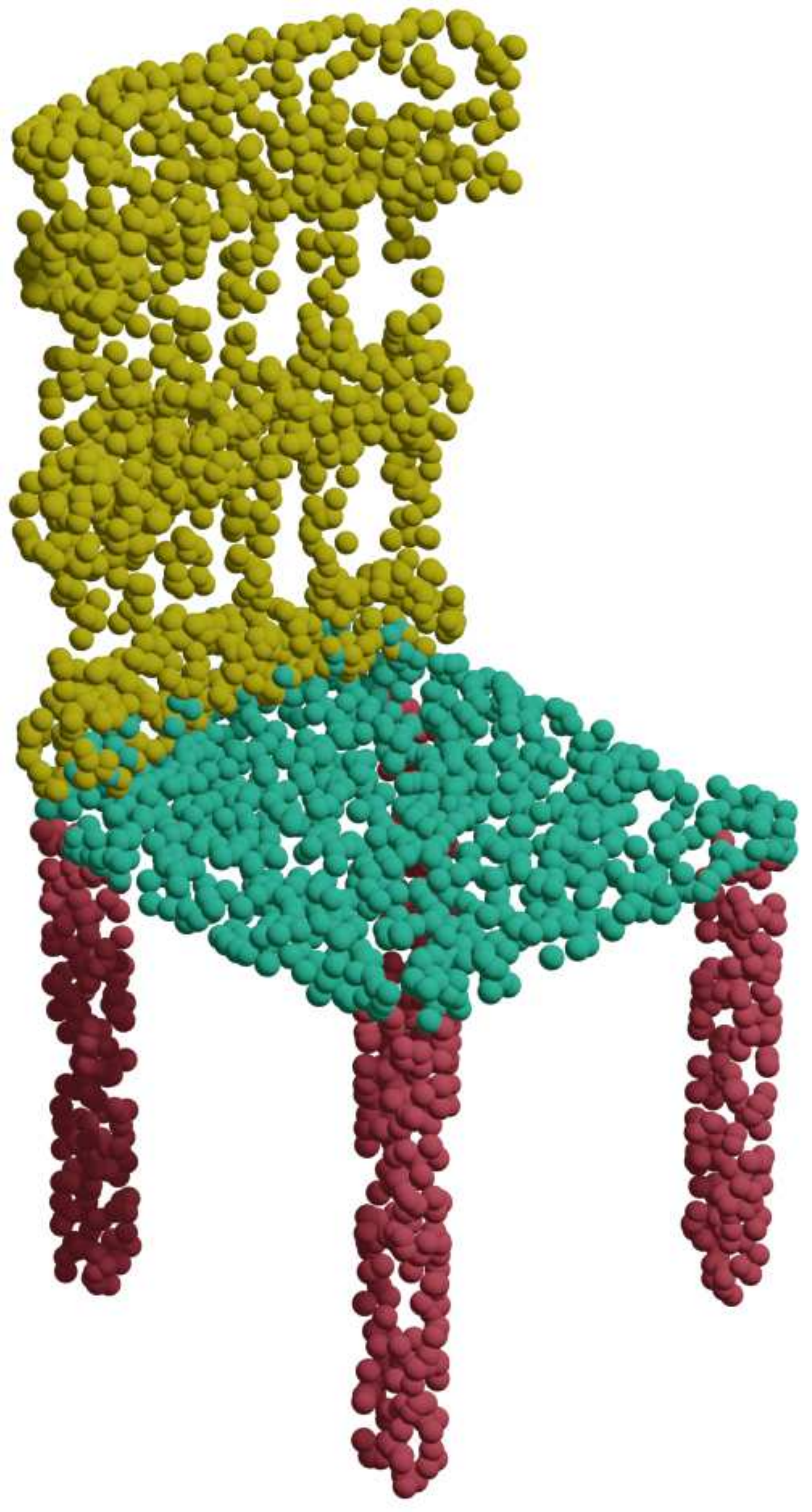}}%
	\hspace{10pt}\subfigure[GT]{\includegraphics[width=0.2\linewidth]{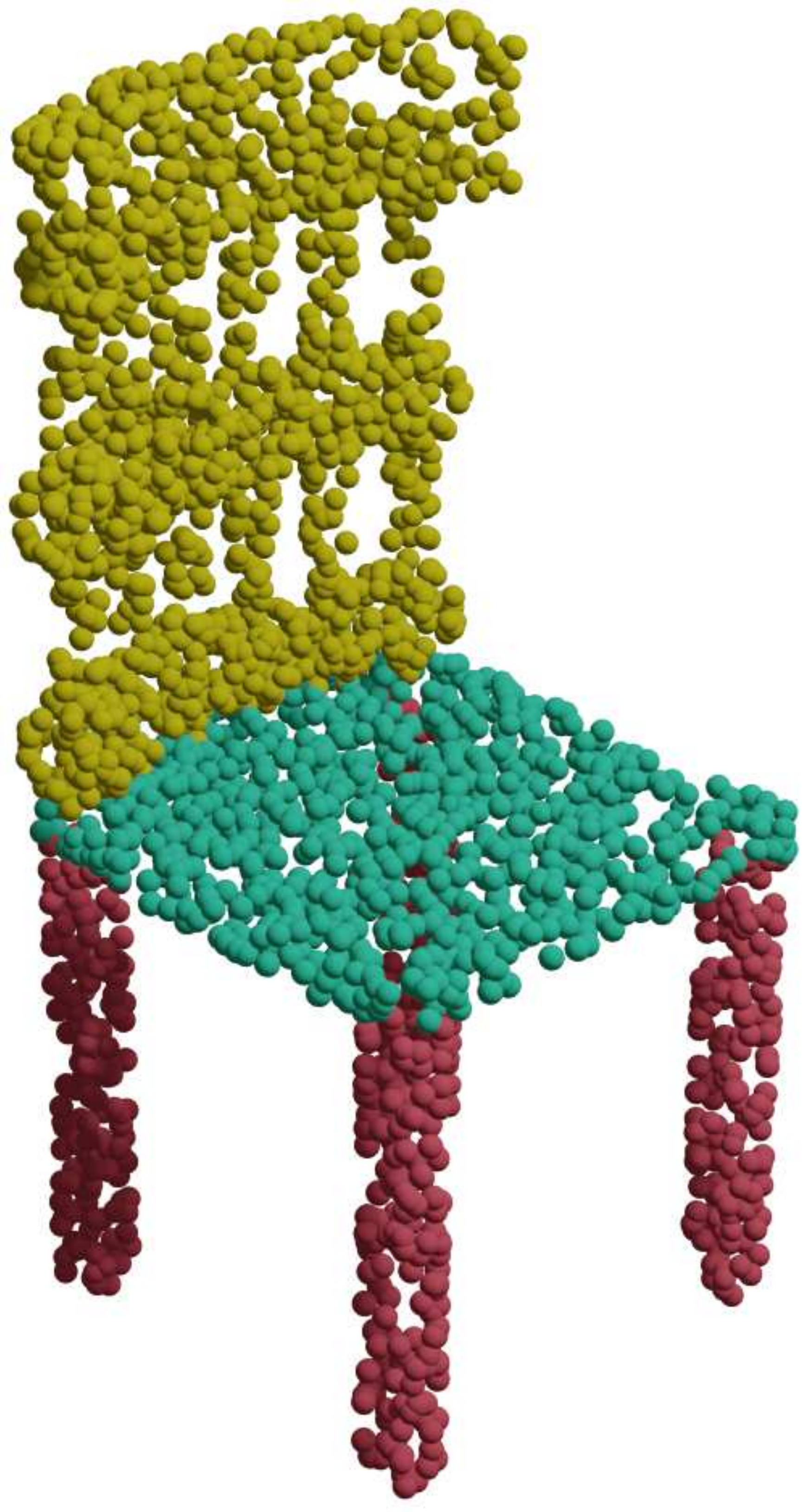}}
	
	\caption{Segmentation results on ShapeNet. The labelled points are visualized in different colors. We compare our adaptive graph convolution with DGCNN \cite{wang2019dynamic} (standard graph convolution) and attentional convolution network (Attention Point). Our method produces better results especially for points close to the object boundaries and edges.}
	\label{fig:partseg}
\end{figure}

\begin{figure*}
	\centering
	\newlength{\unitx}
	\setlength{\unitx}{0.14\linewidth}
	
	\includegraphics[width=\unitx]{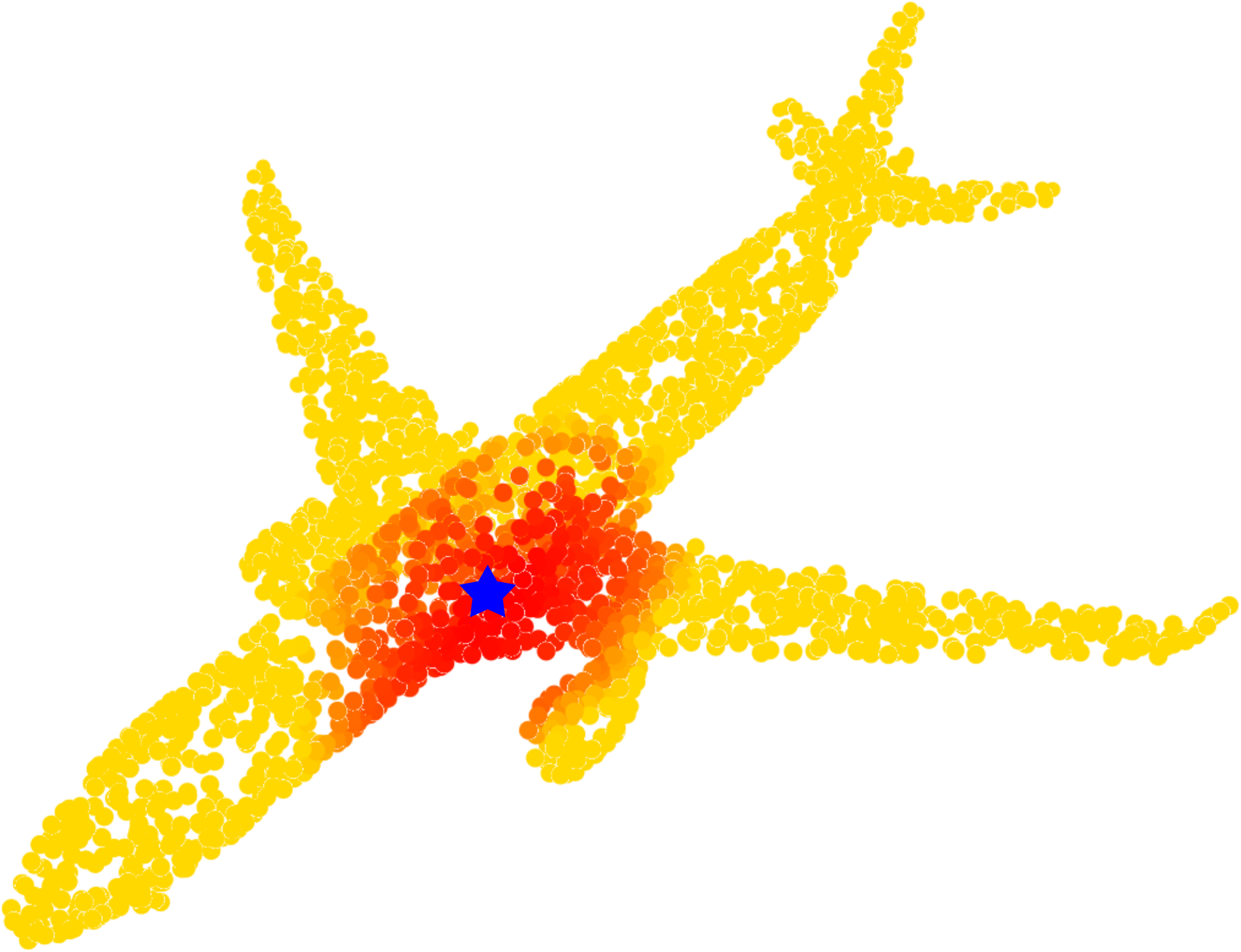}%
	\includegraphics[width=\unitx]{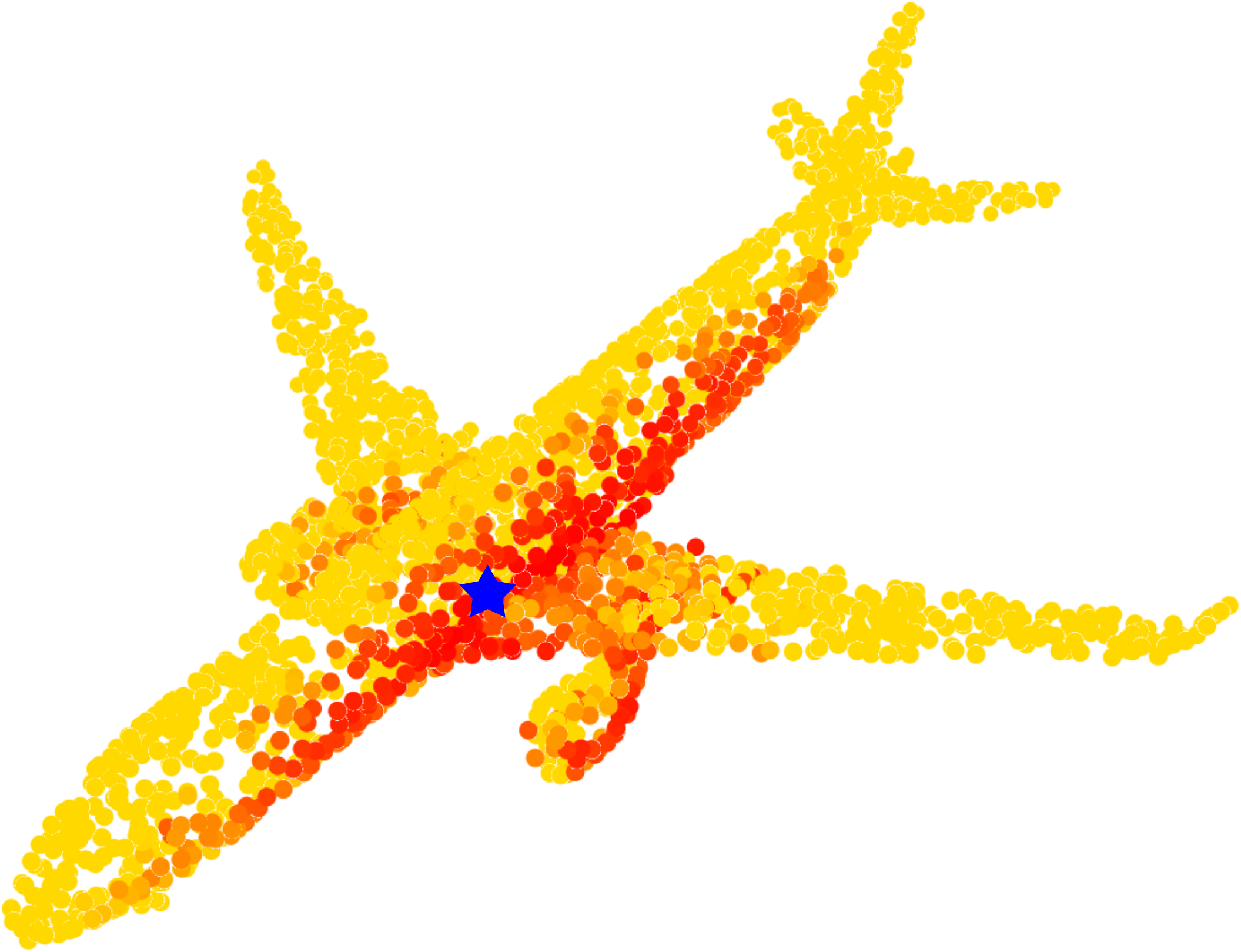}%
	\includegraphics[width=\unitx]{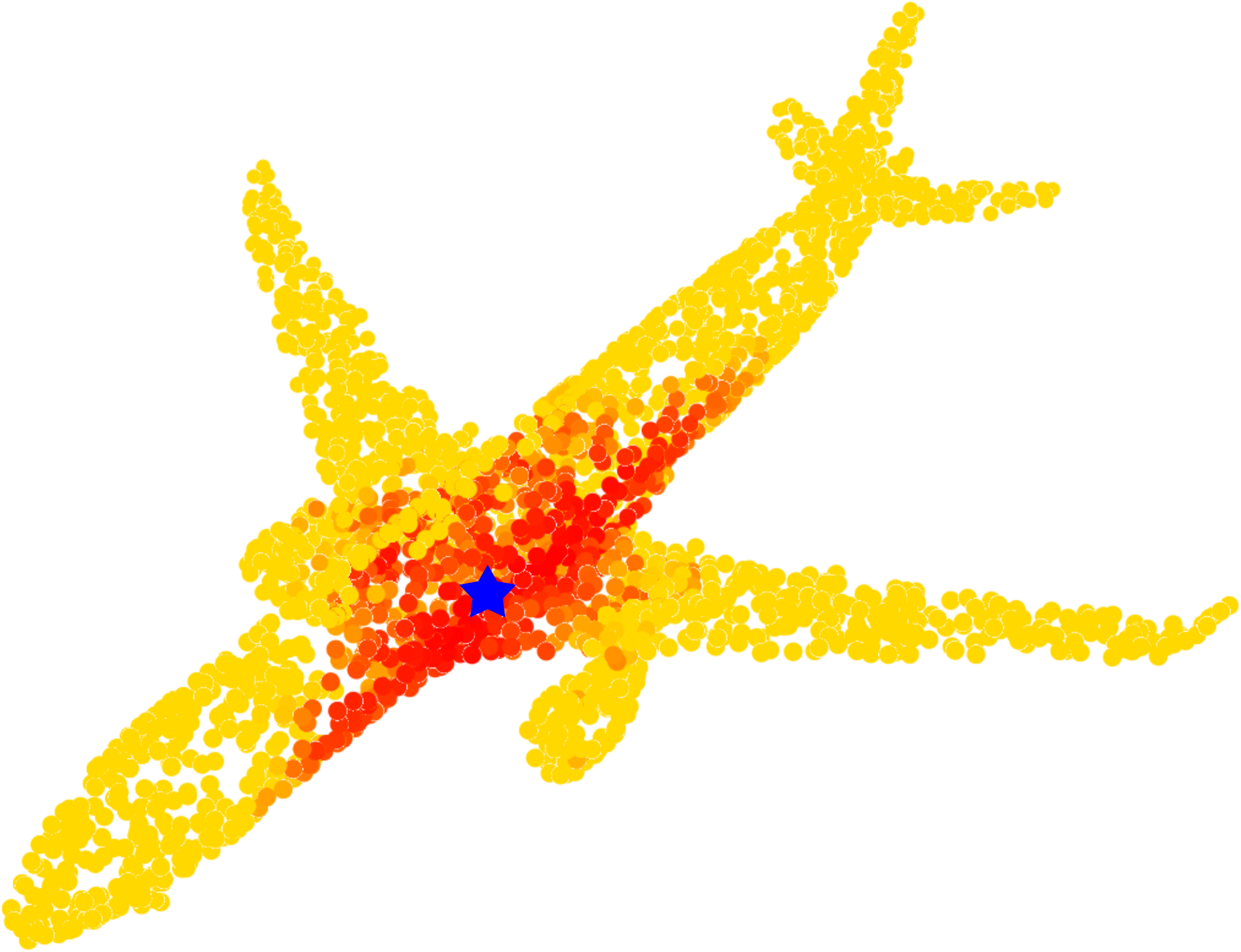}%
	\includegraphics[width=\unitx]{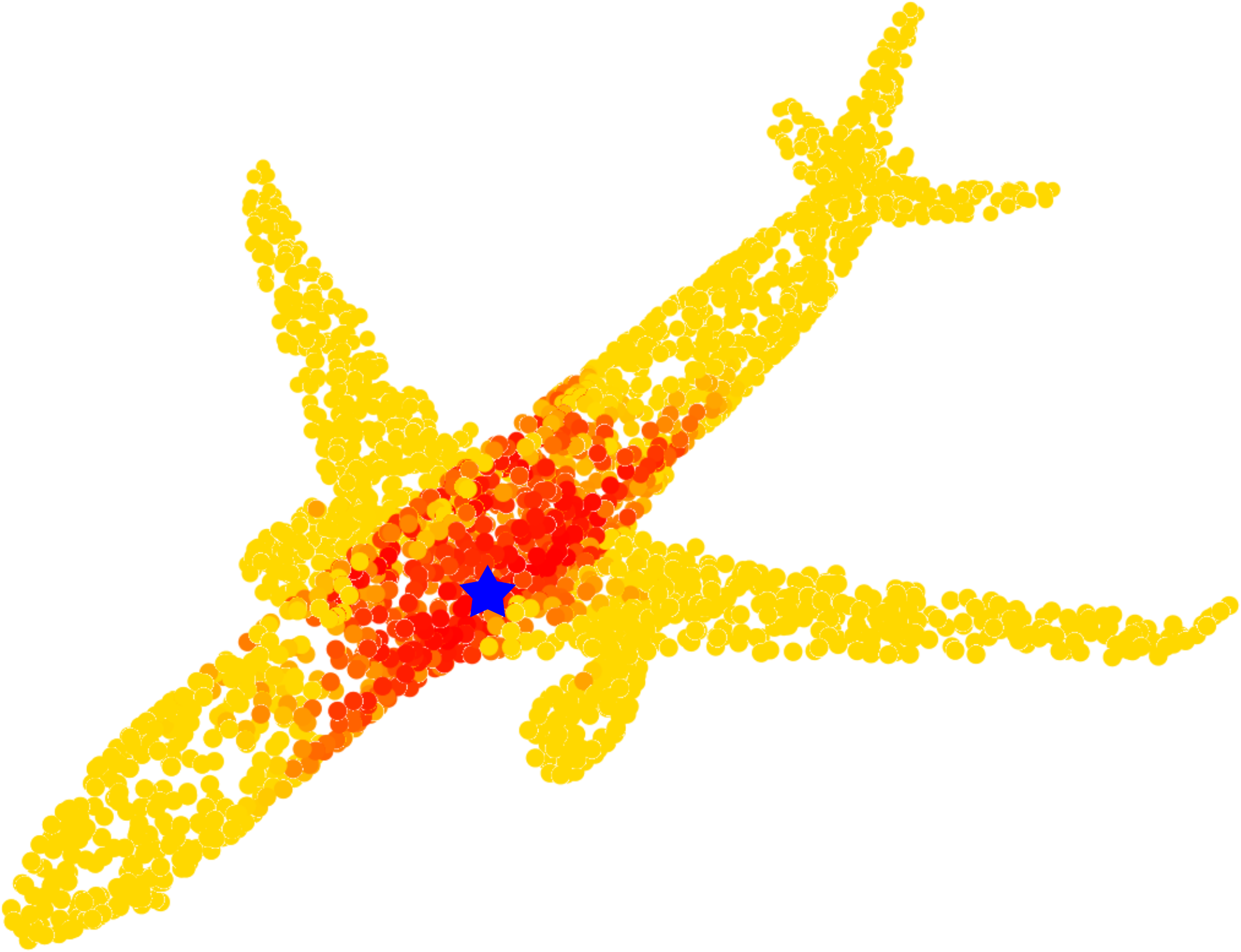}%
	\includegraphics[width=\unitx]{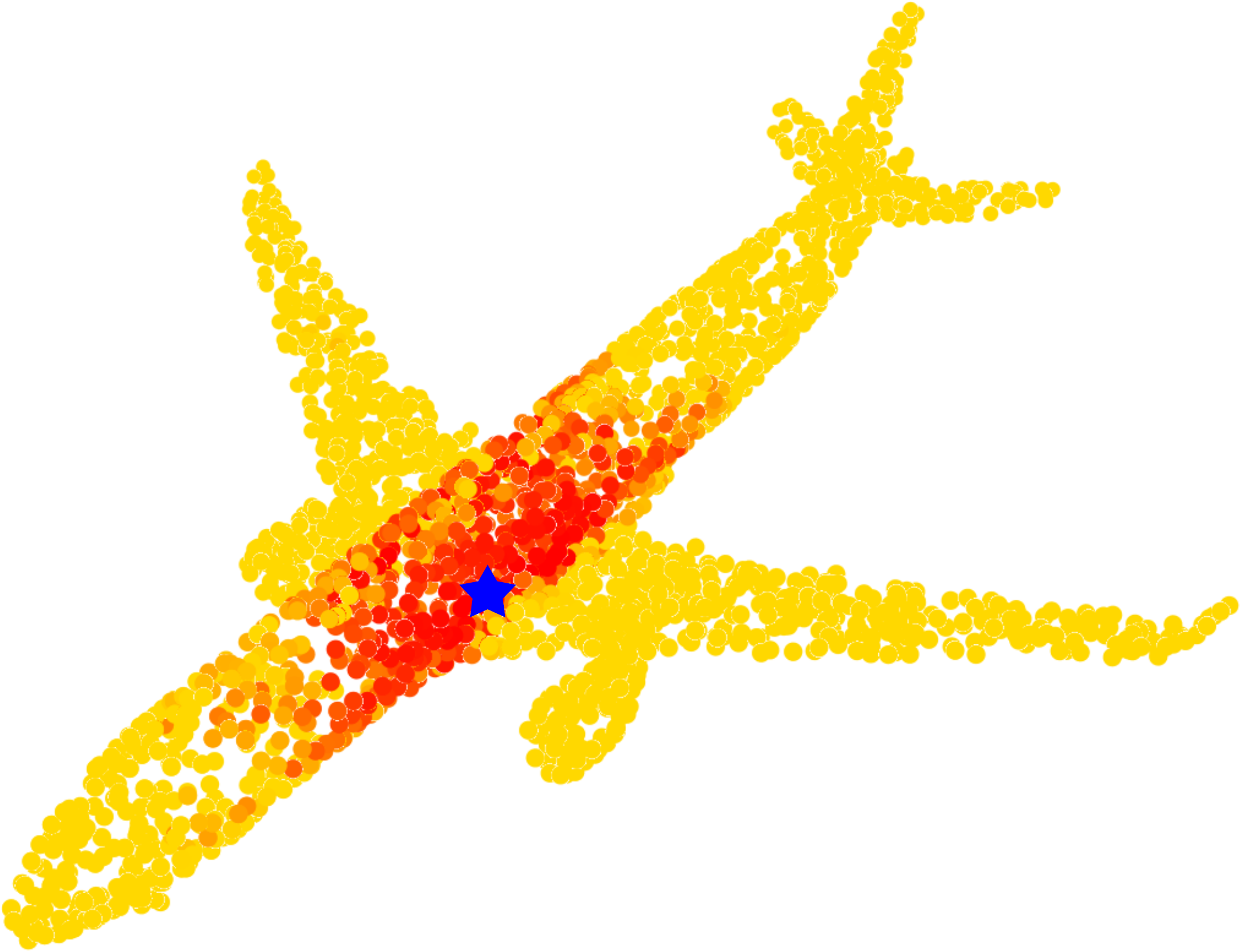}%
	\includegraphics[width=\unitx]{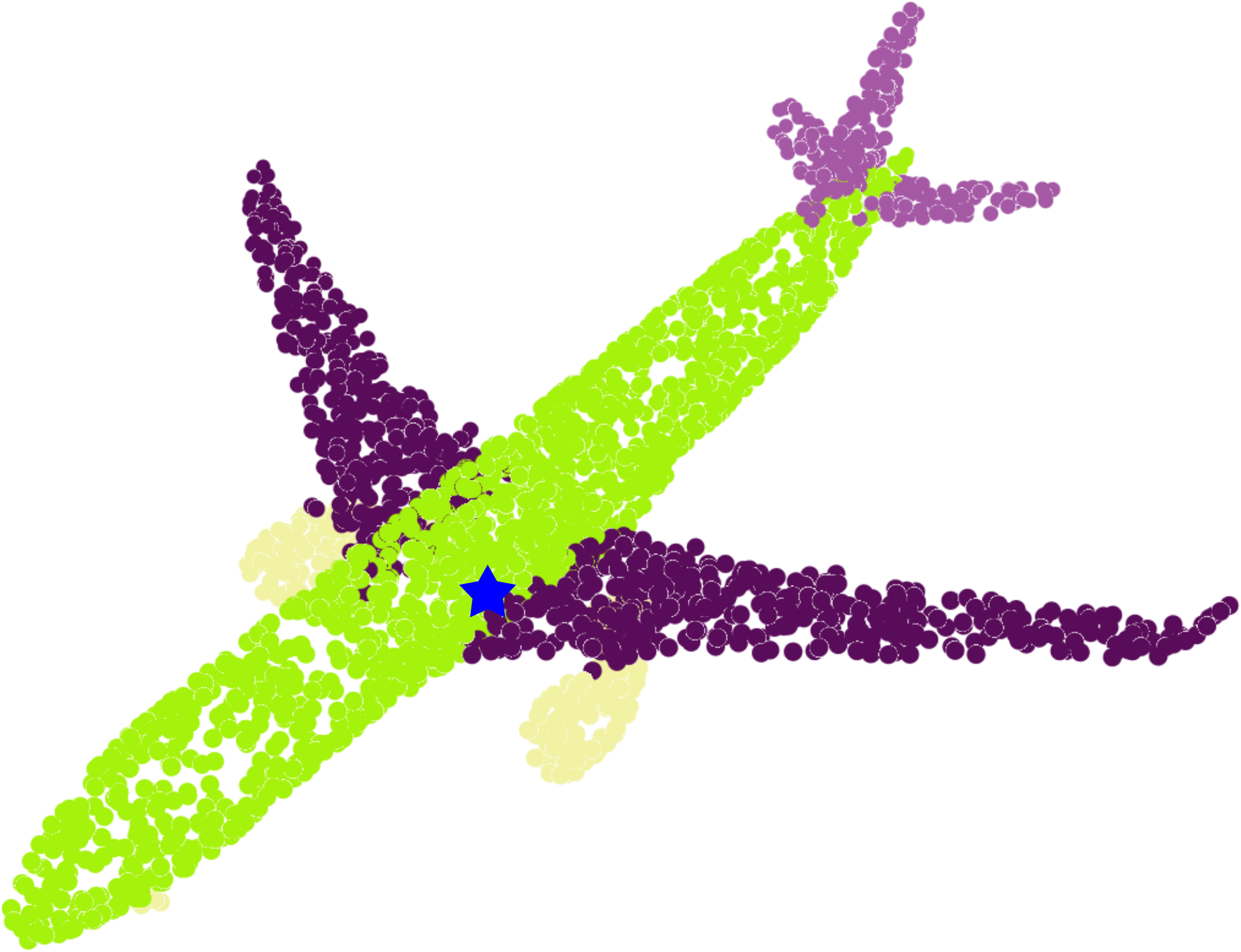}%
	\includegraphics[width=\unitx]{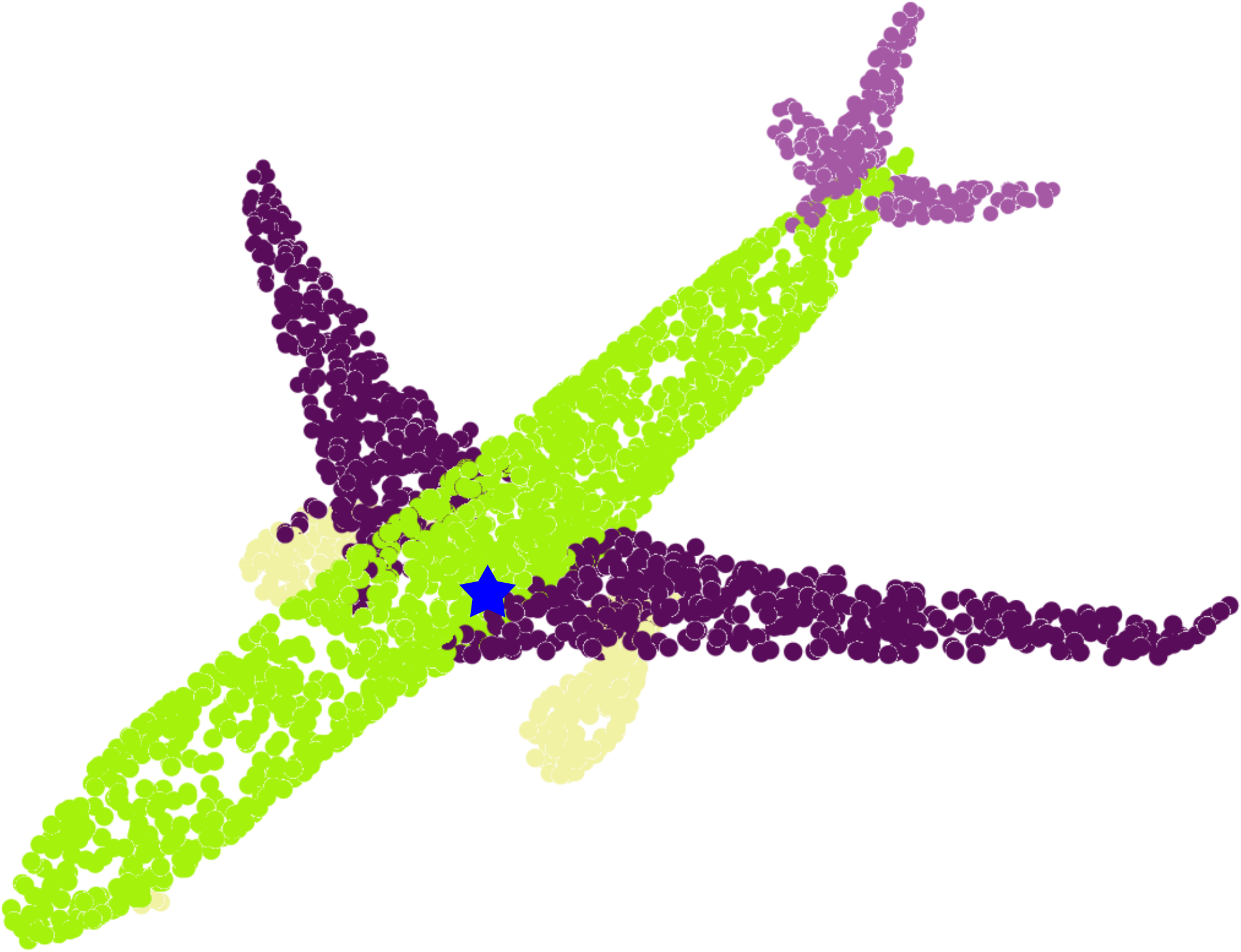}
	
	\subfigure[Spatial]{\includegraphics[width=\unitx]{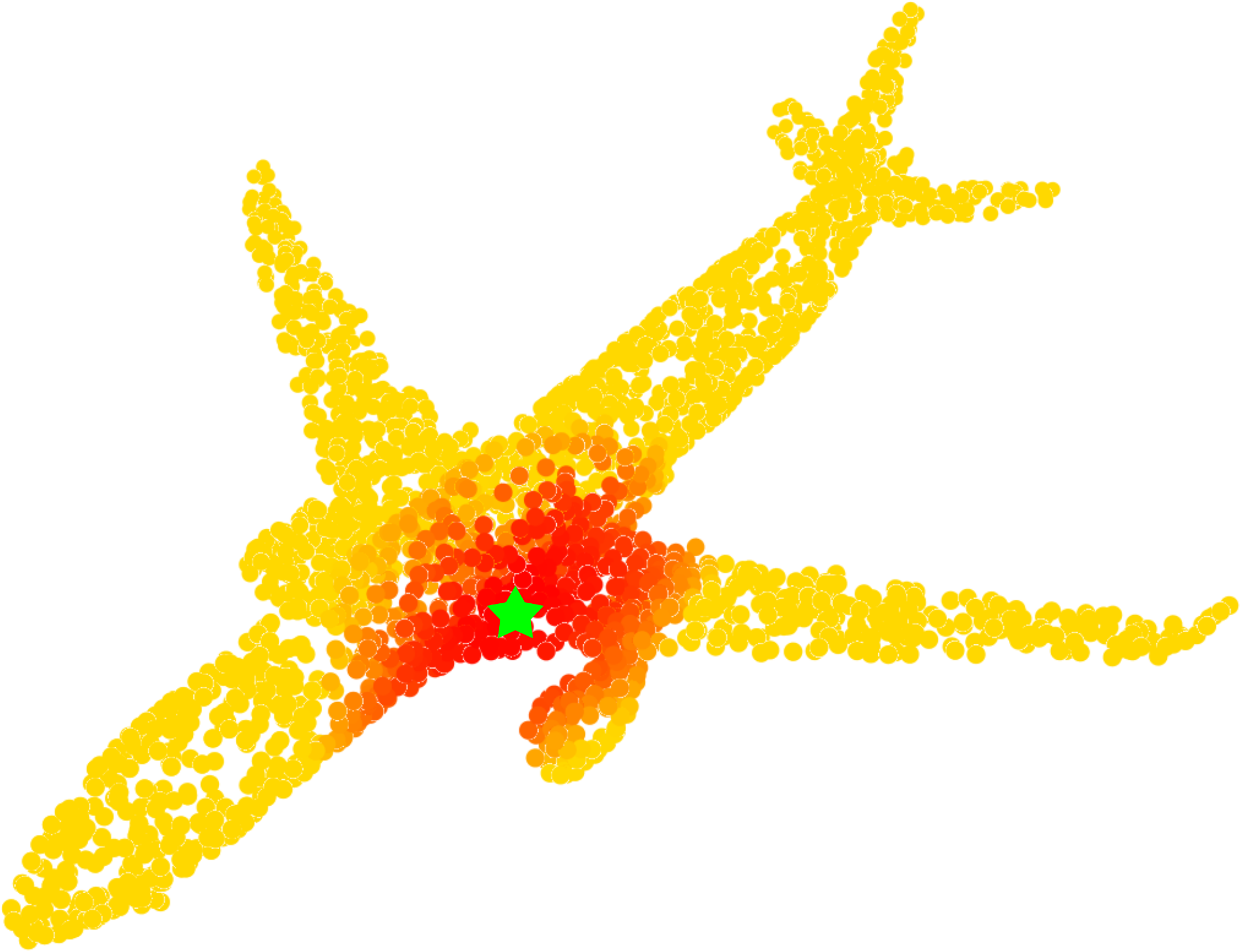}}%
	\subfigure[Layer 1]{\includegraphics[width=\unitx]{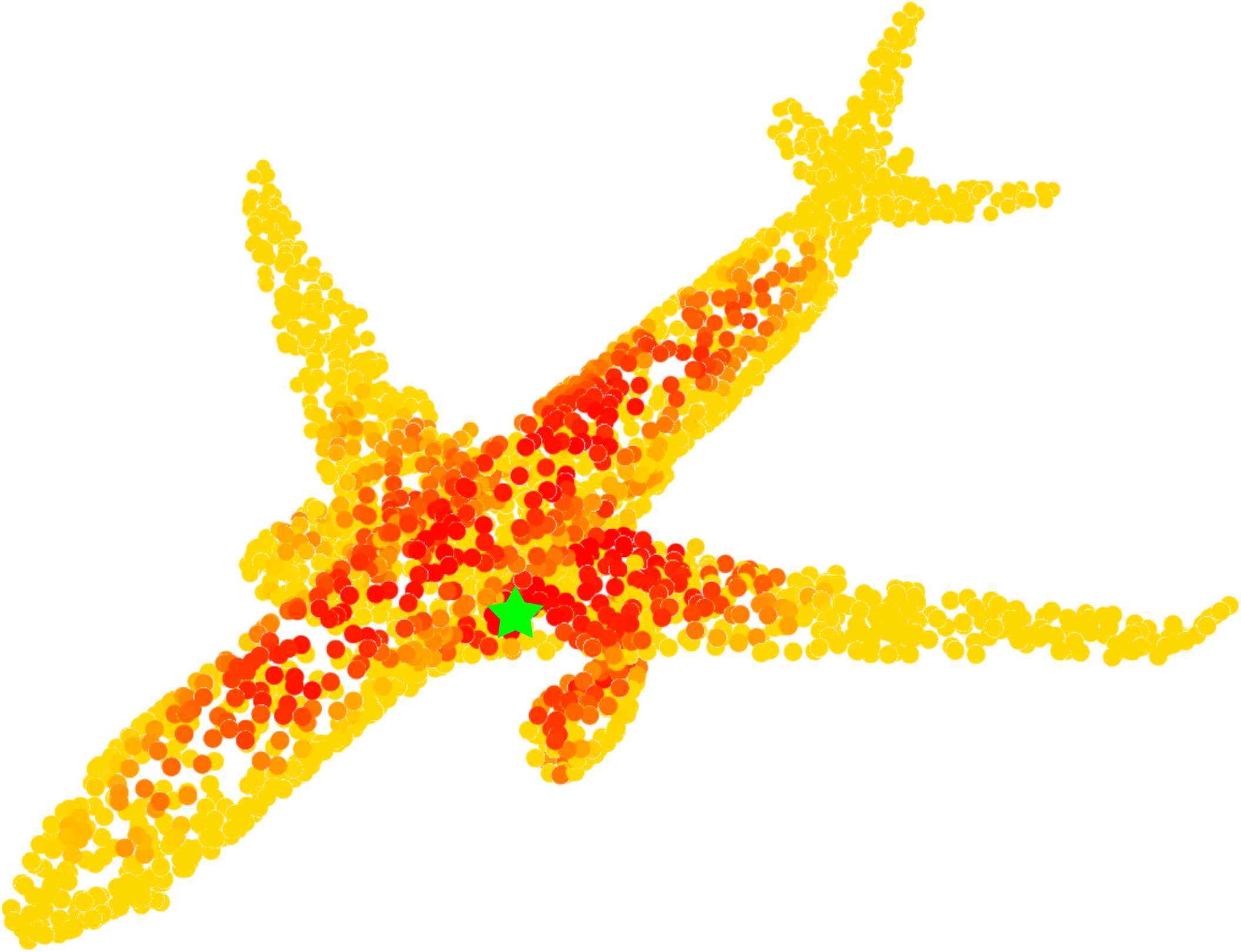}}%
	\subfigure[Layer 2]{\includegraphics[width=\unitx]{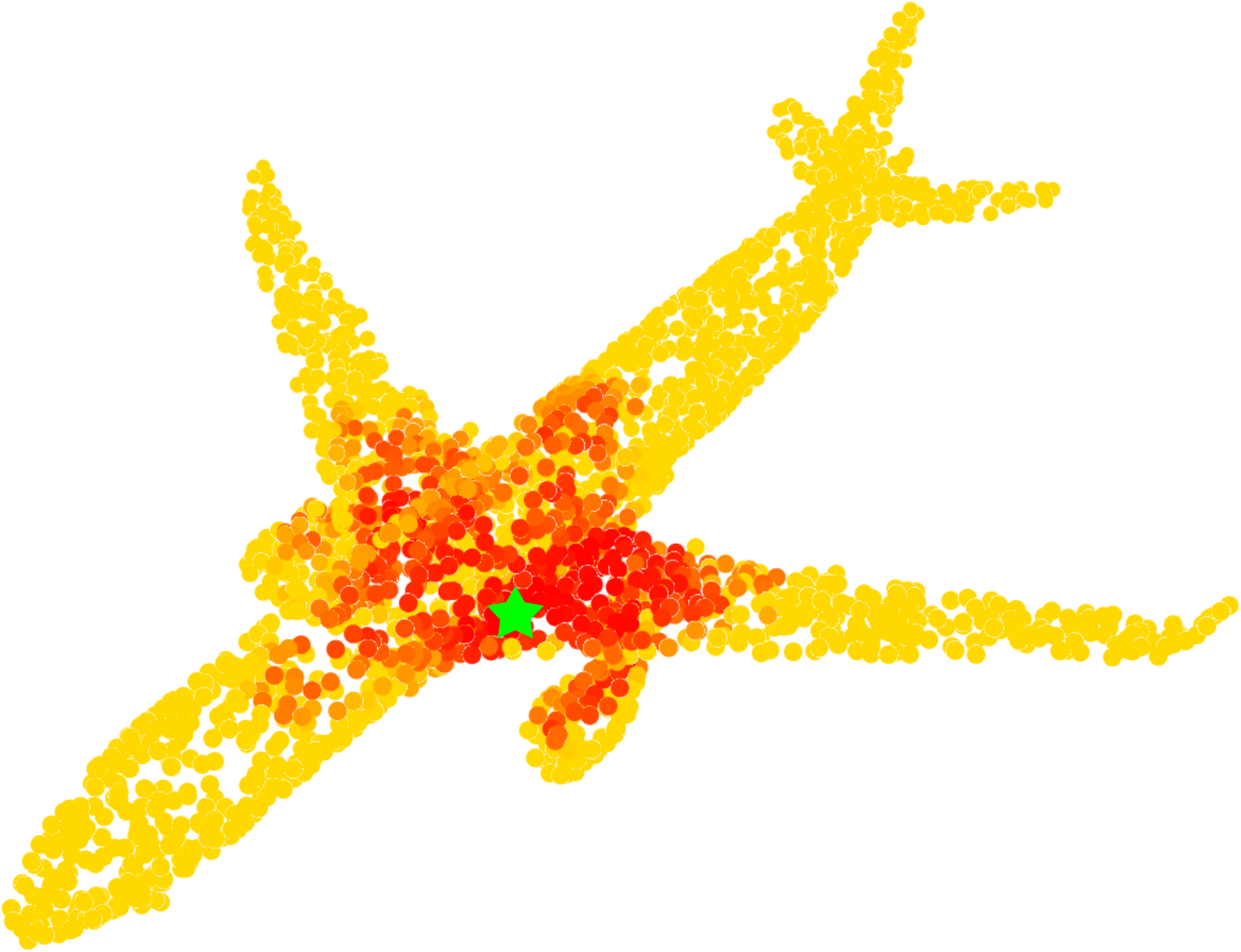}}%
	\subfigure[Layer 3]{\includegraphics[width=\unitx]{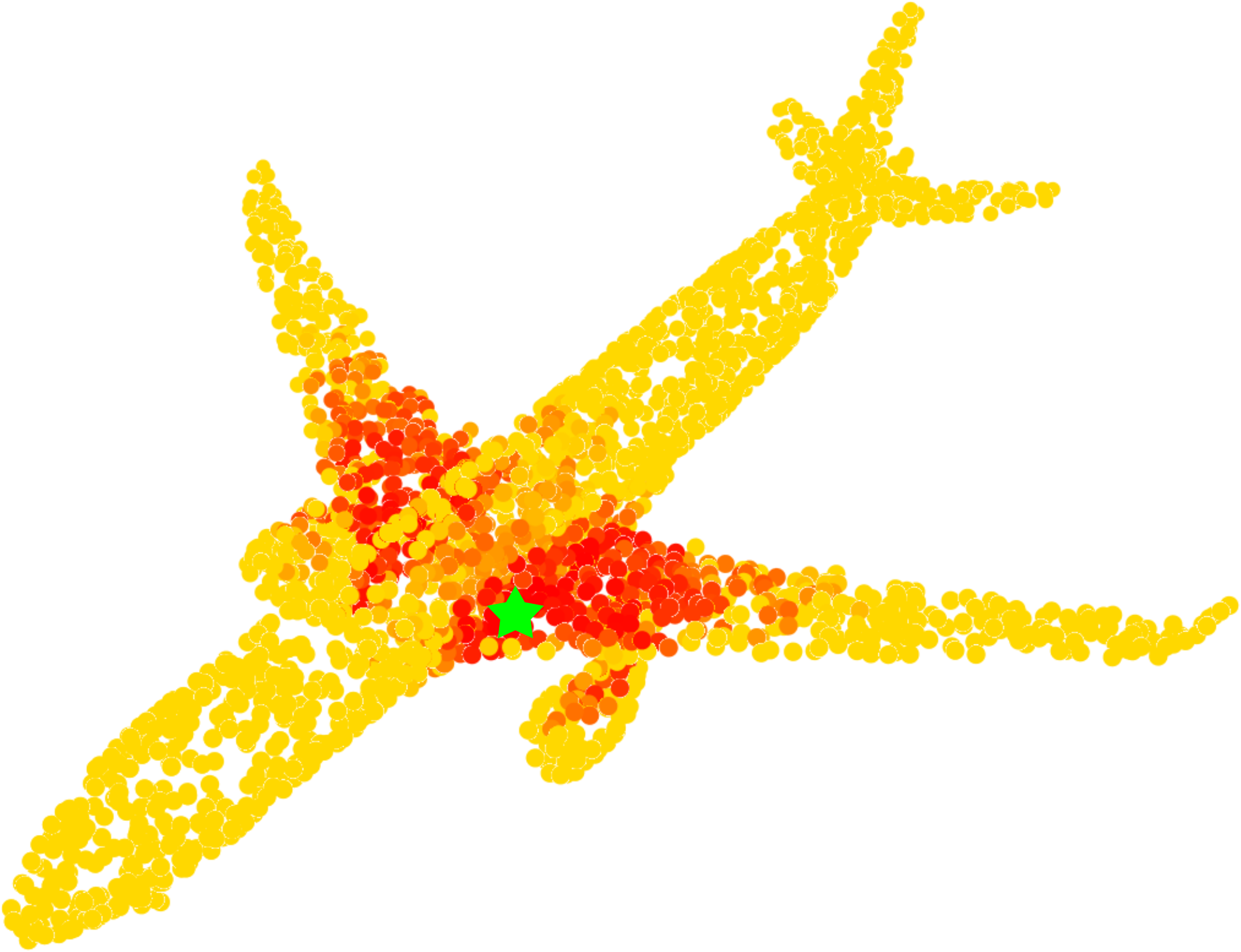}}%
	\subfigure[Layer 4]{\includegraphics[width=\unitx]{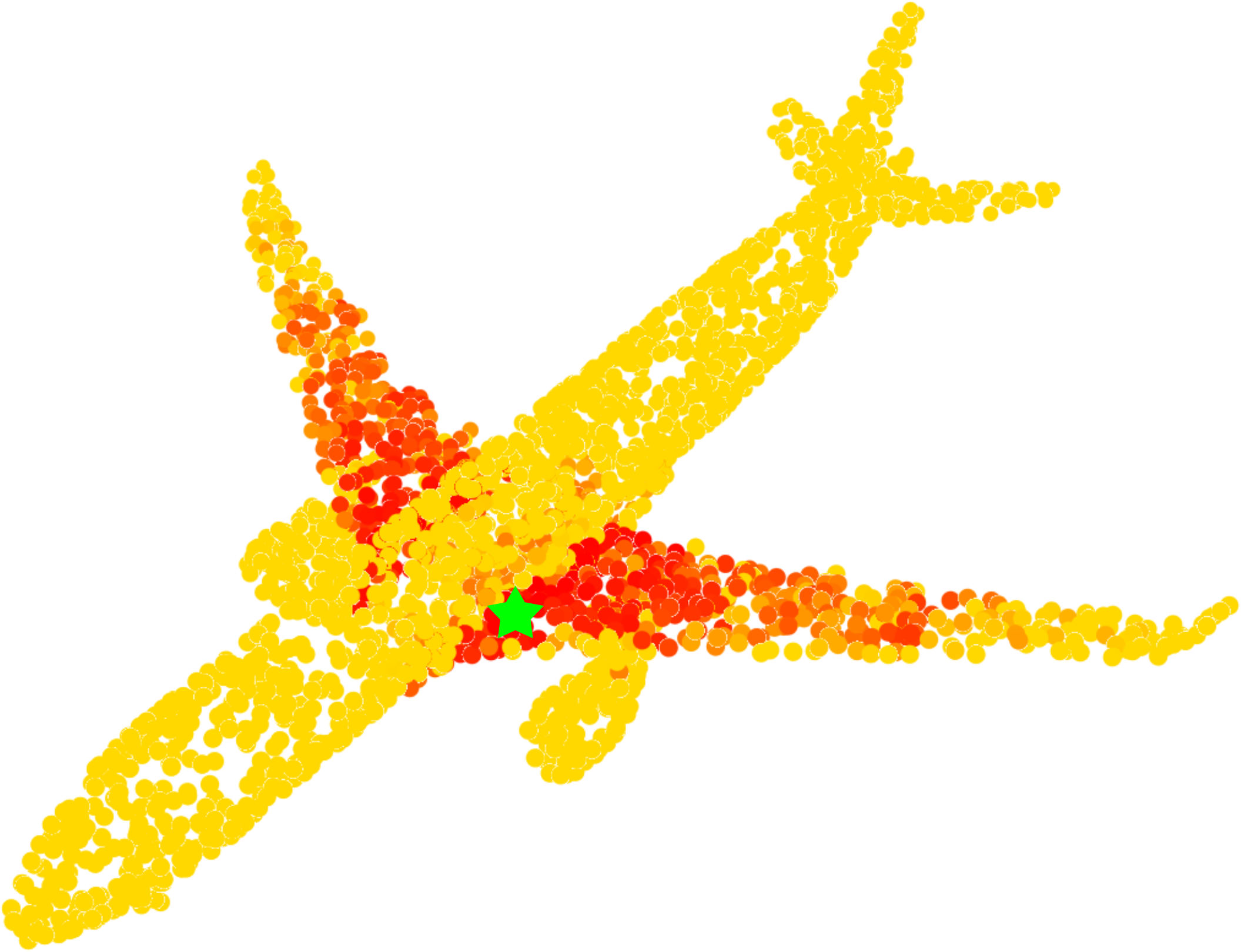}}%
	\subfigure[Ours]{\includegraphics[width=\unitx]{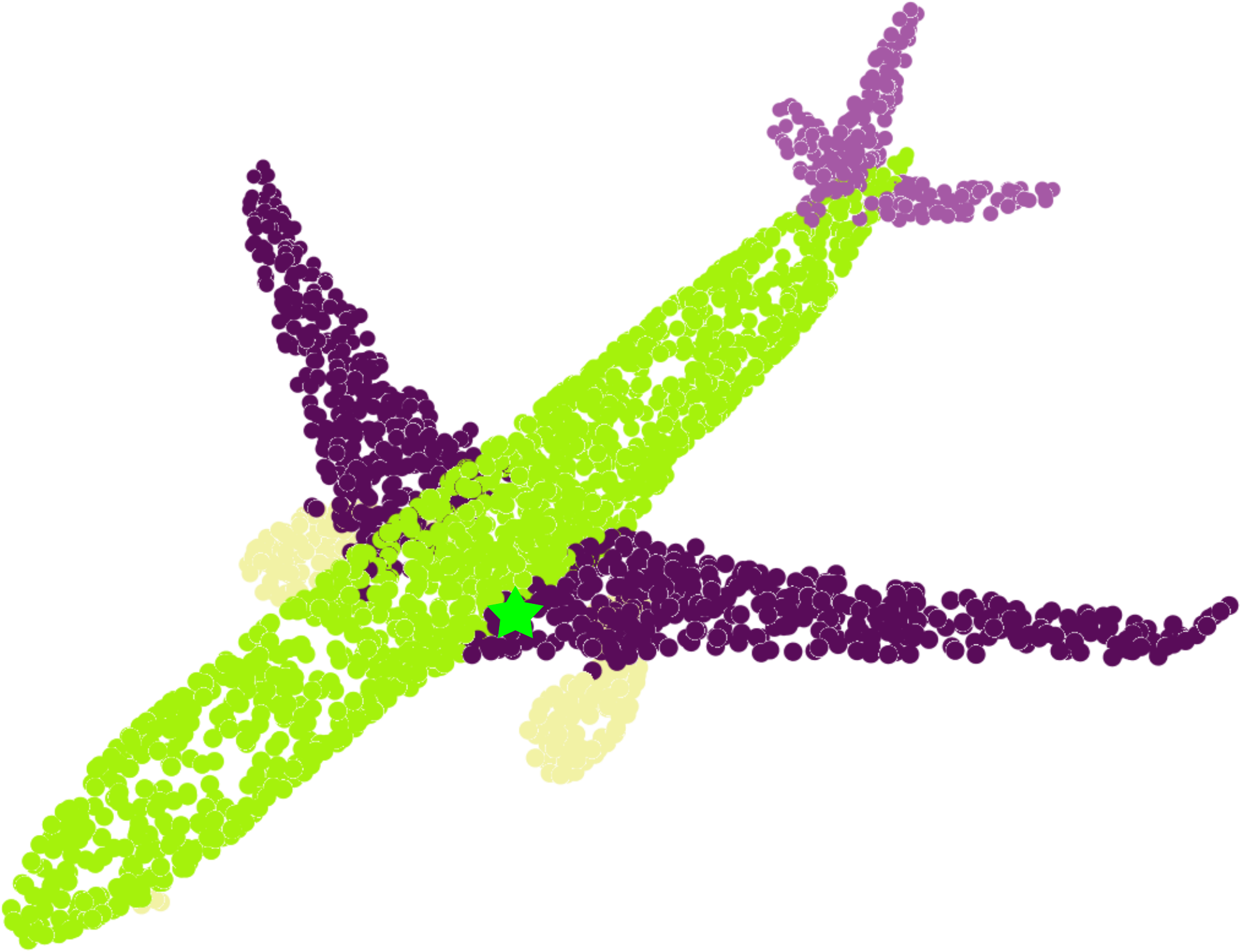}}%
	\subfigure[GT]{\includegraphics[width=\unitx]{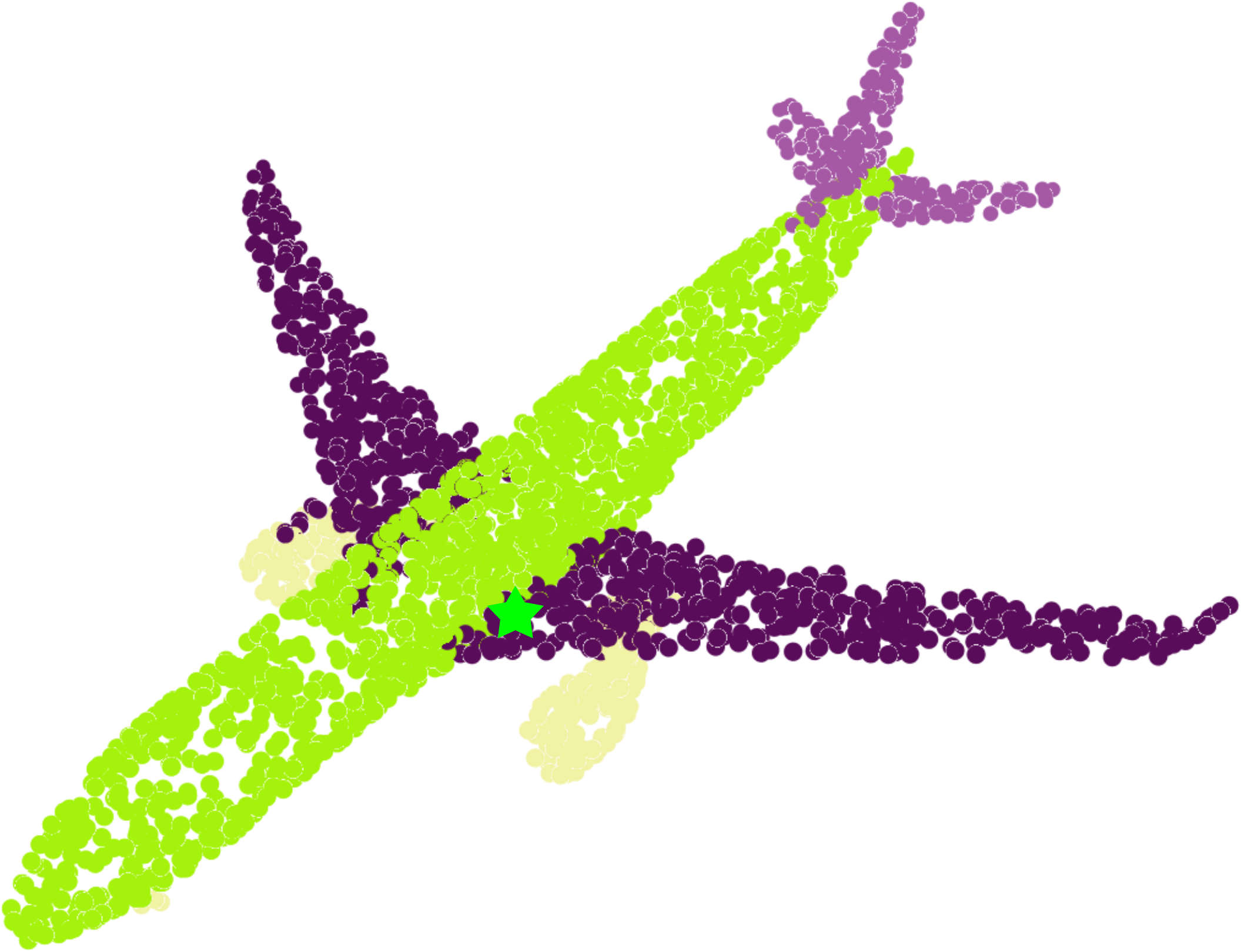}}
	
	\caption{Visualize the Euclidean distances between two points (blue and green stars) and other points in the feature space (red: near, yellow: far).  }
	\label{fig:insight1}
\end{figure*}

\subsection{Robustness test}
We further evaluate the robustness of our model to point cloud density and noise perturbation on ModelNet40 \cite{wu20153d}. We compare our AGConv with several other graph convolutions as discussed in Sec.~\ref{sec:eval:ablation}. All the networks are trained with 1k points and the neighborhood size is set to $k=20$. In order to test the influence of point cloud density, a series of numbers of points are randomly dropped out during testing. For noise test, we introduce additional Gaussian noise with standard deviations according to the point cloud radius. From Fig.~\ref{fig:robust}, we can see that our method is robust to missing data and noise, thanks to the adaptive kernel in which the structural connections are extracted dynamically in a sparser area.

Also, we experiment the influence of $k$ of the nearest neighboring points in Tab.~\ref{table:numberk}. We choose several typical sizes for testing. Reducing the number of neighboring points leads to smaller computational costs while the performance will degenerate due to the limitation of receptive fields. Our network still achieves a promising result when $k$ is reduced to 5. Meanwhile, with certain point density, a larger $k$ does not improve the performance since the local information dilutes within a larger neighborhood.

\begin{table}[t]
	\centering
	\small
	\setlength{\tabcolsep}{3.5mm}
	\begin{tabular}{c|ccc} 
		\toprule[1pt]
		Method & \#parameters & FLOPs & OA(\%) \\
		\midrule[0.3pt]
		\midrule[0.3pt]
		PointNet \cite{QiSMG17}							& 3.5M & 878M & 89.2 \\
		PointNet++ \cite{qi2017pointnet++}				& \textbf{1.48M} & 1.69G & 91.9 \\
		DGCNN \cite{wang2019dynamic}					& 1.81M & 2.57G & 92.9 \\
		KPConv \cite{thomas2019kpconv}					& 14.3M & \textbf{200M} & 92.9 \\
		Ours											& 1.85M & 3.56G & \textbf{93.4} \\
		\bottomrule[1pt]
	\end{tabular}
	\vspace{5pt}
	\caption{The number of parameters, floating point operations and overall accuracy of different models.}
	\label{table:complexity}
\end{table}

\begin{table}[t]
	\centering
	\footnotesize
	\begin{tabular}{c|ccc} 
		\toprule[1pt]
		Method & \#parameters & Time(ms) & OA(\%) \\
		\midrule[0.3pt]
		\midrule[0.3pt]
		Baseline (w/o AGConv)			& \textbf{1.81M} & \textbf{93.1} & 92.5 \\
		AGConv (2-Layer)					& 1.85M & 129.1 & \textbf{93.4} \\
		AGConv (3-Layer) 					& 1.95M & 168.4 & 93.0 \\
		AGConv (4-Layer) 					& 2.35M & 276.0 & 93.2\\
		\bottomrule[1pt]
	\end{tabular}
	\vspace{5pt}
	\caption{Number of parameters, forward pass time (per batch) and overall accuracy for different models using AGConv.}
	\label{table:modelsize}
\end{table}

\begin{figure}
	\centering
	\newlength{\unitS}
	\setlength{\unitS}{0.15\linewidth}
	
	\includegraphics[width=0.11\linewidth]{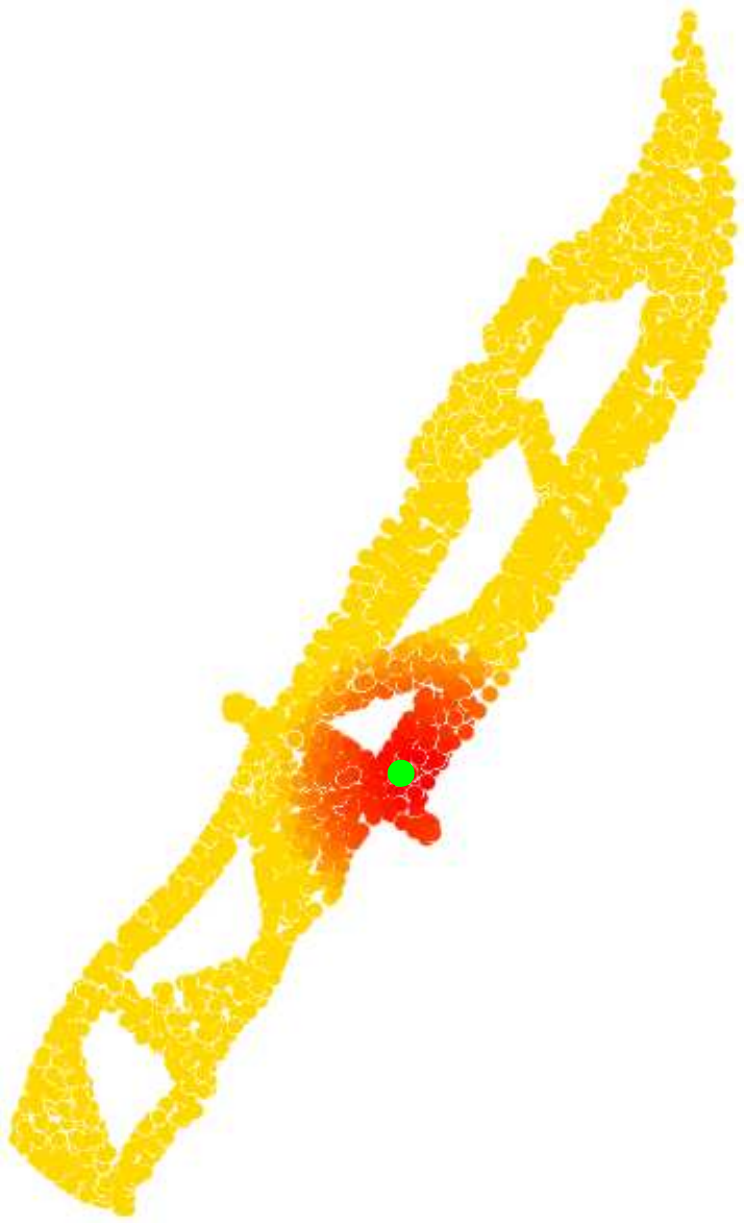}%
	\hspace{16pt}\includegraphics[width=0.11\linewidth]{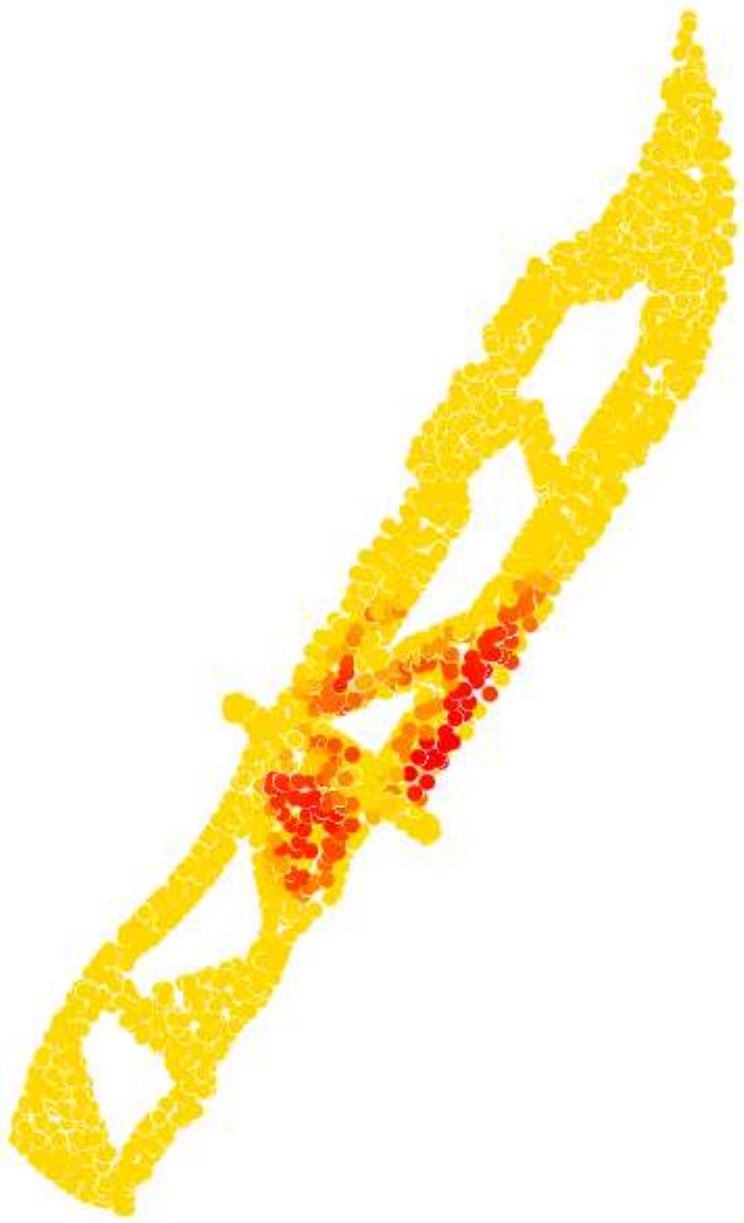}%
	\hspace{16pt}\includegraphics[width=0.11\linewidth]{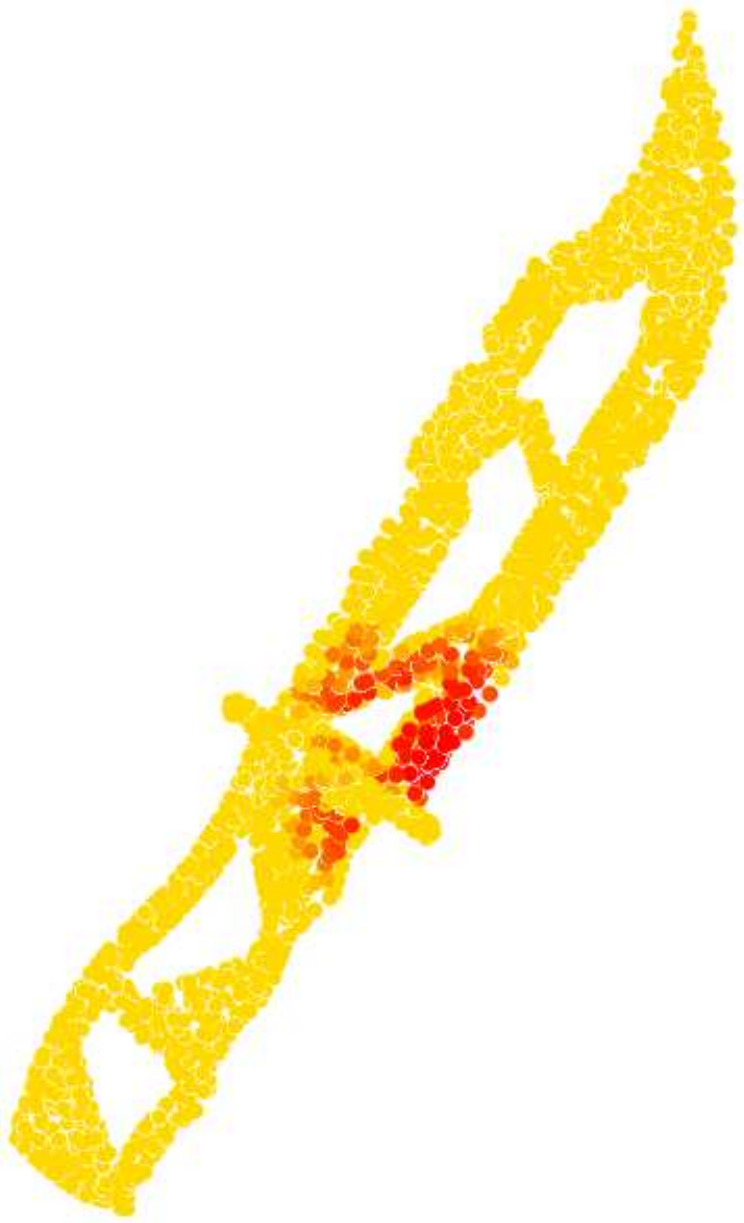}%
	\hspace{16pt}\includegraphics[width=0.11\linewidth]{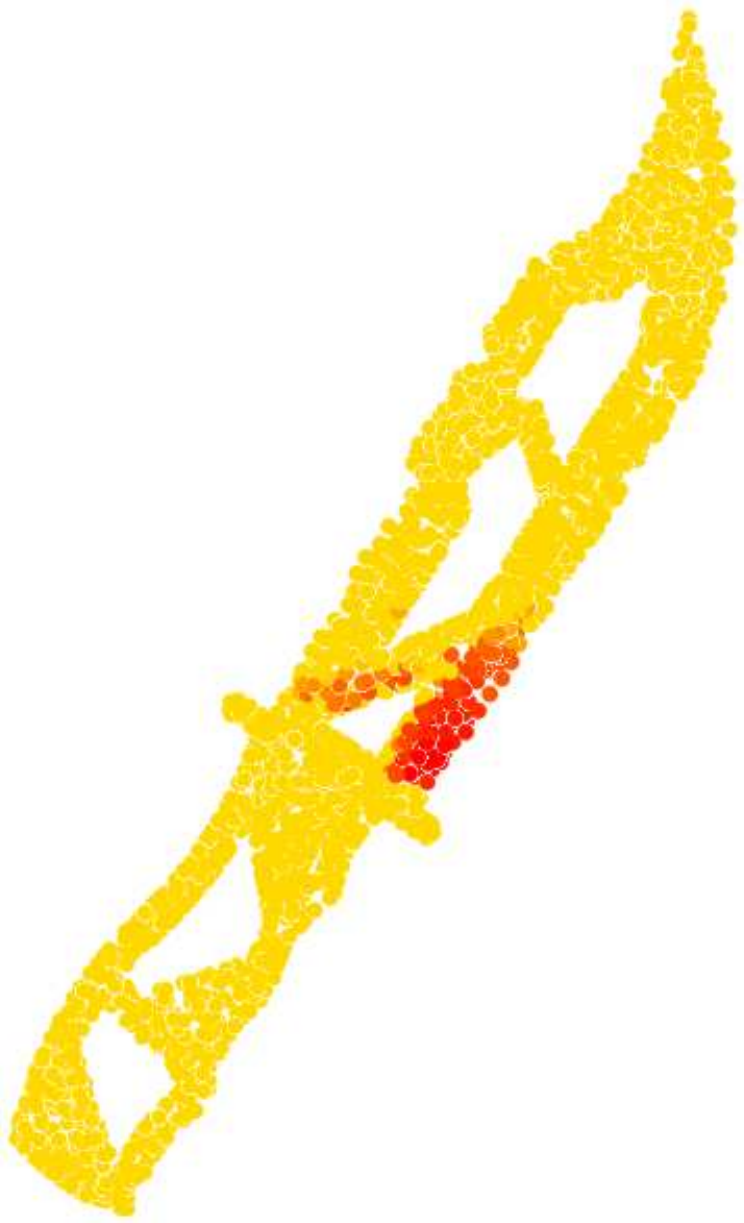}%
	\hspace{16pt}\includegraphics[width=0.11\linewidth]{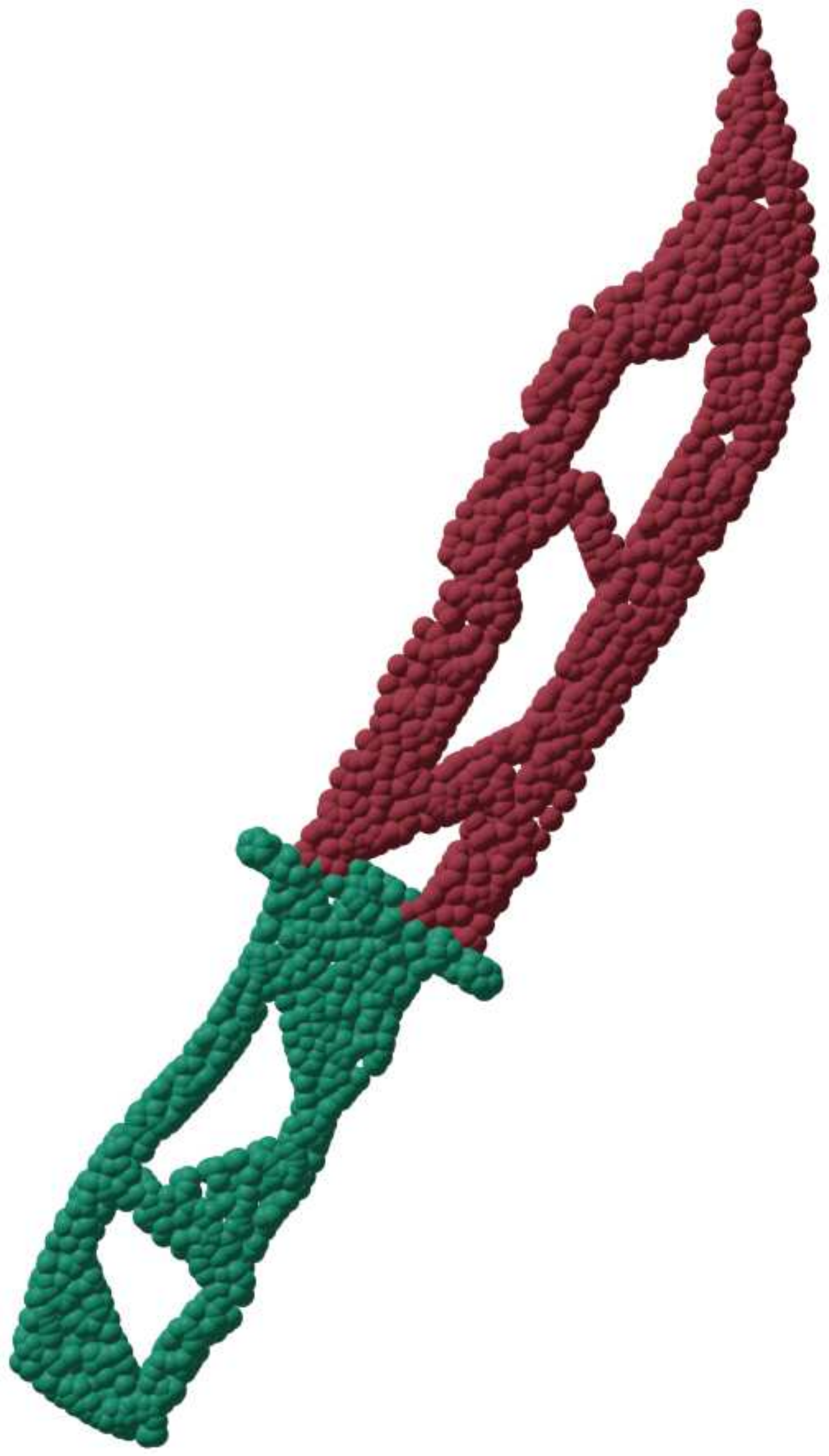}%
	\hspace{16pt}\includegraphics[width=0.11\linewidth]{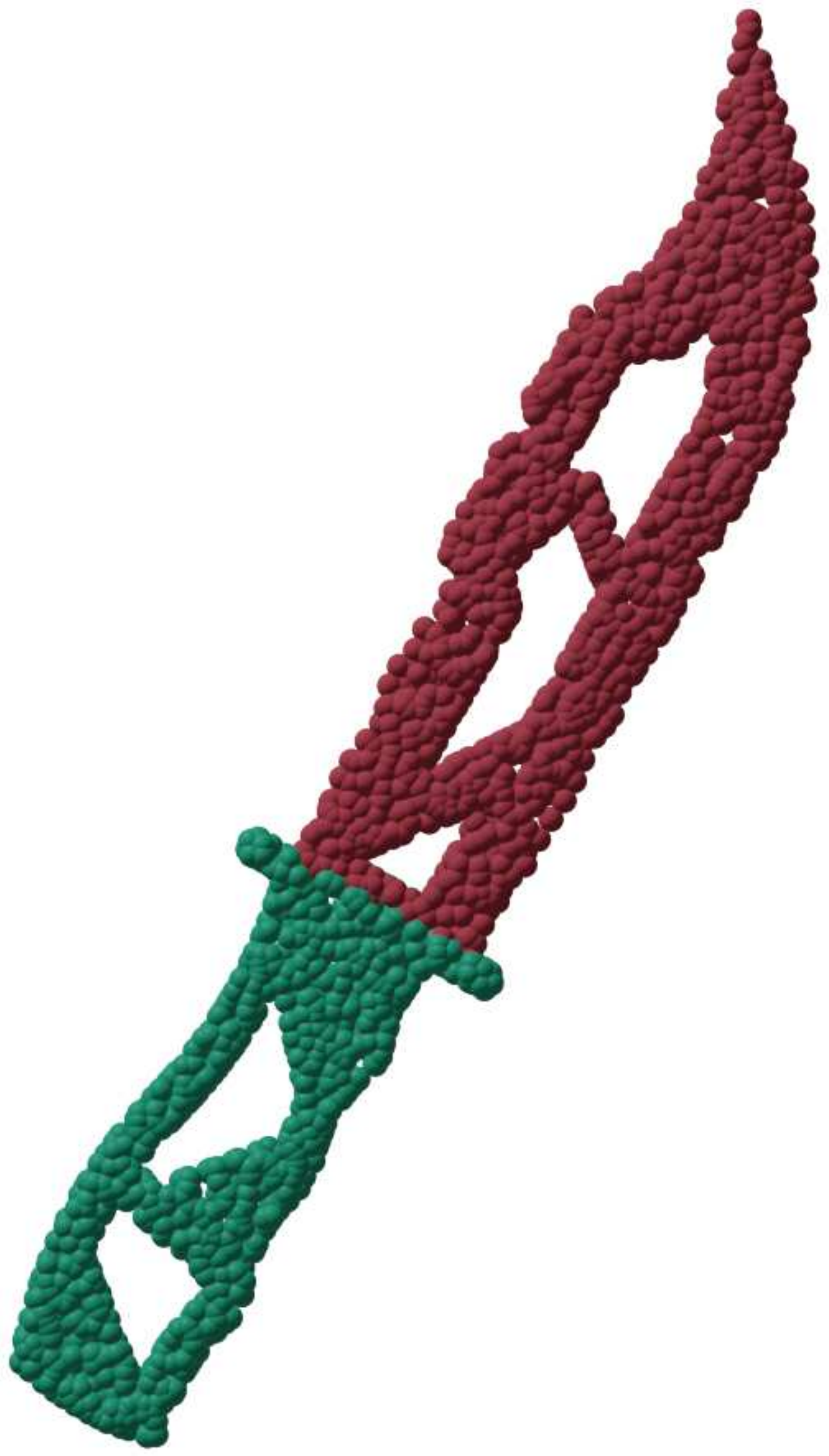}
	
	\subfigure[Spatial]{\label{fig:insight:spatial} \includegraphics[width=\unitS]{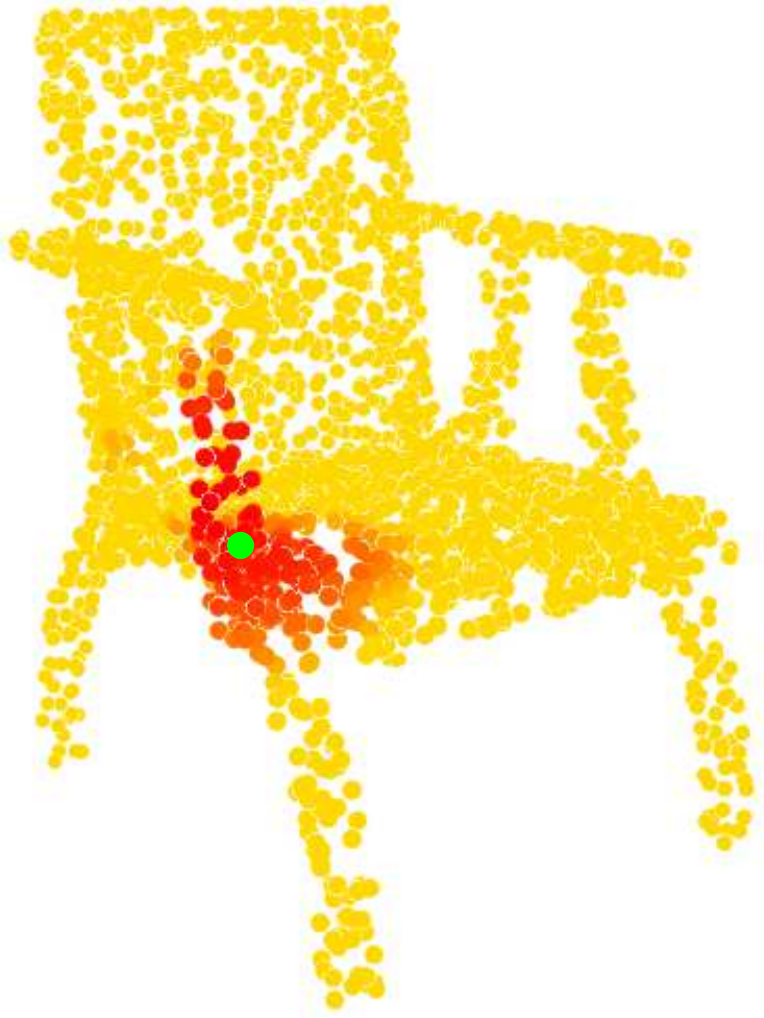}}%
	\subfigure[Layer 1]{\includegraphics[width=\unitS]{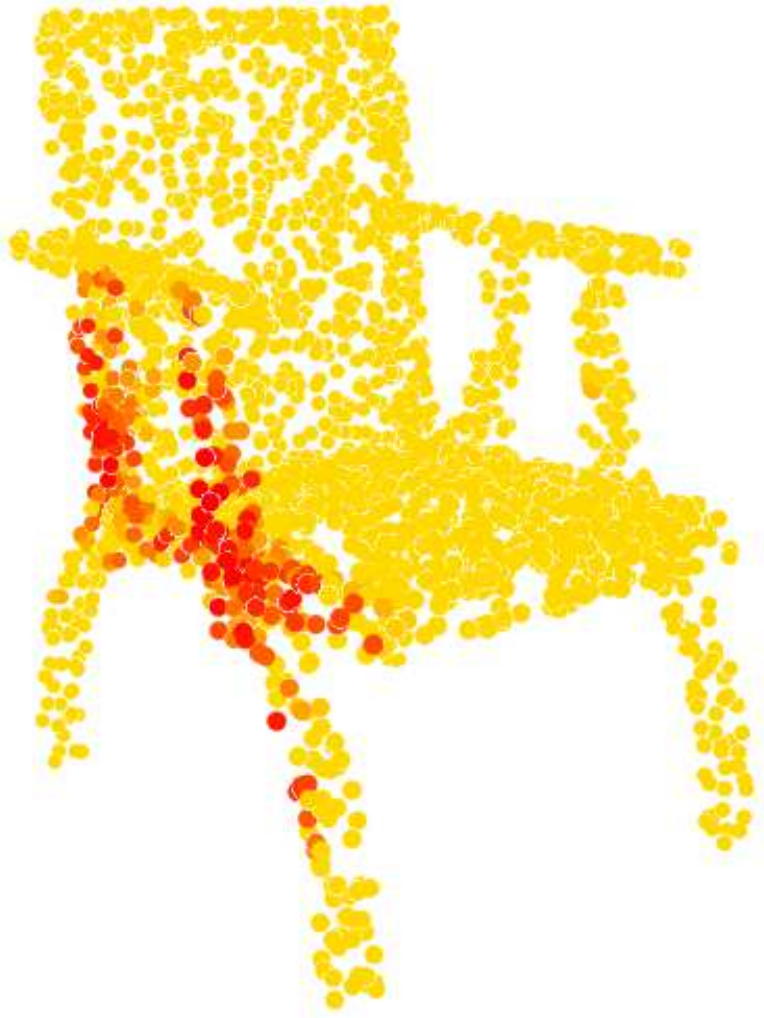}}%
	\subfigure[Layer 2]{\includegraphics[width=\unitS]{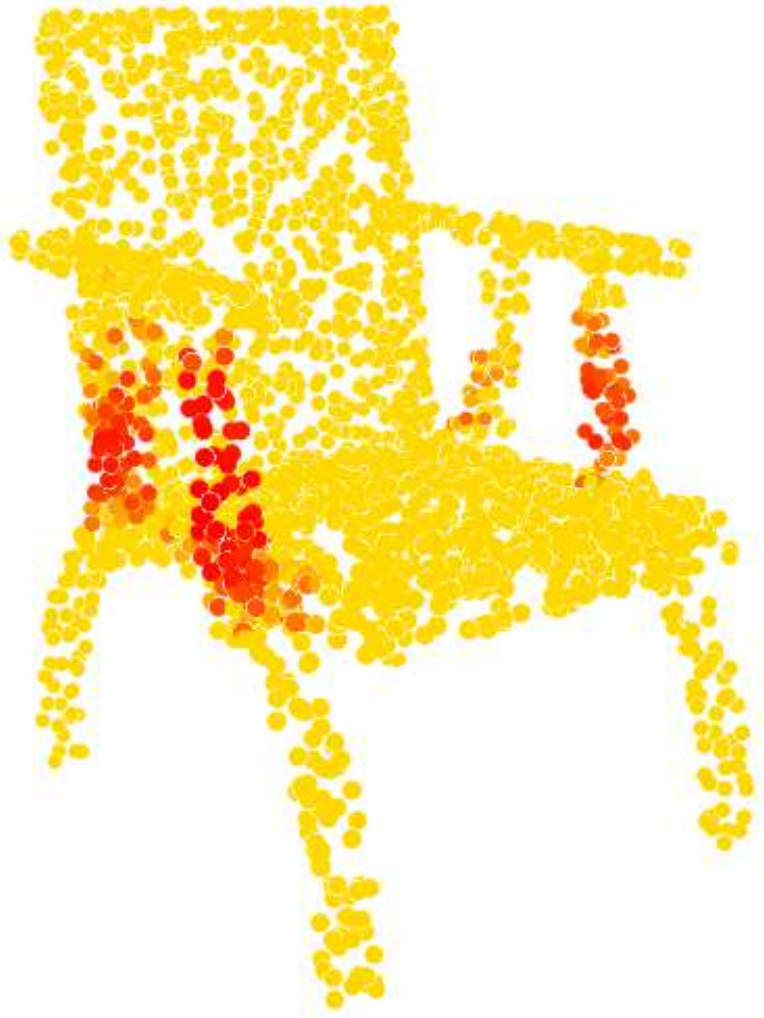}}%
	\subfigure[Layer 4]{\includegraphics[width=\unitS]{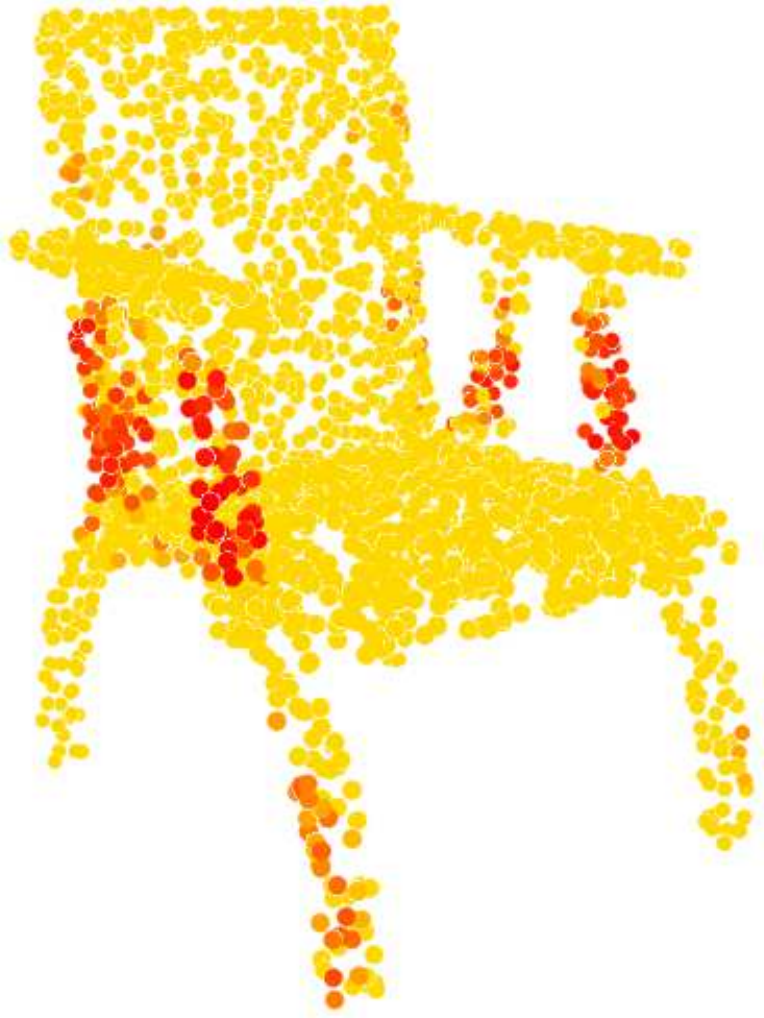}}%
	\subfigure[Ours]{\includegraphics[width=\unitS]{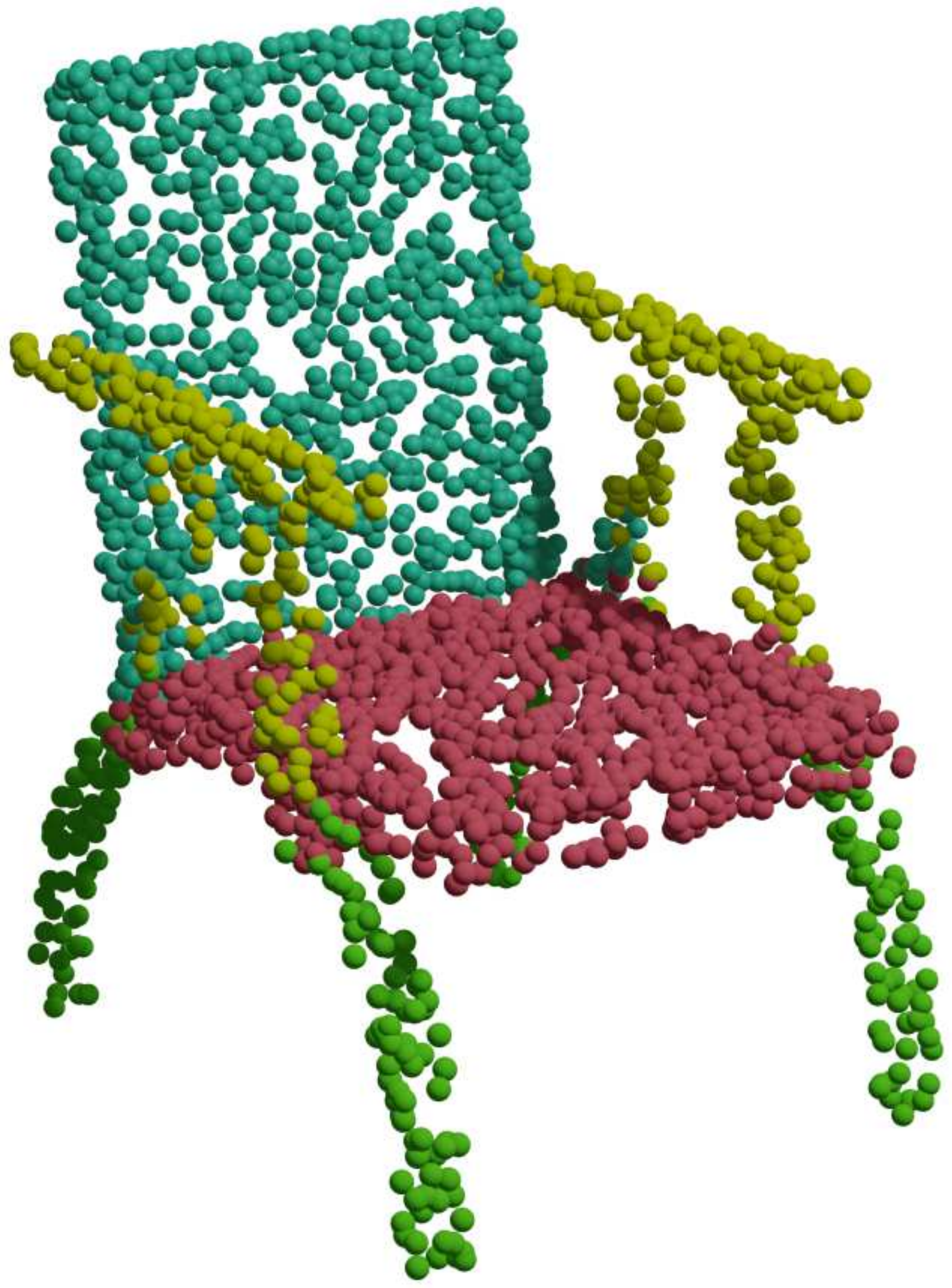}}%
	\subfigure[GT]{\includegraphics[width=\unitS]{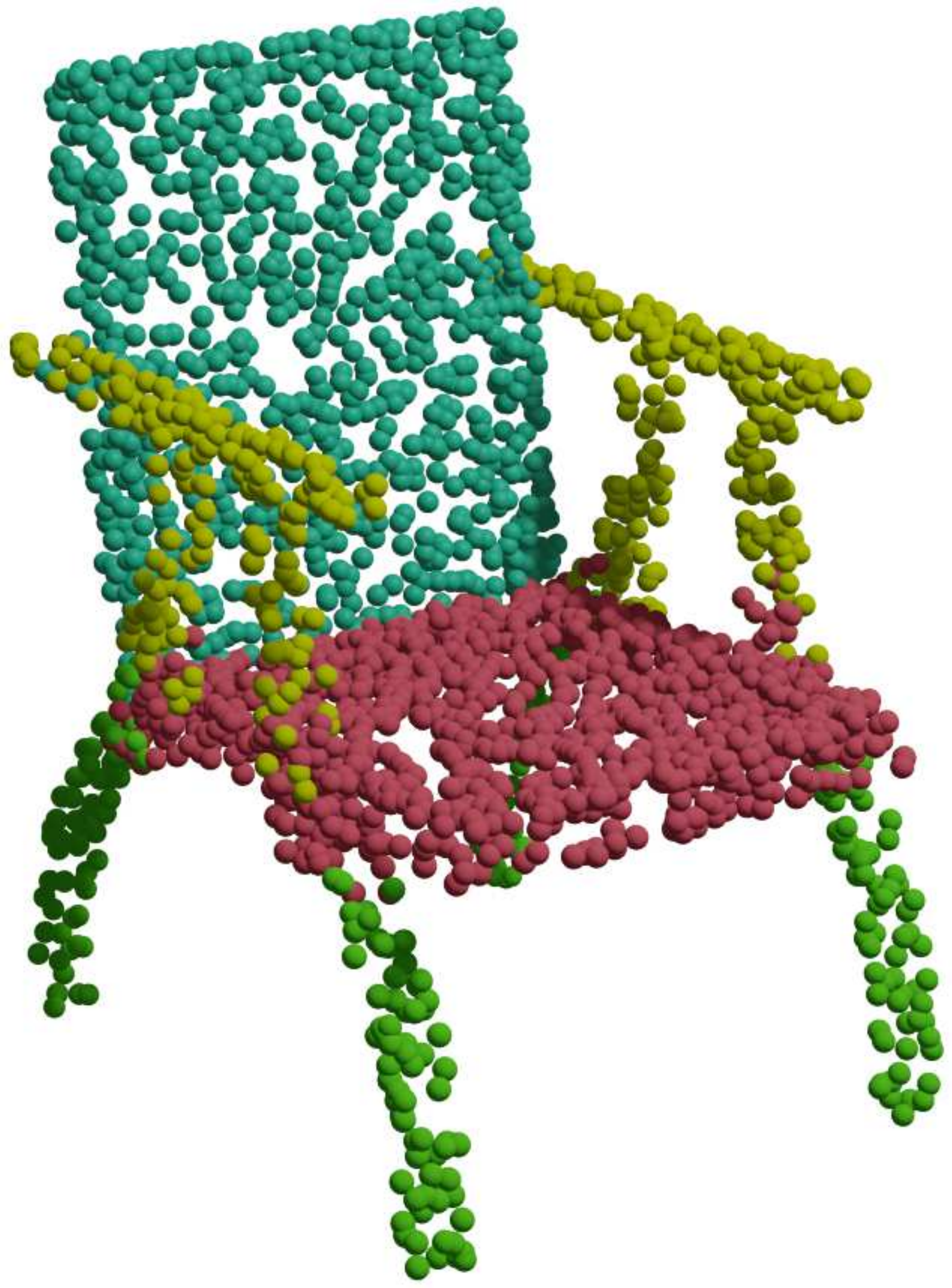}}
	
	\caption{Visualize the Euclidean distances between the target point (green point in (a)) and other points in the feature space. Red color denotes a closer point and yellow one is far from the target. The feature distances in several layers  provide a clear insight that our network can distinguish points belonging to different semantic parts. It also captures non-local similar structures (see the handles in the second row).}
	\label{fig:insight}
\end{figure}

\subsection{Efficiency}
\label{sec:eval:eff}
To compare the complexity of our model with the state-of-the-arts, we show the parameter numbers, floating point operations (FLOPs), and the corresponding results of networks in Tab.~\ref{table:complexity}. Floating point operations are tested on 1024 points. These models are based on ModelNet40 for classification. Our model achieves the best performance of 93.4\% overall accuracy and the model size is relatively small. Compared with DGCNN \cite{wang2019dynamic} which is a standard graph convolution version in our ablation studies, the proposed adaptive kernel performs better while being efficient.

\subsection{Model complexity}
The standard graph convolution in this work contains $2DM$ parameters ($2D$ denotes the dimension of feature input $\Delta f_{ij}$). Here, $D$ and $M$ denote the input and output dimensions respectively. As described in Sec.~\ref{sec:method:architecture}, the kernel function uses a two-layer MLP which contains $2dD + dcM$ parameters where $c$ is the dimension of $\Delta x_{ij}$ ($c = 3\times2$ for point coordinates $(\mathbf{x},\mathbf{y},\mathbf{z})$ input). $d$ is the dimension of a hidden layer of the kernel function (see Fig.~\ref{fig:kernel2}) and it can be adjusted to reduce the model size. We design the network architecture with two layers of AGConv which achieves pleasing performance. We further report the results and parameter numbers on ModelNet40 using different numbers of AGConv layers in Tab.~\ref{table:modelsize}. The baseline uses standard graph convolutions, which is a similar version of DGCNN. The adopted design (2-Layers) significantly improves the network performance while the model size is relatively small. Please note that the performance of a network may be not enhanced by adding more AGConv layers (e.g., adding 3 Layers or 4 Layers), since a two-layer AGConv module is often sufficient to capture geometrically similar neighbors for a low-density dataset such as ModelNet40 in Tab. 10. In practice, we encourage users to add some more AGConv layers if the input data is of high density. For the time performance evaluation, we also report the forward pass times of different models. The proposed AGConv layer is able to improve the performance of existing graph CNNs while being efficient.

\section{Visualization and learned features}
We provide visual results to further demonstrate the effectiveness of AGConv over fixed-kernel methods. We first visualize the segmentation results on ShapeNetPart in Fig.~\ref{fig:partseg}. In this experiment, we compare the results of DGCNN \cite{wang2019dynamic}, attentional graph convolution (Attention Point described in Sec.~\ref{sec:eval:ablation}) and AGConv. Our results are better in challenging regions, such as part boundaries and object edges. This verifies that our method is able to capture distinguishable features for points belonging to different parts. 

To achieve a deeper understanding of AGConv, we explore the feature relations in several intermediate layers of the network to see how AGConv can distinguish points with similar spatial inputs. In this experiment, we train our model on ShapeNetPart for segmentation. In Fig.~\ref{fig:insight1}, two target points (blue and green stars in 1-st and 2-nd rows respectively) are selected which belong to different parts of the object. We then compute the Euclidean distances to other points in the feature space, and visualize them by coloring the points with similar learned features in red. We can see that, while being spatially close, our network can capture their different geometric characteristics and segment them properly. Also, from the 2-nd row of Fig.~\ref{fig:insight1}, points belonging to the same semantic part (the wings) share similar features while they may not be spatially close. This shows that our model can extract valuable information in a non-local manner. As shown in Fig.~\ref{fig:insight}, when the point is close to edges between different semantic parts, our network encourages it to have distinguishable features which captures better geometric information. Thus, it is separated from other parts of the objects, as shown in the first row of Fig.~\ref{fig:insight}. Also, we see that in the second row of Fig.~\ref{fig:insight}, points belonging to the same semantic part share similar features while they may not be spatially close. Note that, Fig.~\ref{fig:insight:spatial} indicates the spatial distances with regard to the central point.


\section{More Point Cloud Analysis Applications}
AGConv, as a plug-and-play module, can flexibly serve more point cloud analysis approaches to boost their performance. We develop AGConv-based completion, denoising, upsampling, circle extraction and registration networks in this section. The graph structure is updated in each AGConv layer according to the feature
similarity among points. For each task, we introduce the network framework, data preparation and comparison in a concise way. \textit{More details will be found in our prepared webpage.}

\subsection{Point cloud completion}
Raw point clouds captured by 3D sensors are often incomplete. When employing these untreated point clouds for semantic understanding, users may receive inaccurate or even wrong results. 
Point cloud completion aims to infer the whole underlying surface from a partial input. Moreover, the completion results should be uniform, dense and possess logically correct geometric structures.

\begin{figure}
	\centering
	\includegraphics[width=1\linewidth]{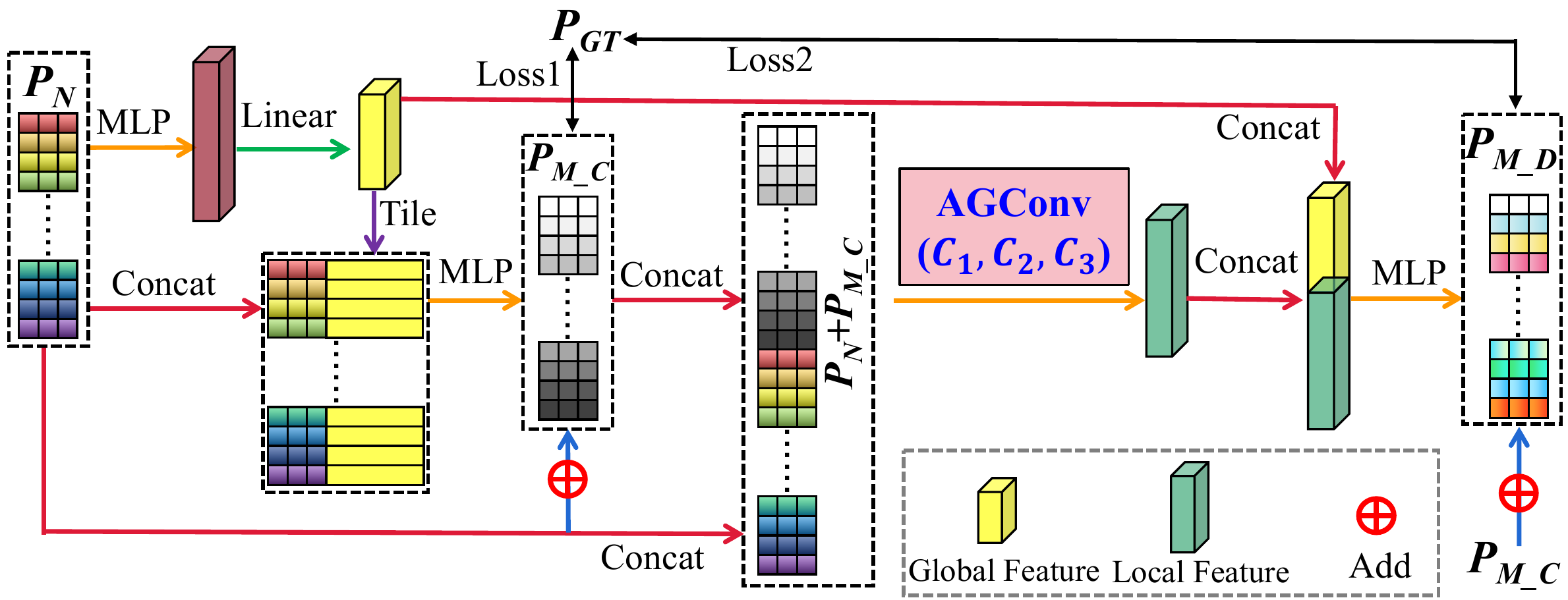}
	\caption{ECG-Net \cite{9093117} is improved by our AGConv for point cloud completion (call iECG-Net). iECG-Net can better extract local features to recover the shape details by AGConv. iECG-Net 1) first generates the coarse result $P_{M\_C}$, and 2) concatenates  $P_{M\_C}$ with the input $P_N$ to produce $P_N+P_{M\_C}$, and 3) produces the final yet fine result $P_{M\_D}$.}
	\label{CompletionNet}
\end{figure}

\vspace{5pt}\noindent\textbf{Network framework.} We leverage ECG-Net \cite{9093117} as our network's backbone in Fig.~\ref{CompletionNet}. In our improved ECG-Net (iECG-Net), the original graph convolution module is replaced by the AGConv module. iECG-Net employs the so-called coarse-to-fine strategy, i.e., first recovering its global yet coarse shape and then increasing
its local details to output the missing point cloud of input. First, we take the incomplete point cloud $P_N$ as input, and use an encoder-decoder structure like PCN \cite{yuan2018pcn} to yield the coarse point cloud $P_{M\_C}$ to represent the missing part. Then, we concatenate the input incomplete point cloud $P_N$ and the coarse missing part $P_{M\_C}$. Finally, we take the concatenation result $(P_N+P_{M\_C})$ to the following detail refinement network that contains the AGConv layer. AGConv not only extracts adequate spatial structure information from $(P_N+P_{M\_C})$, but also extracts local features more efficiently and precisely. The final completion result is $P_{M\_D}$ that represents the missing part. Thus, the whole complete point cloud is  $(P_N+P_{M\_D})$. 
Please note that, in training, we calculate two losses from $P_{M\_C}$ and $P_{M\_D}$ with the ground truth $P_{GT}$ by the Chamfer distance.

\vspace{5pt}\noindent\textbf{Data.} We train and evaluate our iECG-Net in the benchmark dataset Shapenet-Part, which has 13 categories of different objects. Shapenet-Part has 14473 shape elements formatted by point clouds, in which 11705 point clouds are for training and 2768 for testing. In Shapenet-Part, the centers of all point clouds locate at the origin, and their coordinate values of xyz range within [-1,1]. We sample 2048 points uniformly from each point cloud as the complete shape. Then, we select some border points like (1,1,1) or (-1,1,1) as viewpoints and randomly choose a viewpoint for the point clouds in a same training batch, and remove a certain amount of points that are closest to the viewpoint. By such removal operation, we can produce the incomplete point clouds for training and testing. In our experiments, we aim at the problem of large-ratio incomplete point cloud completion. Thus, we set the ratio to be to 50\% for both training and testing.

\begin{figure}
	\centering
	\includegraphics[width=1\linewidth]{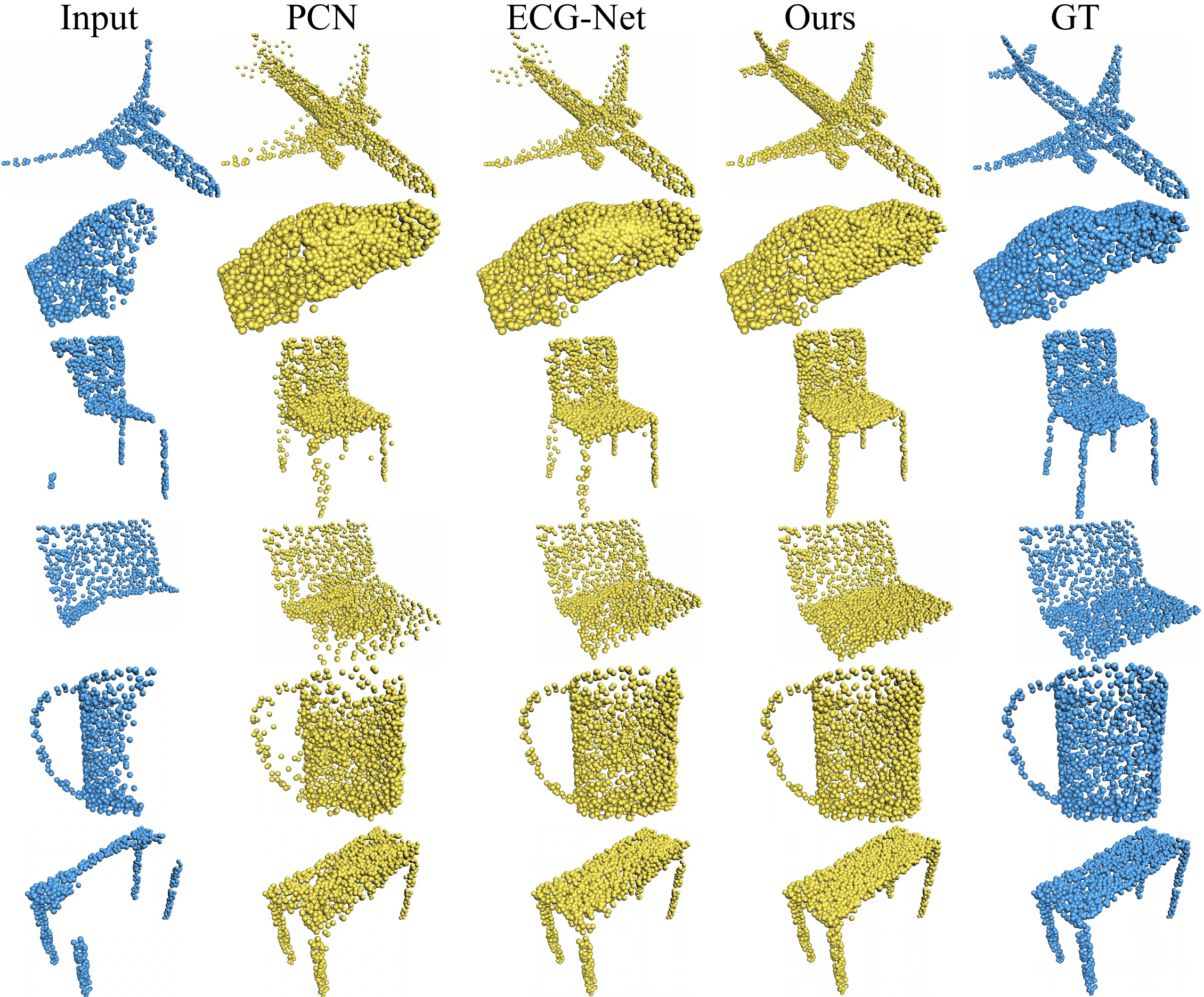}
	
	\caption{Visualization of completion results on Shapenet-Part.}
	\label{completion_f1}
\end{figure}

\vspace{5pt}\noindent\textbf{Comparison.} To demonstrate the effectiveness of AGConv, we compare iECG-Net against several representative completion methods, including L-GAN \cite{lin2018learning}, PCN \cite{yuan2018pcn}, 3D-Capsual \cite{zhao20193d}, TopNet \cite{tchapmi2019topnet}, MSN \cite{liu2020morphing}, PF-Net \cite{huang2020pf}, and ECG-Net \cite{9093117}. In our experiments, we train all the methods without the category information for fairness. We test all methods in 13 categories in Tab.~\ref{completion_t1}. Benefiting from AGConv, iECG-Net possesses higher average completion precision than its competitors in most cases. Moreover, the visualization results are given in Fig.~\ref{completion_f1}, where our iECG-Net can better complete the large-missing parts.

\subsection{Point cloud denoising}

\begin{figure}
	\centering
	\includegraphics[width=1\linewidth]{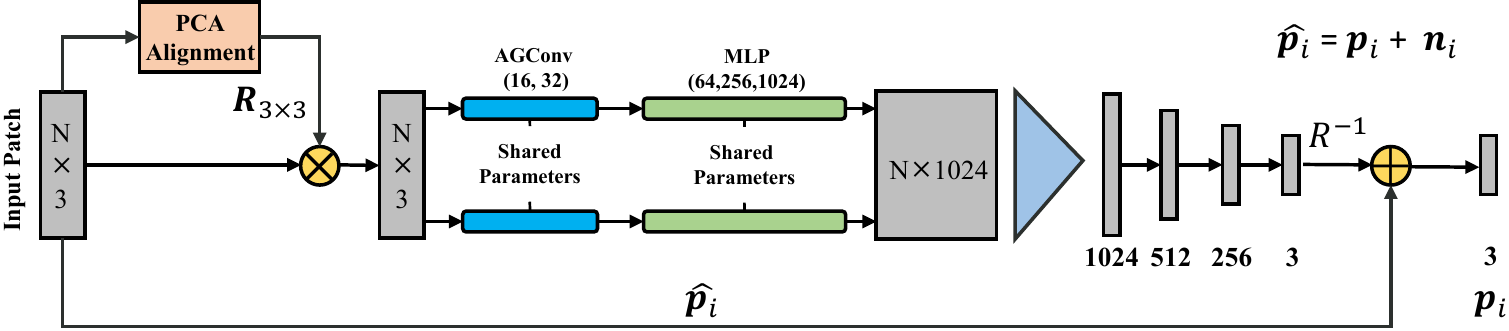}
	\caption{Pointfilter \cite{9207844} is improved by our AGConv for point cloud denoising (call iPointfilter). The main idea of iPointfilter is to project each noisy point onto the clean surface according to its neighboring structure. Given a noisy patch with $N$ points, PCA is utilized for alignment and then the aligned patch is fed into iPointfilter. Both AGConv and MLP are used to extract features, and then all the features are aggregated by max pooling. Finally, three fully connected layers are used to regress a displacement vector between the noisy point cloud and the ground truth. The outputs of the first two layers are processed by Batch Normalization and Relu, and the last layer only uses the tanh activation function to constrain the displacement vector. }
	\label{fig:DenoiseNet}
\end{figure}

3D imaging devices are frequently used to
capture the virtual models of physical objects. These models represented by point clouds are usually noisy due to measurement and reconstruction errors, and should
be denoised to facilitate subsequent applications.
Point cloud denoising aims to eliminate noise or spurious information from a noisy point cloud, while preserving its real geometry.

\vspace{5pt}\noindent\textbf{Network framework.} We leverage Pointfilter \cite{9207844} as our network's backbone in Fig.~\ref{fig:DenoiseNet}. In our improved Pointfilter (iPointfilter), the original encoder is replaced by the AGConv module. iPointfilter takes the noisy point as input, and outputs a displacement vector to move this noisy point to the underlying (noise-free) surface. 
The encoder attempts to obtain a complex representation for the input patch, and is mainly composed of two parts, i.e., a feature extractor to obtain features of different scales from the neighborhood, and a collector to aggregate the features as a latent vector. 
The extractor and collector are implemented by our AGConv layer and max pooling layer, respectively. 
The decoder is used as a regressor to return a displacement vector of the noisy point, which is realized by three fully connected layers.  

\vspace{5pt}\noindent\textbf{Data.} We train our iPointfilter on the benchmark dataset from Pointfilter \cite{9207844}, which contains 22 clean models (11 CAD models and 11 non-CAD models). Each model is generated from a random sampling of 100K points from the original surface. The clean models are then perturbed by Gaussian noise with the standard deviations from 0.0\% to 2.5\% of the bounding box's diagonal length. The training set consists of 132 models. In addition to the (x,y,z) coordinate of each point, the point normals of clean models are also required for training. To test the model, we randomly selected 20 models from the dataset of PU-GAN \cite{li2019pu} and added different levels of Gaussian noise onto them.

\begin{figure}
	\centering
	\includegraphics[width=1\linewidth]{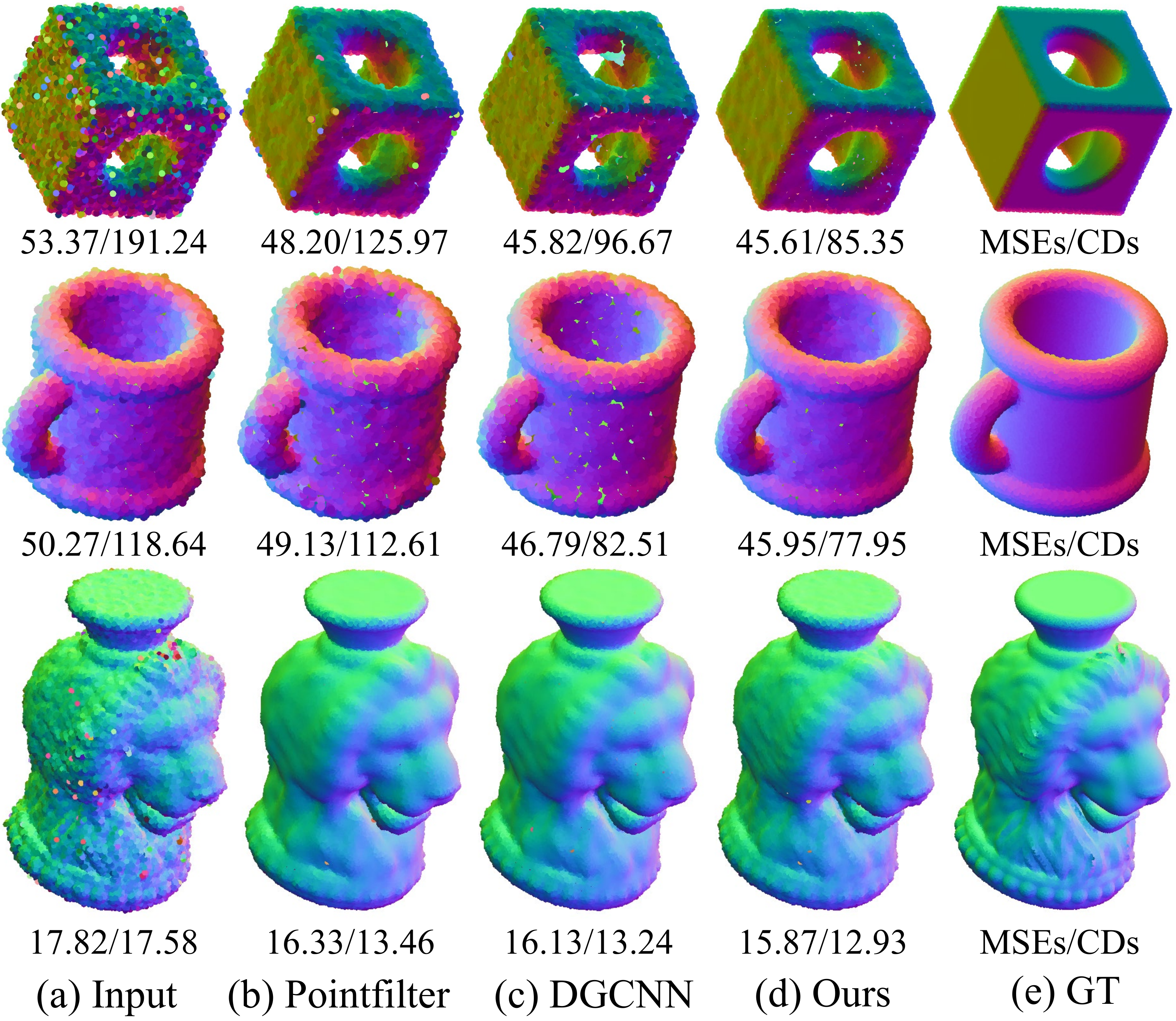}
	\caption{Visual comparisons of point clouds with 0.5\% noise (left: MSEs ($10^{-3}$), right: CDs ($10^{-5}$)).}
	\label{fig:Denoie_comparison}
\end{figure}

\vspace{5pt}\noindent\textbf{Comparison.} To evaluate the effectiveness of iPointfilter, we replace the encoder of Pointfilter with DGCNN \cite{wang2019dynamic}.  As shown in Fig.~\ref{fig:Denoie_comparison}, our method can produce more evenly distributed results on the first two models and maintain more sharp features on the third models than the results of Pointfilter \cite{9207844} and DGCNN \cite{wang2019dynamic}. To comprehensively evaluate our iPointfiler, we calculate the mean square error (MSE) and the chamfer distance (CD) over the 20 synthetic models in the test set. Tab.~\ref{table:denoising-results} shows our method that averagely achieves the lowest errors.

\begin{table}[t]
	\centering
	\small
	\setlength{\tabcolsep}{3.5mm}
	\begin{tabular}{c|cc} 
		\toprule[1pt]
		Methods & MSE ($10^{-3}$) & CD ($10^{-5}$) \\
		\midrule[0.3pt]
		\midrule[0.3pt]
		Noisy   								& 44.77 & 104.12 \\
		Pointfilter                             & 42.31 & 62.35 \\
		DGCNN						            & 42.26 & 59.07 \\
		Ours									& \textbf{40.19} & \textbf{44.96} \\
		\bottomrule[1pt]
	\end{tabular}
	\vspace{5pt}
	\caption{Average errors of all filtered point clouds over our test synthetic models (20 models with 0.5\% noise).}
	\label{table:denoising-results}
\end{table}

\begin{table*}[!ht]
\centering
\footnotesize
\begin{tabular}{ccccccccc}
\toprule[1pt]
Category   & LGAN-AE        & PCN   & 3D-Capsule & TopNet   & MSN         & PF-Net               &ECG-Net       & Ours            \\
\hline\hline
Airplane   & 2.814          & 2.626 & 2.991      & 2.251 & 1.698          & \textbf{0.984}       & 1.095             & 1.010                  \\
Bag        & 8.837          & 8.673 & 8.492      & 7.887 & 9.745          & \textbf{3.543}       & 3.995             & 4.121                  \\
Cap        & 7.609          & 7.126 & 7.706      & 6.524 & 5.491          & 5.473                & 4.668             & \textbf{3.576}         \\
Car        & 5.416          & 5.789 & 6.236      & 5.514 & 5.716          & 2.390                & 2.496             & \textbf{2.356}         \\
Chair      & 4.787          & 4.153 & 4.045      & 3.597 & 3.072          & 2.053                & 2.124             & \textbf{1.916}         \\
Guitar     & 1.251          & 1.113 & 1.294      & 0.976 & 0.836          & \textbf{0.407}       & 0.478             & 0.442                  \\
Lamp       & 7.476          & 6.918 & 7.669      & 6.534 & 3.517          & 4.185                & 3.467             & \textbf{3.182}         \\
Laptop     & 3.376          & 3.262 & 3.627      & 2.671 & 1.619          & 1.448                & 1.408             & \textbf{1.348}         \\
Motorbike  & 4.156          & 4.012 & 4.048      & 3.546 & 2.963          & 1.923                & 2.034             & \textbf{1.888}         \\
Mug        & 6.516          & 6.845 & 7.051      & 6.781 & 8.795          & \textbf{3.377}       & 3.775             & 3.478                  \\
Pistol     & 3.261          & 3.163 & 3.212      & 2.620 & 1.647          & 1.381                & \textbf{1.237}    & 1.271                  \\
Skateboard & 3.022          & 2.906 & 3.346      & 2.717 & 1.760          & 1.327                & 1.354             & \textbf{1.247}         \\
Table      & 4.781          & 4.746 & 5.157      & 4.036 & 4.342          & 2.053                & 1.982             & \textbf{1.922}         \\
\hline
Mean       & 4.869          & 4.717 & 4.990      & 4.281 & 3.938          & 2.349                & 2.316             & \textbf{2.135}         \\
\bottomrule[1pt]
\end{tabular}
\vspace{5pt}
\caption{Completion results on Shapenet-Part with 13 categories by CD. The last row show the mean CD loss of all these categories, scaled by 1000.}
\label{completion_t1}
\end{table*}

\subsection{Point cloud upsampling}
Point cloud upsampling aims to generate dense point clouds from their sparse input. The generated data should recover the fine-grained structures at a higher resolution, and the upsampled points are expected to uniformly lie on the underlying surface.

\begin{figure*}
	\centering
	\includegraphics[width=1\linewidth]{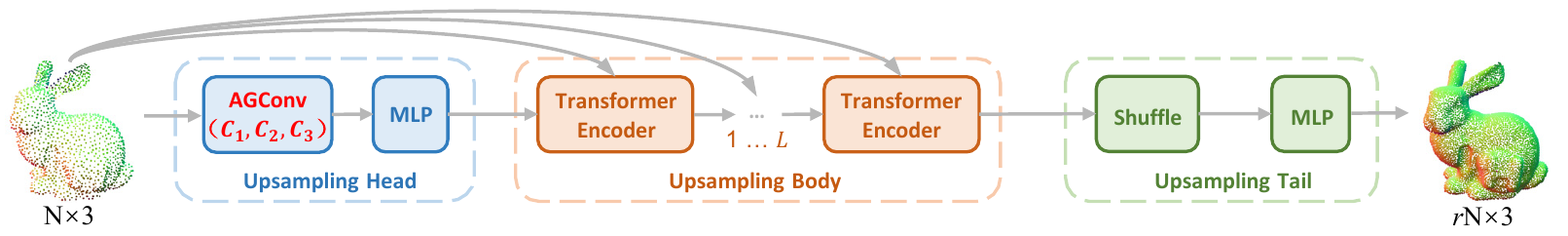}
	\caption{PU-Transformer \cite{2021PU} is improved by our AGConv for point cloud upsampling (call iPU-Transformer).}
	\label{fig:UpsamplingNet}
\end{figure*}

\vspace{5pt}\noindent\textbf{Network framework.}
We leverage PU-Transformer \cite{2021PU} as our network's backbone in Fig.~\ref{fig:UpsamplingNet}.
In our improved PU-Transformer (iPU-Transformer), we add the AGConv module in the upsampling head.
Given a sparse point cloud $P\in R^{N\times3}$ as input, iPU-Transformer can generate a dense point cloud $Q\in R^{rN\times3}$, where $r$ denotes the upsampling scale. In Fig.~\ref{fig:UpsamplingNet}, we first exploit AGConv and MLP to construct the upsampling head, which extracts a preliminary feature map from the input. Then, based on the feature map and the inherent 3D coordinates, the upsampling body gradually encodes a more comprehensive feature map via the cascaded Transformer encoders. Finally, in the upsampling tail, the shuffle operation \cite{2016Real} is used to form a dense feature map and reconstruct the 3D coordinates of $Q$ via an MLP.

\vspace{5pt}\noindent\textbf{Data.}
We train and test iPU-Transformer on the PU1K dataset in PU-GCN \cite{2019PU}. PU1K covers 50 object categories, in which 1,020 3D meshes are used for training and 127 ones for testing. To match the patch-based upsampling methods, the training data is generated from patches of 3D meshes via Poisson disk sampling. Specifically, the training data includes a total of 69,000 samples (patches), where each sample has 256 points (low resolution) and a corresponding ground-truth of 1,024 points ($4\times$ high resolution). 

\begin{figure}
	\centering
	\includegraphics[width=1\linewidth]{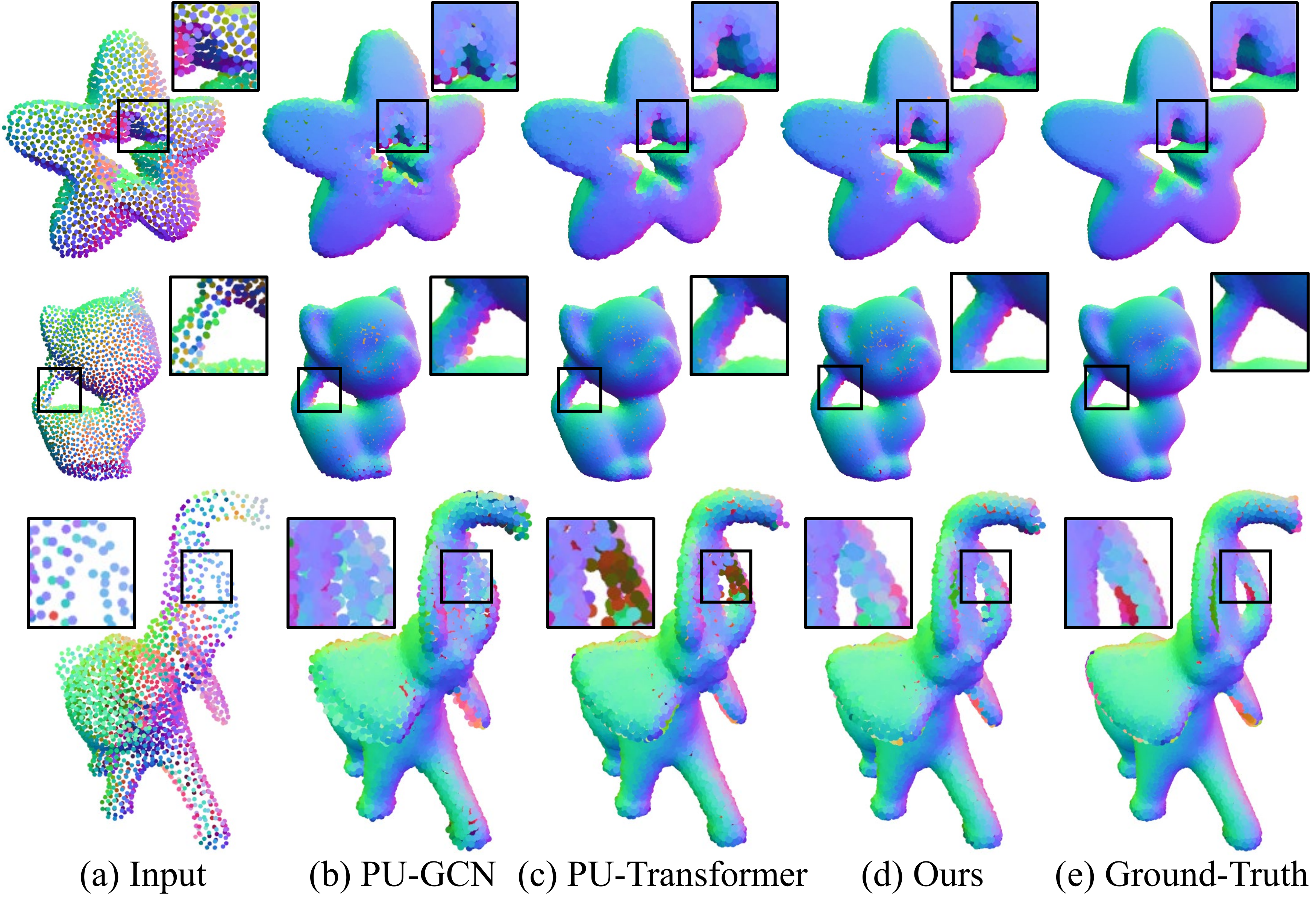}
	\caption{Visualization of upsampling results on models (with 2,048 points for the $4\times$ upsampling) from PU-GAN \cite{li2019pu}.}
	\label{fig:Upsampling-V}
\end{figure}

\vspace{5pt}\noindent\textbf{Results.}
We follow
PU-GCN \cite{2019PU} and PU-Transformer \cite{2021PU}. To be specific, we first cut the input point cloud into multiple seed patches covering all $N$ points. Then, we apply our trained model to upsample the seed patches with a scale of $r$. Finally, the farthest point sampling algorithm is used to combine all upsampled patches as a dense output point cloud with $rN$ points. We test the point clouds with 2,048 points for the $4\times$ upsampling experiments. We quantitatively evaluate the upsampling performance of PU-GCN \cite{2019PU}, PU-Transformer \cite{2021PU}, and our method in Tab. \ref{table:Upsample-Q} based on three widely used metrics: (i) Chamfer distance (CD), (ii) Hausdorff distance (HD), and (iii) Point-to-Surface distance (P2F). iPU-Transformer obtains a lower value under these metrics than its competitors. Moreover, we visually compare our iPU-Transformer with PU-GCN, and PU-Transformer in Fig. \ref{fig:Upsampling-V}. Benefiting from AGConv, iPU-Transformer better upsamples point clouds that have detailed features. 


\begin{table}
	\centering
	\small
	\setlength{\tabcolsep}{3.5mm}
	\begin{tabular}{c|ccc} 
		\toprule[1pt]
		\multirow{2}{*}{Method} & \textbf{CD} & \textbf{HD} & \textbf{P2F} \\
								& ($\times10^{-3}$)   & ($\times10^{-3}$) & ($\times10^{-3}$)  \\ 
		\midrule[0.3pt]
		\midrule[0.3pt]
		PU-GCN \cite{2019PU}	& 0.585 & 7.577 & 2.499 \\
		PU-Transformer \cite{2021PU}	& 0.451 & 3.843 & 1.277\\
		Ours						& \textbf{0.434} & \textbf{3.534} & \textbf{1.251} \\
		\bottomrule[1pt]
	\end{tabular}
	\vspace{5pt}
	\caption{Quantitative comparisons on PU1K \cite{2019PU}.}
	\label{table:Upsample-Q}
\end{table}

\subsection{Circle extraction}
Geometric primitive extraction from man-made engineering objects is essential for many practically meaningful applications, such as reverse engineering  and 3D inspection. The shape of circle is one of the fundamental geometric primitives of man-made engineering objects. Thus, extraction of circles from scanned point clouds is a quite important task in geometry data processing.

\begin{figure}
	\centering
	\includegraphics[width=1\linewidth]{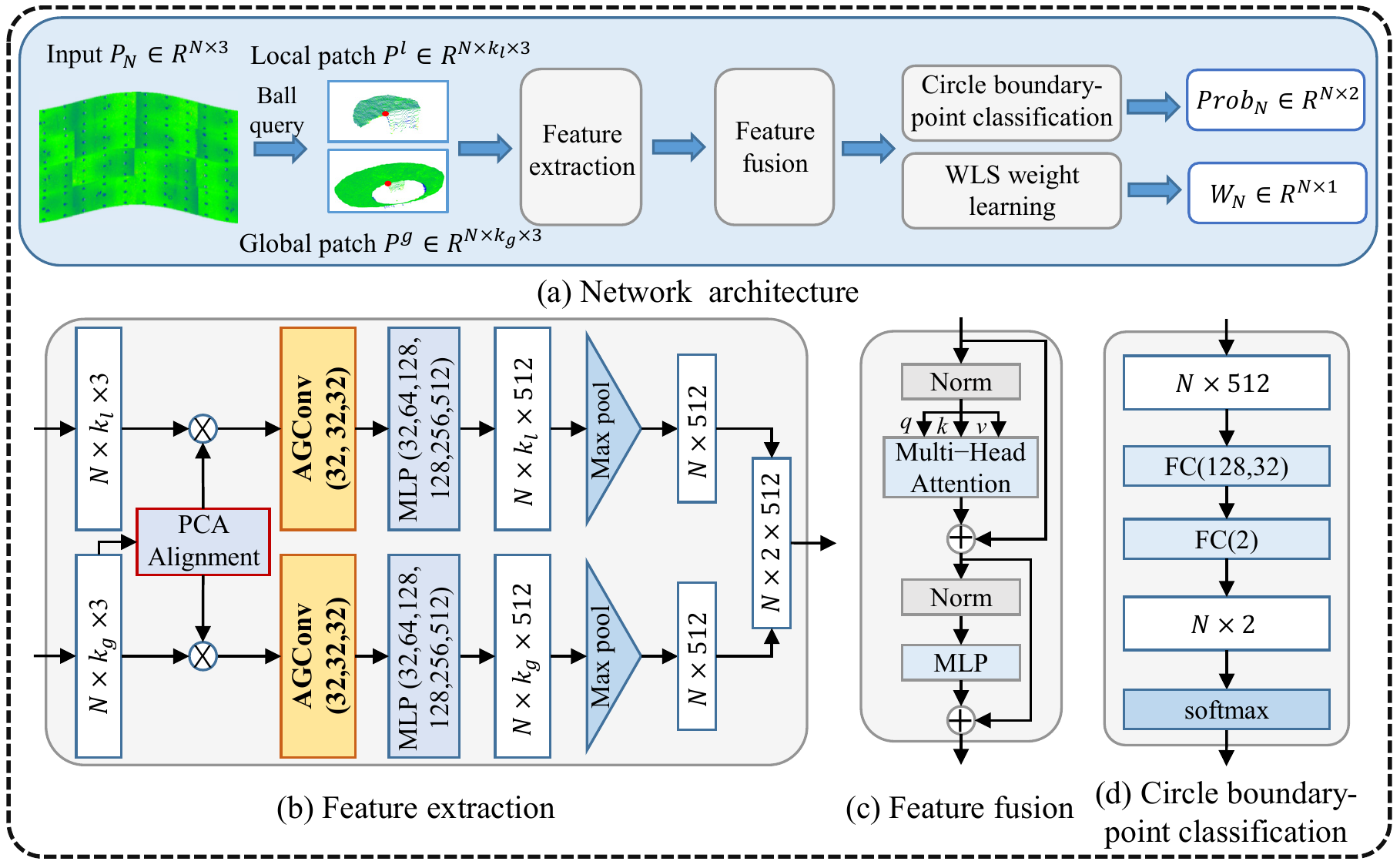}
	\caption{Circle-Net \cite{Honghua2022} (a classification-and-fitting network) is improved by our AGConv for point cloud circle extraction (call iCircle-Net).}
	\label{fig:CircleExtractionNet}
\end{figure}

\vspace{5pt}\noindent\textbf{Network framework.}
We leverage Circle-Net \cite{Honghua2022} as our backbone in Fig.~\ref{fig:CircleExtractionNet}.
In our improved Circle-Net (iCircle-Net), the original graph convolution module is replaced by AGConv. iCircle-Net leverages an end-to-end classification-and-fitting network: The circle-boundary learning module detects all potential circle-boundary points from a raw point cloud by considering local and global neighboring contexts of each
point; the circle parameter learning module for weighted least squares is developed,
without designing any weight metric to avoid the influence of outliers during fitting; the two modules are co-trained with a comprehensive loss to enhance the quality of extracted circles. 

First, we build two different neighborhoods, for perceiving both local and global context information for each point. The deep features of the two patches are then extracted by AGConv and MLPs. Then, a transformer module is used to fuse the features of the two patches, and the fused features are exploited to regress each point's label. After classification, we use a neural network to estimate the weight of each point in the detected circle-boundary candidate points for weighted least squares circle fitting.

\vspace{5pt}\noindent\textbf{Data.}
We train and evaluate our iCircle-Net in the benchmark dataset from Circle-Net \cite{Honghua2022}, which contains 55 CAD models, including curved and flat thin-walled planes with multiple circular structures. Each CAD model is virtually scanned by a simulated scanner, which is developed by Blender, to generate virtually scanned data with different noise intensities and different resolutions. In addition, the circular structures in CAD models have different radii and depths, and the simulated raw data is scanned from different views to mimic more general scanning scenarios. The ground-truth circle primitives is extracted directly from CAD models since the virtually scanned data is consistent with the corresponding CAD models. Through the above schemes, the virtual point cloud data is similar to the real-scanned. Totally, $4,500$ point clouds are created for training.

\begin{figure}
	\centering
	\includegraphics[width=1\linewidth]{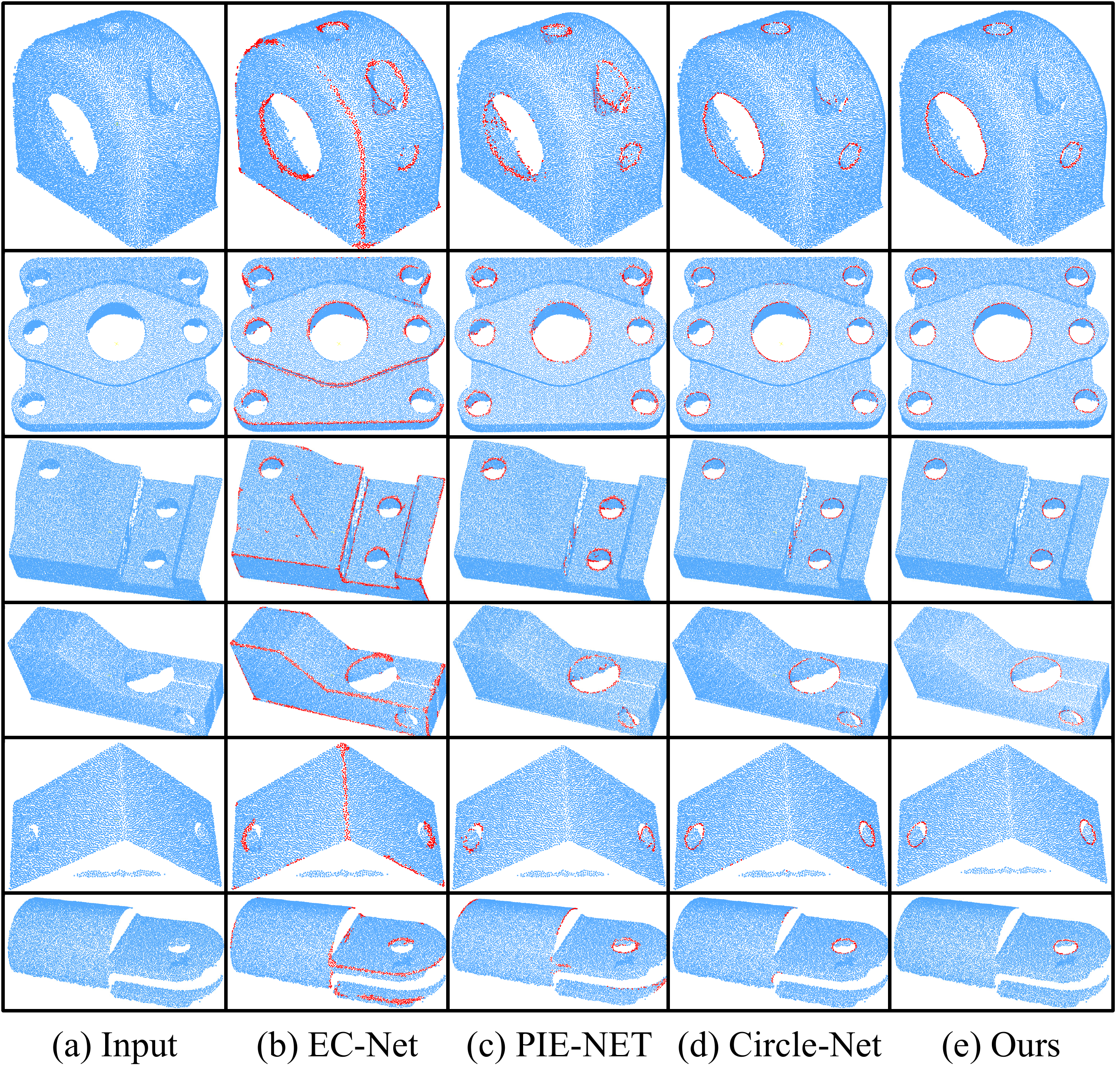}
	\caption{Visualization of circle extraction results on six real scans.}
	\label{fig:CircleExtraction-V}
\end{figure}

\vspace{5pt}\noindent\textbf{Results.}
To demonstrate the effectiveness of AGConv, we compare iCircle-Net against  representative circle extraction methods, including EC-Net \cite{yu2018ec}, PIE-NET \cite{wang2020pie}, and Circle-Net \cite{Honghua2022}. We test all methods with a variety of virtually-scanned clouds in Tab. \ref{table:CircleExtraction-Q}. Benefiting from AGConv, iCircle-Net possesses higher circle boundary detection precision than its competitors. Moreover, the visualization results on several real-scanned point clouds are given in Fig. \ref{fig:CircleExtraction-V}, where our iCircle-Net achieves less detection errors.

\begin{table}
	\centering
	\small
	\setlength{\tabcolsep}{3.5mm}
	\begin{tabular}{c|ccc} 
		\toprule[1pt]
		Method & $Precision (\%)$   & $Recall (\%)$   & $F1 (\%)$\\
		\midrule[0.3pt]
		\midrule[0.3pt]
		EC-Net \cite{yu2018ec}		& 32.60 & 26.82 & 29.43 \\
		PIE-NET \cite{wang2020pie}	& 77.95 & 69.79 & 73.64 \\
		Circle-Net \cite{Honghua2022}				& 86.06 & 77.62 & 81.58\\
		Ours						& \textbf{87.33} & \textbf{78.24} & \textbf{82.54} \\
		\bottomrule[1pt]
	\end{tabular}
	\vspace{5pt}
	\caption{Quantitative comparisons of circle boundary detection on virtual scans.}
	\label{table:CircleExtraction-Q}
\end{table}

\subsection{Point cloud registration}
3D sensors are becoming increasingly available and affordable, which benefit to accurately represent scanned surfaces, detailing the shape, and the size of various objects. However, the raw point clouds directly captured by these sensors unavoidably require a registration step to synthesize a complete model or a large-scale scene from multiple partial scans. Point cloud registration aims to find a rigid transformation to align two point clouds accurately.

\begin{figure}
	\centering
	\includegraphics[width=1\linewidth]{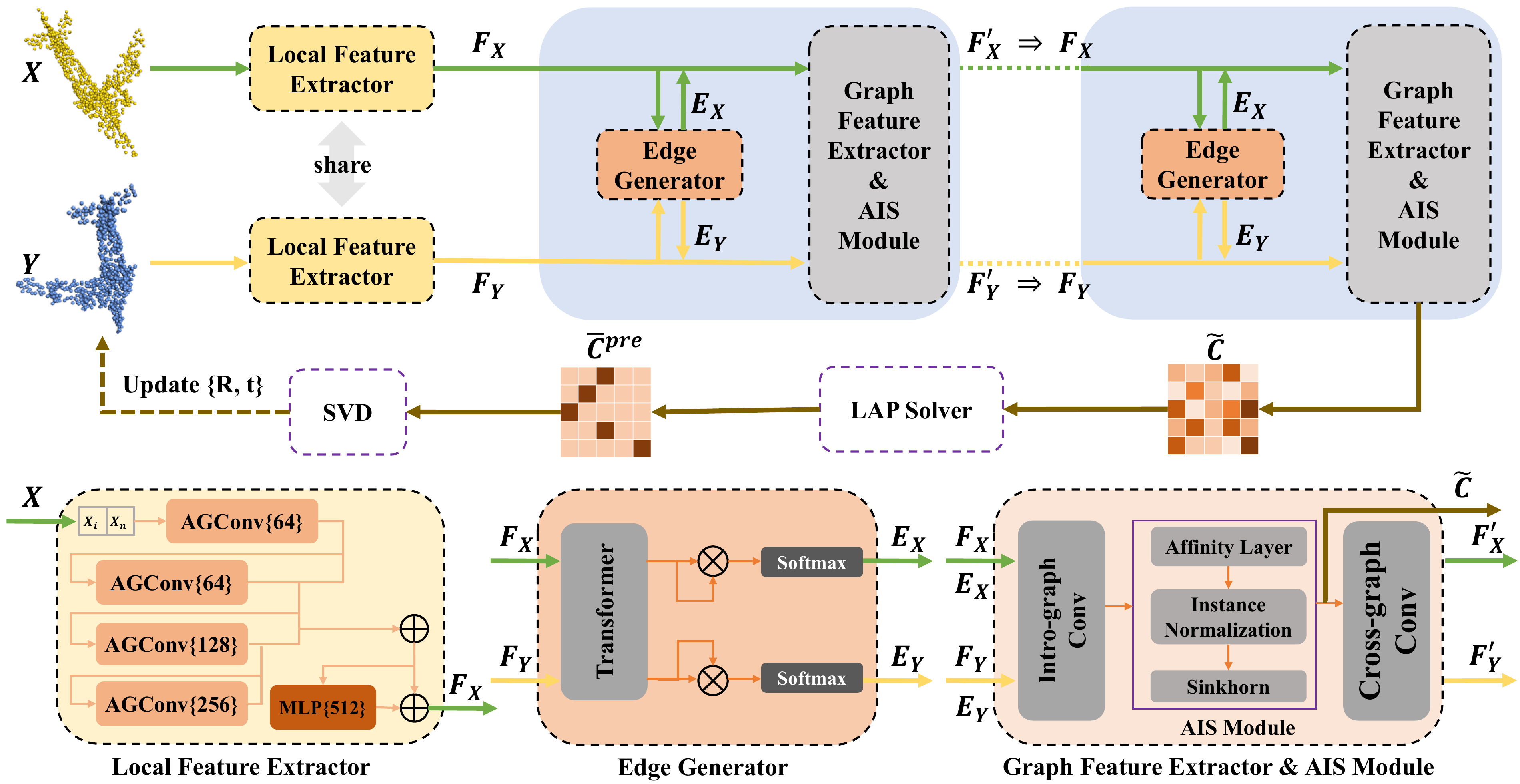}
	\caption{RGM \cite{fu2021robust} is improved by AGConv for registration (called iRGM).}
	\label{fig:iRGM}
\end{figure}

\vspace{5pt}\noindent\textbf{Network framework.}
We utilize RGM \cite{fu2021robust} as our network's backbone in Fig.~\ref{fig:iRGM}. In our improved RGM (iRGM), we replace the original graph convolution with the proposed AGConv module. iRGM consists of four components: a local feature extractor, an edge generator, a graph feature extractor \& AIS module, and an LAP-SVD. First, we use the shared local feature extractor with AGConv to extract discriminative features for each point in \textbf{X} and \textbf{Y}. Then, the edge generator produces edges and builds both the source graph and target graph, and the graphs are inputted into the graph feature extractor. The AIS module predicts the soft correspondence matrix $\widetilde{C}$ between nodes of the two graphs. Finally, the soft correspondences are converted to hard correspondences using the LAP solver, and the transformation is solved by SVD. We also update the transformation iteratively, similar to ICP. 

\vspace{5pt}\noindent\textbf{Data.}
All experiments are conducted on ModelNet40 \cite{wu20153d}. It includes 12,311 meshed CAD models from 40 categories. Following RGM \cite{fu2021robust}, we randomly sample 2,048 points from the mesh faces and re-scale the points into a unit sphere. Each category consists of official train/test splits. To select models for evaluation, we take 80\% and 20\% of the official train split as the training set and validation set, respectively, and the official test split for testing. For each object in the dataset, we randomly sample 1,024 points as the source point cloud \textbf{X}, and then apply a random transformation on \textbf{X} to obtain the target point cloud \textbf{Y} and shuffle the point order. For the transformation applied, we randomly sample three Euler angles in the range of ${[0, 45]}^{\circ}$ for rotation and three displacements in the range of $[-0.5, 0.5]$ along each axis for translation. Moreover, we focus on the challenging partial-to-partial case. In order to generate partial overlapping pairs, we create a random plane passing through each point cloud independently, translate it along its normal, and retain 70\% of the points as RPM-Net \cite{yew2020rpm}.

\begin{figure}
	\centering
	\includegraphics[width=1\linewidth]{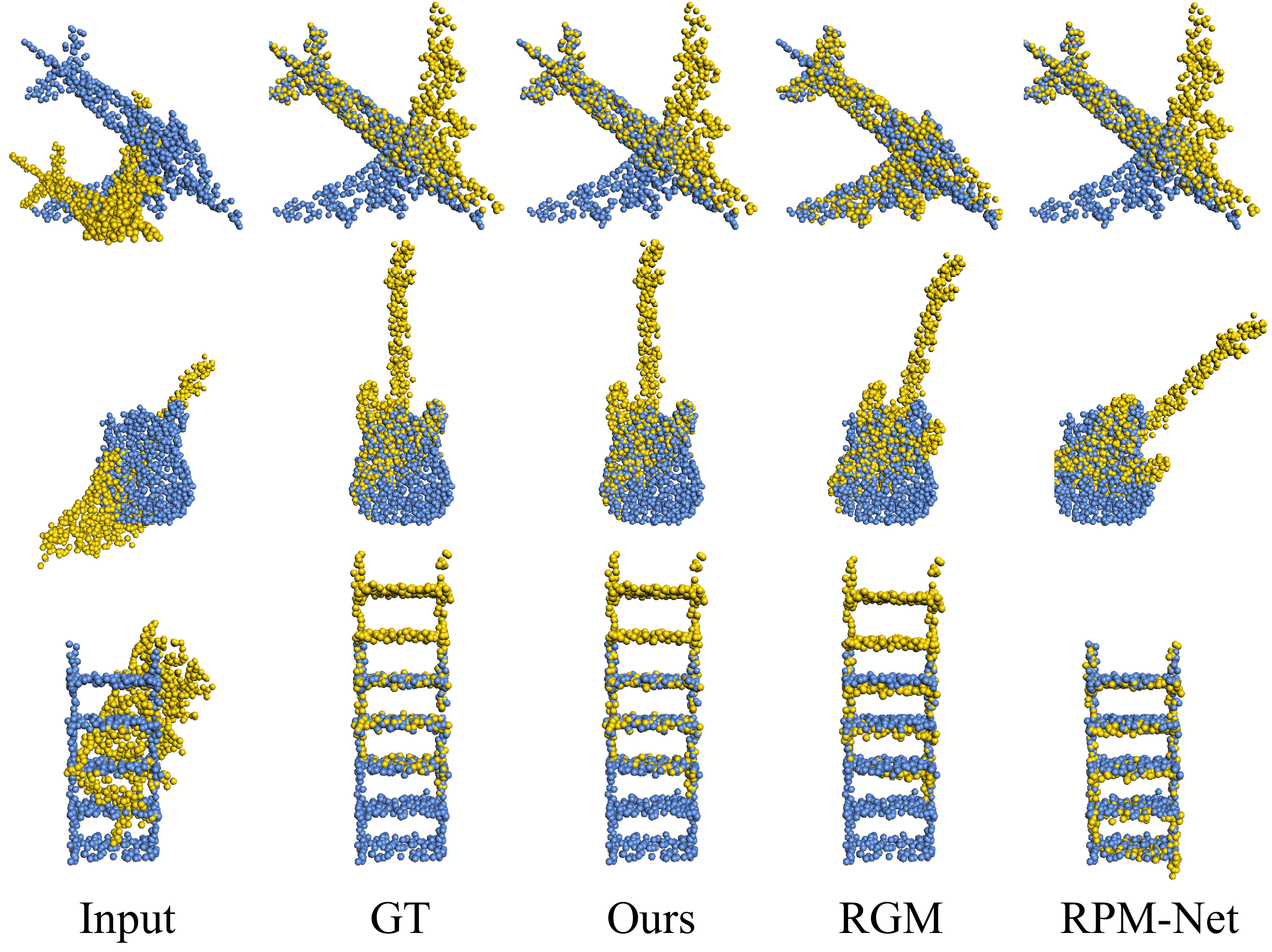}
	\caption{Visualization of point cloud registration on ModelNet40 \cite{wu20153d}.}
	\label{fig:PCR_VIS}
\end{figure}

\vspace{5pt}\noindent\textbf{Results.}
To show the effectiveness of AGConv, we compare iRGM against two methods, i.e., RPM-Net \cite{yew2020rpm} and RGM \cite{fu2021robust}. We train them in the same way and evaluate their performance over four metrics: the mean isotropic errors (MIE) and the mean absolute errors (MAE) of rotation and translation, as shown in Tab. \ref{table:PCR}. The robust features extracted by AGConv boost the performance of iRGM, which surpass the two methods over all metrics. The visualization results in Fig.~\ref{fig:PCR_VIS} demonstrate that our iRGM aligns two point clouds more accurately.

\begin{table}
	\centering
	\small
	\begin{tabular}{c|cccc} 
		\toprule[1pt]
		Method & MAE($R$)   & MAE($t$)   & MIE($R$)  & MIE($t$)\\
		\midrule[0.3pt]
		\midrule[0.3pt]
		RPM-Net \cite{yew2020rpm}		& 0.869 & 0.0082 & 1.711 & 0.0175\\
		RGM \cite{fu2021robust}				& 0.541 & 0.0046 & 1.032 & 0.0096\\
		Ours						& \textbf{0.493} & \textbf{0.0041} & \textbf{0.967} & \textbf{0.0088}\\
		\bottomrule[1pt]
	\end{tabular}
	\vspace{5pt}
	\caption{Quantitative comparisons of registration methods on ModelNet40 \cite{wu20153d}.}
	\label{table:PCR}
\end{table}
\label{sec:eval}

\section{Conclusion}

Deep learning on 2D images has boomed because of its superiority in solving computer vision tasks. Deep learning on 3D point clouds has also drawn much interests in recent years. 
However, it is still far from being satisfactory to leverage the potential of deep learning for understanding point clouds.
In this paper, we propose a novel adaptive graph convolution (AGConv) for point cloud analysis. The main contribution of our method lies in the designed adaptive kernel in the graph convolution, which is dynamically generated according to the point features. Instead of using a fixed kernel that captures correspondences indistinguishably between points, our AGConv can produce learned features that are more flexible to shape geometric structures.  
We have applied AGConv to train end-to-end deep networks for several point cloud analysis tasks, including the low-level geometry processing tasks, i.e., completion, denoising, upsampling and registration, and the high-level geometry processing tasks, i.e., classification, segmentation and circle extraction. In all these tasks, AGConv outperforms the state-of-the-arts on the benchmark datasets. Collecting-and-annotating large-scale point clouds is time-consuming and expensive. To alleviate it, we attempt to propose unsupervised learning approaches to learn features from unlabeled point cloud datasets by AGConv and geometry domain knowledge in future.

\label{sec:conclusion}

\vspace{5pt}


%

\ifCLASSOPTIONcompsoc
\else
\fi


\ifCLASSOPTIONcaptionsoff
  \newpage
\fi

\bibliographystyle{IEEEtran}
\bibliography{main}

\end{document}